\documentclass{article}

\usepackage{microtype}
\usepackage{graphicx}
\usepackage{subfigure}
\usepackage{booktabs} 

\usepackage{hyperref}

\usepackage[accepted]{icml2024}

\usepackage{amsmath}
\usepackage{amssymb}
\usepackage{mathtools}
\usepackage{amsthm}

\usepackage[utf8]{inputenc} 
\usepackage[T1]{fontenc}    
\usepackage{url}            
\usepackage{amsfonts}       
\usepackage{nicefrac}       
\usepackage{xcolor}         

\usepackage{tikz}
\usetikzlibrary{positioning}
\usepackage{changepage}
\usepackage{sidecap}
\usepackage{booktabs}
\usetikzlibrary{calc}
\usetikzlibrary{shapes, fit, backgrounds, decorations.pathreplacing}

\usepackage{ulem} 
\usepackage{array}
\usepackage{amsthm}
\usepackage{algorithm}
\usepackage{algorithmic}
\usepackage{amsfonts}
\usepackage{enumerate}
\usepackage{subfigure}
\usepackage{bm}
\usepackage{tabularx}
\usepackage{enumitem}
\usepackage{multirow}
\usepackage{xcolor}
\usepackage{nicefrac}
\usepackage{booktabs}
\usepackage{bbding}
\usepackage{caption}
\usepackage{graphicx}
\usepackage{bbding}
\usepackage{wrapfig}
\usepackage{lipsum}
\usepackage{amssymb}
\usepackage{multibib}
\usepackage{tcolorbox}
\usepackage{bm}
\usepackage{makecell}
\usepackage{amsthm}
\usepackage{amssymb}
\usepackage{amsmath}
\usepackage{diagbox}
\usepackage{wrapfig}
\usepackage{graphicx}
\usepackage{subfigure}
\usepackage{enumitem}
\usepackage{soul}
\usepackage{pifont}
\usepackage{bbm}
\usepackage{pgf}
\usepackage{tikz}
\usetikzlibrary{positioning}
\usepackage{changepage}
\usepackage{sidecap}
\usepackage{booktabs}
\usetikzlibrary{calc}
\usetikzlibrary{shapes, fit, backgrounds, decorations.pathreplacing}

\usepackage{listings}
\usepackage{placeins}

\usepackage{minitoc}

\usepackage[capitalize,noabbrev]{cleveref}

\theoremstyle{plain}

\theoremstyle{definition}

\theoremstyle{remark}

\usepackage[textsize=tiny]{todonotes}

\usepackage{colortbl,color}
\definecolor{shadecolor}{rgb}{0.92,0.92,0.92}
\definecolor{grey}{rgb}{0.5, 0.5, 0.5}
\definecolor{greyC}{RGB}{180,180,180}
\definecolor{greyL}{RGB}{235,235,235}

\definecolor{boxcolor1}{HTML}{D6EAF8}
\definecolor{boxcolor2}{HTML}{64B5F6}
\usepackage{caption}
\usepackage{etoc}
\etocdepthtag.toc{mtchapter}
\etocsettagdepth{mtchapter}{subsection}
\etocsettagdepth{mtappendix}{none}

\icmltitlerunning{
Envisioning Outlier Exposure by Large Language Models for Out-of-Distribution Detection
}

\begin{document}
	
	\twocolumn[
	\icmltitle{Envisioning Outlier Exposure by Large Language Models for Out-of-Distribution Detection
 }
        
	\icmlsetsymbol{equal}{*}
	\begin{icmlauthorlist}
		\icmlauthor{Chentao Cao}{hkbu}
		\icmlauthor{Zhun Zhong}{hfut,Nottingham}
		\icmlauthor{Zhanke Zhou}{hkbu}
		\icmlauthor{Yang Liu}{ucsc}
		\icmlauthor{Tongliang Liu}{sydney}
		\icmlauthor{Bo Han}{hkbu}
	\end{icmlauthorlist}
	
	\icmlaffiliation{hkbu}{TMLR Group, Department of Computer Science, Hong Kong Baptist University}
        \icmlaffiliation{hfut}{School of Computer Science and Information Engineering, Hefei University of Technology}
	\icmlaffiliation{Nottingham}{School of Computer Science, University of Nottingham}
	\icmlaffiliation{ucsc}{Computer Science and Engineering, University of California, Santa Cruz}
        \icmlaffiliation{sydney}{Sydney AI Centre, The University of Sydney}

        \icmlcorrespondingauthor{Bo Han}{bhanml@comp.hkbu.edu.hk}
	\icmlcorrespondingauthor{Zhun Zhong}{zhunzhong007@gmail.com}

	\icmlkeywords{OOD Detection, LLM}
	
	\vskip 0.3in
	]
	
	
	
	\printAffiliationsAndNotice{}  

\begin{abstract}
Detecting out-of-distribution (OOD) samples is essential when deploying machine learning models in open-world scenarios. Zero-shot OOD detection, requiring no training on in-distribution (ID) data, has been possible with the advent of vision-language models like CLIP. Existing methods build a text-based classifier with only closed-set labels. However, this largely restricts the inherent capability of CLIP to recognize samples from large and open label space. In this paper, we propose to tackle this constraint by leveraging the expert knowledge and reasoning capability of large language models (LLM) to Envision potential Outlier Exposure, termed EOE, without access to any actual OOD data. Owing to better adaptation to open-world scenarios, EOE can be generalized to different tasks, including far, near, and fine-grained OOD detection. Technically, we design (1) LLM prompts based on visual similarity to generate potential outlier class labels specialized for OOD detection, as well as (2) a new score function based on potential outlier penalty to distinguish hard OOD samples effectively. Empirically, EOE achieves state-of-the-art performance across different OOD tasks and can be effectively scaled to the ImageNet-1K dataset.
The code is publicly available at: \url{https://github.com/tmlr-group/EOE}.
\end{abstract}

\section{Introduction}
Machine learning models excel in closed-set scenarios, where training and testing datasets share identical distribution. However, in open-world settings, especially in high-stakes scenarios like autonomous driving where the consequence of making an error can be fatal, these models often encounter out-of-distribution~(OOD) samples that fall outside the label space of the training dataset, leading to unpredictable and frequently erroneous model behaviors. Consequently, there is a growing interest in OOD detection~\citep{yang2021oodsurvey, zhang2023openood, salehi2021unified},  aiming to distinguish OOD samples from test-time data while maintaining classification accuracy.

Most existing OOD detection methods~\citep{hendrycks17baseline, lee2018simple,hendrycks2019oe, liu2020energy, sehwag2021ssd} can effectively detect OOD samples based on a well-trained in-distribution~(ID) classifier. However, they are constrained to ID datasets with different label spaces.
Besides, these methods solely depend on vision patterns, ignoring the connection between visual images and textual labels. Recently, MCM~\cite{ming2022delving} introduced the setting of zero-shot OOD detection, which aims to leverage the capabilities of large-scale vision-language models~(VLMs), \textit{e.g.}, CLIP~\citep{radford2021learning}, to detect OOD samples across diverse ID datasets without training samples. By constructing a textual classifier with only ID class labels, MCM achieves impressive performance compared to previous OOD detection methods. 

\begin{figure*}
\begin{center}
\includegraphics[width=\textwidth]{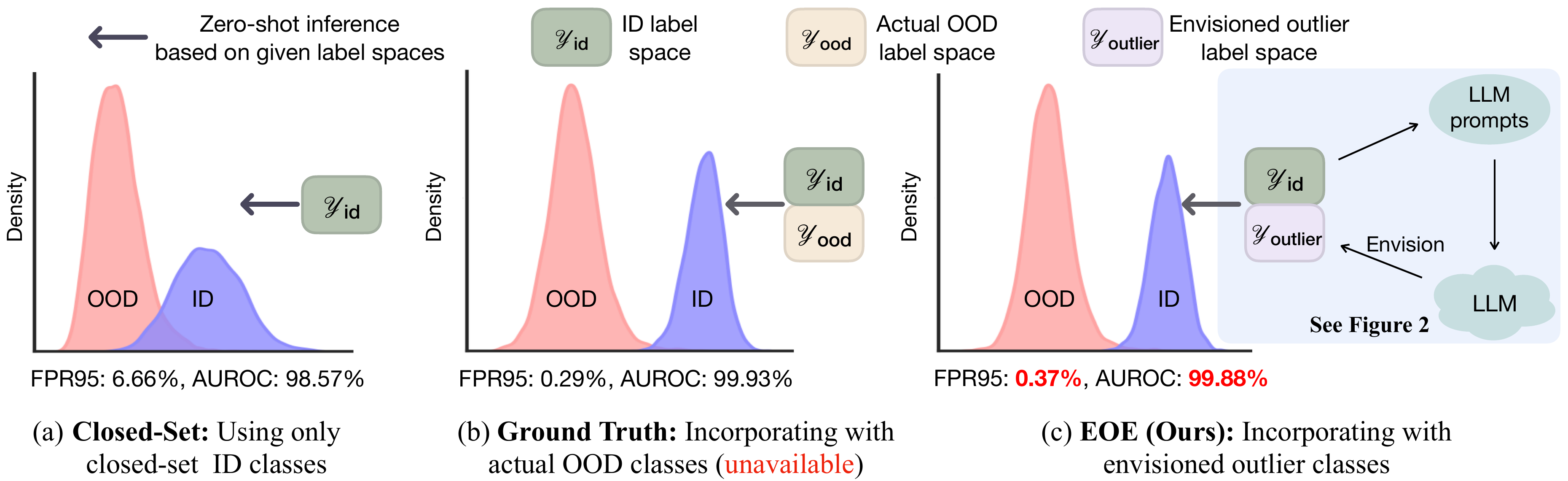}
\end{center}
  \caption{Comparison of zero-shot OOD detection score distribution. Compared to the model using (a) only closed-set ID classes, (b) adding actual OOD class labels can largely increase the OOD detection performance. (c) By adding the outlier classes generated by our method, the OOD detection results can also be significantly improved without using the actual OOD class labels. We use CUB-200-2011~~\citep{wah2011caltech} as ID classes and Places~\citep{zhou2017places} as OOD classes.
  }
  \label{fig1: toy illustration}
\vspace{-.12in}
\end{figure*}

However, such an approach often fails when encountering hard OOD samples, as shown in Figure~\ref{fig1: toy illustration}~(a). One might wonder 1) if this issue arises because the pre-trained models (\textit{e.g.}, CLIP) are not strong enough or require further fine-tuning; or 2) if it is attributable to the usages of these pre-trained models, \textit{e.g.}, \textit{an exclusive reliance on closed-set ID classes}.
Surprisingly, our findings suggest that CLIP can achieve superior OOD detection results by incorporating actual OOD class labels, as depicted in Figure~\ref{fig1: toy illustration}~(b).
This reinforces that relying solely on ID class labels is inadequate for distinguishing hard OOD samples. Unfortunately, we are unable to access the actual OOD label space in practical open-world scenarios. Therefore, we raise an open question:
\begin{center}
\textit{Is it possible to generate the potential outlier class labels for OOD detection without access to test-time data?}
\end{center}

To answer this question, we take a step further in this work and ponder whether we can employ large language models~(LLMs) to address this challenge. 
We propose a knowledge-enhanced approach that harnesses the expert knowledge and reasoning capabilities of LLMs to Envision potential Outlier Exposure, termed EOE, without relying on any actual or auxiliary OOD data, as shown in Figure~\ref{fig1: toy illustration}~(c).
Technically, we design LLM prompts to generate potential outlier class labels specialized for OOD detection, following a visual similarity rule. 
For example, "\textit{Give three categories that visually resemble a horse}", in which "\textit{horse}" is an ID class. 
Furthermore, we introduce a new score function based on potential outlier penalty to distinguish hard OOD samples effectively.
Different from ZOC~\citep{esmaeilpour2022zero} and CLIPN~\citep{wang2023clipn}, which also attempt to generate ``NOT ID'' classes, ZOC requires additional training on a text-based image description generator, and CLIPN necessitates a large dataset to train the extra CLIP encoder. EOE only utilizes ID class labels to generate outlier classes.

The proposed EOE brings significant performance improvements 
and enjoys the advantages of: 
(1)~\textit{OOD-Agnostic}, which does not require any prior knowledge of the unknown OOD data; (2)~\textit{Zero-Shot}, which serves various task-specific ID datasets with a single pre-trained model; (3)~\textit{Scalability and Generalizability}, which effectively adapts to large-scale datasets such as ImageNet-1K~\citep{deng2009imagenet} with $1000$ ID classes,
and is flexible and generalizable
across far, near, and fine-grained OOD detection tasks. 

Our contributions can be summarized as follows:
\vspace{-.03in}
\begin{itemize}[leftmargin=.0in]
\item We propose a new perspective that leverages expert knowledge from LLM to envision potential outlier class labels, facilitating OOD detection~(Section~\ref{section3}).
\vspace{-.03in}
\item We propose EOE, providing LLM prompts based on the visual similarity rule to envision potential outlier class labels. A score function is further designed based on potential outlier penalty, helping the model effectively distinguish between ID samples and OOD samples~(Section~\ref{section3}).
\vspace{-.03in}
\item 
Our EOE is superior, significantly outperforming existing methods.
EOE achieves improvements of
$2.47\%$, $2.13\%$, $3.59\%$, and $12.68\%$ on the far OOD, near OOD, fine-grained OOD, and ImageNet-1K far OOD detection tasks in terms of FPR95~(Section~\ref{exp}).
\end{itemize}

\begin{figure*}[t]
\begin{center}
\includegraphics[width=\textwidth]{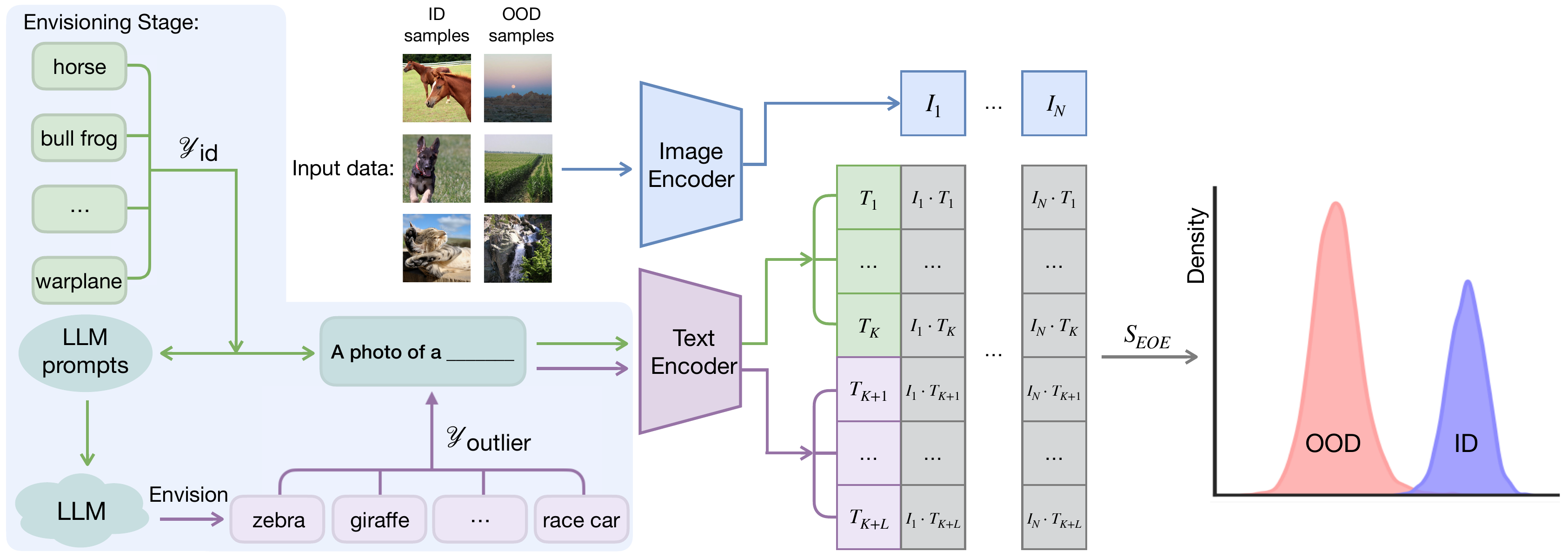}
\end{center}
  \caption{The framework of the proposed EOE. Given a set of ID class labels \( \mathcal{Y}_\text{id} \), we first leverage the designed prompts to generate a set of outlier class labels, \( \mathcal{Y}_\text{outlier} \), by using a LLM. Then, we input both the ID and generated OOD class labels into the text encoder for building the textual classifier. During the test stage, given an input image, we obtain the visual feature by the image encoder and calculate the similarities between the visual feature and the textual classifier. Finally, the OOD score is obtained by scaling the similarities with the proposed detection score function $S_{\text{EOE}}$.
  }
   \label{fig: framework}
  \vspace{-.05in}
\end{figure*}

\section{Preliminaries} 
\textbf{Contrastive Language-Image Pre-training~(CLIP)}~\citep{radford2021learning} is trained on 400 million (image, text) pairs collected from the internet using self-supervised contrastive representation learning~\citep{tian2020contrastive}. The model comprises an image encoder $\mathcal{I}(\cdot)$, adopting either ViT~\citep{dosovitskiy2020vit} or ResNet~\citep{he2016deep} architecture, and a text encoder $\mathcal{T}(\cdot)$, utilizing the Transformer~\citep{vaswani2017attention} architecture.
During testing, the model evaluates the similarity of visual and textual features to choose the best-matching class label.
CLIP enables open-set textual inputs as class labels without retraining or fine-tuning based on specific training data. This feature renders it widely applied to zero-shot downstream tasks, such as visual classification, detection, segmentation, and so on.

\textbf{Large Language Models (LLMs)} refer to natural language processing models trained on massive data, with more than hundreds of billions or more parameters, such as GPT-3~\citep{brown2020language} GPT-4~\citep{gpt4}, PaLM~\citep{chowdhery2022palm} and LLaMA~\citep{touvron2023llama}. These models demonstrate the proficiency to understand and generate natural language text, thus enabling them to undertake a multitude of linguistic tasks. Considering the comprehensive nature of their training datasets, which span a wide variety of knowledge domains, the expert knowledge embedded in LLMs can be employed to provide class labels visually similar to the ID class to meet our needs.

\textbf{Zero-Shot OOD Detection}~\citep{esmaeilpour2022zero, ming2022delving, wang2023clipn} enables detecting OOD samples across diverse ID datasets using the same pre-trained model, \textit{e.g.}, CLIP, without re-training on any unseen ID data. It can be viewed as a binary classification problem:
\begin{equation}
    G_{\lambda}(x; \mathcal{Y}_\text{id}, \mathcal{I}, \mathcal{T})=\begin{cases} 
      \text{ID} & S(x)\ge \lambda \\
      \text{OOD} & S(x) < \lambda 
   \end{cases},
\end{equation}
where $G(\cdot)$ is the OOD detector, $x$ denotes the input image, $x \! \in \! \mathcal{X}, \mathcal{X} \! = \! \{\text{ID}, \text{OOD}\}$ and $\mathcal{Y}_\text{id}$ defines the space of ID class labels. The OOD detection score function $S$ is derived from the similarity between the visual representation $\mathcal{I}(x)$ and textual representation $\mathcal{T}(t)$, $t$ is the textual input to the text encoder,
such as \textit{"a photo of a <ID class>"}, 
and $\lambda$ is the threshold to distinguish ID/OOD classes.

\section{Envisioning Outlier Exposure for Zero-shot OOD Detection}
\label{section3}
In this paper, we aim to enhance zero-shot OOD detection performance by harnessing LLM to generate potential outlier class labels. However, several challenges require attention: (1)~How to guide LLM to generate the desired outlier class label? 
(2)~How can we sharpen the distinction between ID and OOD samples given the envisioned outlier class label? To address these issues, we propose LLM prompts specifically tailored for OOD detection and introduce a novel score function for better differentiation. The overall framework of our method is illustrated in Figure~\ref{fig: framework}.

\subsection{Acquiring Envisioned Outlier Class Labels}
We first categorize OOD detection tasks into three types: 
far, near, and fine-grained OOD detection. 
Then, we elaborate the corresponding three LLM prompts anchored on \textit{visual similarity} to generate outlier class labels as follows, which are general to different datasets for each OOD task. 

\textbf{Far OOD Detection} refers to identifying OOD classes that are distant from the ID classes in the label space, with most being effectively discerned. Building upon the existing ID class labels, we first guide LLM to summarize these classes and determine their respective major categories. Subsequently, we ask LLM to provide outlier class labels that are visually similar to these major categories. Since LLM first summarizes the ID classes into major categories, our approach can be easily extended to large-scale datasets, such as ImageNet-1K. The LLM prompt for far OOD detection is illustrated in Figure~\ref{fig: far ood LLM prompts}. 
Wherein, the $\mathcal{Y}_\text{id}$ represents the set of ID class labels, $\mathcal{Y}_\text{id} = \{y_1, y_2, \cdots, y_K\}$, and the $K$ signifies the total number of categories encompassed within the ID class labels. Similarly, $\mathcal{Y}_\text{outlier}$ indicates the set of envisioned outlier class labels generated by LLM, and $\mathcal{Y}_\text{outlier} = \{n_1, n_2, \cdots, n_L\}$. 

\begin{figure}[!t]
    \centering
    \begin{tikzpicture}
        \small
        \definecolor{chatcolor1}{HTML}{5fedb7} 
        \definecolor{shadecolor}{gray}{0.9}
        \definecolor{promptcolor}{HTML}{D6EAF8}  
        \fontfamily{cmss}\selectfont
        
        \node[align=left, text width=0.45\textwidth, fill=shadecolor, rounded corners=1mm, anchor=north west] (node1) at (0,0) {\textbf{Q:} I have gathered images of $K$ distinct categories: $\mathcal{Y}_\text{id}$. Summarize what broad categories these categories might fall into based on \textbf{visual features}. Now, I am looking to identify $L$ classes that \textbf{visually resemble} these broad categories but have no direct relation to these broad categories. Please list these $L$ categories for me.};
        \node[align=left, text width=0.22\textwidth, fill=chatcolor1, rounded corners=1mm, anchor=north east] (node2) at ([yshift=-0.2cm]node1.south -| {$(0,0)+(0.45\textwidth,0)$}) {\textbf{A:} These $L$ categories are:};

        \node[anchor=west, font=\selectfont] at (node1.south west |- node2) {\textbf{Far OOD prompt}};
        
        \node[draw, black, line width=1pt, rounded corners=1mm, inner sep=4pt, fit=(node1) (node2)] {};

    \end{tikzpicture}
    \caption{LLM prompt for far OOD detection, consisting of both the contents of \textbf{Q} and \textbf{A}.}
    \label{fig: far ood LLM prompts}
    \vspace{-8pt}
\end{figure}

\textbf{Near OOD Detection} pertains to identifying OOD classes that are relatively close to the ID class, \textit{e.g.}, "\textit{horse}" and "\textit{zebra}", presenting an increased propensity to come across OOD samples that bear visual resemblances to ID classes. Consequently, for each ID class label, we instruct LLM to provide $l$ outlier class labels that exhibit visual resemblance with ID class labels, \textit{i.e.}, $l \times K = L$. 
Overlapping classes in $\mathcal{Y}_\text{outlier}$ with $\mathcal{Y}_\text{id}$ are removed by string matching. 
The LLM prompt for near OOD detection is shown in Figure~\ref{fig: near ood LLM prompts}.
\begin{figure}[!t]
    \centering
    \begin{tikzpicture}
        \small
        \definecolor{chatcolor1}{HTML}{5fedb7} 
        \definecolor{shadecolor}{gray}{0.9}
        \definecolor{promptcolor}{HTML}{D6EAF8}  
        \fontfamily{cmss}\selectfont
        
        \node[align=left, text width=0.45\textwidth, fill=shadecolor, rounded corners=1mm, anchor=north west] (node1) at (0,0) {\textbf{Q:} Given the image category $y_i$, please suggest \textbf{visually similar} categories that are not directly related or belong to the same primary group as $y_i$. Provide suggestions that \textbf{share visual characteristics} but are from broader and different domains than $y_i$.};
        \node[align=left, text width=0.26\textwidth, fill=chatcolor1, rounded corners=1mm, anchor=north east] (node2) at ([yshift=-0.2cm]node1.south -| {$(0,0)+(0.45\textwidth,0)$}) {\textbf{A:} There are $l$ classes similar to $y_i$, and they are from broader and different domains than $y_i$:};

        \node[anchor=west, font=\selectfont] at (node1.south west |- node2) {\textbf{Near OOD prompt}};
        \node[draw, black, line width=1pt, rounded corners=1mm, inner sep=4pt, fit=(node1) (node2)] {};
    \end{tikzpicture}
    \caption{LLM prompt for near OOD detection.}
    \label{fig: near ood LLM prompts}
    \vspace{-8pt}
\end{figure}

\begin{figure}[!t]
    \centering
    \begin{tikzpicture}
        \small
        \definecolor{chatcolor1}{HTML}{5fedb7} 
        \definecolor{shadecolor}{gray}{0.9}
        \definecolor{promptcolor}{HTML}{D6EAF8}  
        \fontfamily{cmss}\selectfont
        
        \node[align=left, text width=0.45\textwidth, fill=shadecolor, rounded corners=1mm, anchor=north west] (node1) at (0,0) {\textbf{Q:} I have a dataset containing $K$ different species of \textit{class-type}. I need a list of $L$ distinct \textbf{\textit{class-type} species} that are NOT present in my dataset, and ensure there are no repetitions in the list you provide. For context, the species in my dataset are: $\mathcal{Y}_\text{id}$.};
        \node[align=left, text width=0.2\textwidth, fill=chatcolor1, rounded corners=1mm, anchor=north east] ((node2) at ([yshift=-0.2cm]node1.south -| {$(0,0)+(0.45\textwidth,0)$}) {\textbf{A:} The other $L$ \textit{class-type} species not in the dataset are:};

        \node[anchor=west, font=\selectfont] at (node1.south west |- node2) {\textbf{Fine-grained OOD prompt}};
        \node[draw, black, line width=1pt, rounded corners=1mm, inner sep=4pt, fit=(node1) (node2)] {};
    
    \end{tikzpicture}
    \caption{LLM prompt for fine-grained OOD detection.}
    \label{fig: fine-grained LLM prompts}
    \vspace{-8pt}
\end{figure}

\textbf{Fine-grained OOD Detection}, also known as open-set recognition~(OSR)~\citep{vaze2021open}, focuses on semantic shift instead of distributional shift primarily comprised in previous OOD detection. 
In fine-grained OOD detection, both ID and OOD samples fall under the same major category~(\textit{e.g.}, "\textit{bird}"), and intrinsic visual similarities exist among subclasses~(\textit{e.g.}, "\textit{Frigatebird}", "\textit{Ovenbird}"). Therefore, it is more appropriate to instruct the LLM to provide different subclasses within the same major category directly. 
The LLM prompt for fine-grained OOD detection is presented below Figure~\ref{fig: fine-grained LLM prompts}, where $\textit{class-type}$ refers to the major category, such as "\textit{bird}".\footnote{For detailed LLM prompts and the outlier class labels generated by LLM, please refer to Appendix~\ref{example of LLM prompt}.}


\begin{table*}[tb]
\caption{ Zero-shot \textbf{far} OOD detection results. The \textbf{bold} indicates the best performance on each dataset. The \colorbox{gray!22}{gray} indicates that the comparative methods require an additional massive auxiliary dataset. Ground Truth is the result of incorporating the actual OOD class labels, which are unavailable in fact.}
\label{tab: far-ood}
\centering
\fontsize{7}{8}\selectfont
\setlength\tabcolsep{4.5pt}
\begin{tabular}{ccccccccccccc}
\toprule
\multirow{3}{*}{\textbf{ID Dataset}} & \multirow{3}{*}{\textbf{Method}} & \multicolumn{8}{c}{\textbf{OOD Dataset}} & \multicolumn{2}{c}{\multirow{2}{*}{\textbf{Average}}} \\
& & \multicolumn{2}{c}{iNaturalist} & \multicolumn{2}{c}{SUN} & \multicolumn{2}{c}{Places} & \multicolumn{2}{c}{Texture} & \multicolumn{2}{c}{} \\            
\cmidrule(lr){3-4}\cmidrule(lr){5-6}\cmidrule(lr){7-8}\cmidrule(lr){9-10}\cmidrule(lr){11-12} 
& & \textbf{FPR95$\downarrow$} & \textbf{AUROC$\uparrow$} & \textbf{FPR95$\downarrow$} & \textbf{AUROC$\uparrow$} & \textbf{FPR95$\downarrow$} & \textbf{AUROC$\uparrow$} & \textbf{FPR95$\downarrow$} & \textbf{AUROC$\uparrow$} & \textbf{FPR95$\downarrow$} & \textbf{AUROC$\uparrow$} \\    
\midrule
\rowcolor{gray!22}
\cellcolor{white}
\multirow{6}{*}{\textbf{CUB-200-2011}} 
& CLIPN & 0.10 & 99.97 & 0.06 & 99.98 & 0.33 & 99.91 & 0.17 & 99.95 & 0.17 & 99.95 \\
& Energy & 0.46 & 99.89 & 0.03 & 99.99 & 0.30 & 99.92 & 0.02 & 100.00 &0.20 & 99.95 \\
& MaxLogit & 0.35 & 99.92 & 0.06 & 99.99& 0.35& 99.91 & 0.00 & 100.00 & 0.19& 99.95 \\
& MCM & 9.83 & 98.24 & 4.93 & 99.10 & 6.66 & 98.57 & 6.99 & 98.75 & 7.10 & 98.66 \\
& EOE~(Ours) & 0.06 & 99.98 & 0.03 & 100.00 & 0.37 & 99.88 & 0.01 & 100.00 & \textbf{0.12} & \textbf{99.96} \\
& Ground Truth & - & - &- & - & 0.29 & 99.93 & 0.00 & 99.99 & - & - \\
\midrule
\rowcolor{gray!22}
\cellcolor{white}
\multirow{6}{*}{\textbf{STANFORD-CARS}} 
& CLIPN & 0.00 & 99.99 & 0.02 & 99.99 & 0.13 & 99.96 & 0.02 & 99.99 & 0.04 & 99.98 \\
& Energy & 0.01 & 100.00 & 0.04 & 99.99 & 0.42 & 99.90 & 0.04 & 99.99 & 0.13 & 99.97 \\
& MaxLogit & 0.00 & 100.00 & 0.02 & 99.99 & 0.26 & 99.94 & 0.00 & 100.00 & 0.07 & 99.98 \\
& MCM &0.05&	99.77&	0.02&	99.95&	0.24&	99.89&	0.02&	99.96&	0.08 &	99.89 \\
& EOE~(Ours) & 0.00 & 100.00 & 0.01 & 100.00 & 0.06 & 99.99 & 0.00 & 100.00 & \textbf{0.02} & \textbf{100.00} \\
& Ground Truth & - & - & - & - & 0.07 & 99.99 & 0.00 & 100.00 & - & - \\
\midrule
\rowcolor{gray!22}
\cellcolor{white}
\multirow{6}{*}{\textbf{Food-101}} 
& CLIPN & 0.70 & 99.83 & 0.10 & 99.96 & 0.26 & 99.94 & 5.35 & 98.19 & 1.60 & 99.48 \\
& Energy & 0.92 & 99.75 & 0.18 & 99.92 & 0.54 & 99.86 & 12.43 & 96.55 & 3.52 & 99.02 \\
& MaxLogit & 0.56 & 99.86 & 0.09 & 99.95 & 0.49 & 99.88 & 8.33 & 97.44 & 2.37 & 99.28 \\
& MCM & 0.64& 	99.78& 	0.90& 	99.75& 	1.86& 	99.58& 	4.04& 	98.62& 	1.86& 	99.43\\
& EOE~(Ours) & 0.06 & 99.99 & 0.00 & 100.00 & 0.10 & 99.98 & 2.45 & 99.04 & \textbf{0.65 }& \textbf{99.76} \\
& Ground Truth & - & - & - & - & 0.02 & 99.99 & 0.59 & 99.83 & - & - \\
\midrule
\rowcolor{gray!22}
\cellcolor{white}
\multirow{6}{*}{\textbf{Oxford-IIIT Pet}} 
& CLIPN & 0.01 & 99.99 & 1.08 & 99.78 & 0.97 & 99.80 & 1.42 & 99.61 & 0.87 & 99.80 \\
& Energy & 0.06 & 99.97 & 0.05 & 99.98 & 0.23 & 99.94 & 0.35 & 99.88 & 0.17 & 99.94 \\
& MaxLogit & 0.02 & 99.98 & 0.05 & 99.97 & 0.20 & 99.94 & 0.27 & 99.91 & 0.14 & 99.95 \\
& MCM & 2.80&	99.38&	1.05&	99.73&	2.11&	99.56&	0.80&	99.81&	1.69&	99.62\\
& EOE~(Ours) & 0.00 & 100.00 & 0.01 & 99.99 & 0.15 & 99.96 & 0.12 & 99.97 & \textbf{0.07} & \textbf{99.98} \\
& Ground Truth &- & - & - & - & 0.08 & 99.98 & 0.09 & 99.98 & -& - \\
\midrule
\rowcolor{gray!22}
\cellcolor{white}
\multirow{6}{*}{\textbf{Average}} 
& CLIPN & 0.20 & 99.95 & 0.32 & 99.93 & 0.42 & 99.90 & 1.74 & 99.44 & 0.67 & 99.80 \\
& Energy & 0.36 & 99.90 & 0.07 & 99.97 & 0.37 & 99.91 & 3.21 & 99.10 & 1.01 & 99.72 \\
& MaxLogit & 0.23 & 99.94 & 0.06 & 99.98 & 0.33 & 99.92 & 2.15 & 99.34 & 0.69 & 99.79 \\
& MCM & 3.33 & 99.29 & 1.73 & 99.63 & 2.71 & 99.40 & 2.96 & 99.28 & 2.68 & 99.40\\
& EOE~(Ours) & 0.03 & 99.99& 0.01 & 100.00 & 0.17 & 99.95 & 0.64 & 99.75 & \textbf{0.21} & \textbf{99.92} \\
& Ground Truth & - & - & - &- & 0.12 & 99.97 & 0.17 & 99.95 & - & - \\
\bottomrule
\end{tabular}
\vspace{-.15in}
\end{table*}

\subsection{A New OOD Detection Score}
The purpose of the OOD detection score function is to better distinguish between ID and OOD score distributions. By having the outlier labels available, we can effectively enhance the score function's ability to differentiate between ID and OOD samples. We introduce a new score function $S_{\text{EOE}}$ based on potential outlier $\mathcal{Y}_\text{outlier}$ penalty to distinguish hard OOD samples effectively. First, the label-wise matching score $s_i(x)$ is expressed as:
\begin{equation}
    s_i(x) = \frac{\mathcal{I}(x) \cdot \mathcal{T}(t_i)}{\lVert \mathcal{I}(x)\rVert \cdot \lVert \mathcal{T}(t_i) \rVert};~~~t_i \in \mathcal{Y}_\text{id} \cup \mathcal{Y}_\text{outlier}.
    \label{label-wise matching score}
\end{equation}
Subsequently, the proposed score $S_{\text{EOE}}(\cdot)$ is defined, taking into account the outlier class labels, as follows:
\begin{multline}
    S_{\text{EOE}}(x;\mathcal{Y}_\text{id},\mathcal{Y}_\text{outlier},\mathcal{T},\mathcal{I}) = \underset{\scriptscriptstyle i \in [1, K]}\max \frac{e^{s_i(x)}}{\sum_{j=1}^{\scriptscriptstyle K+L} e^{s_j(x)}} \\- {\beta \cdot \underset{\scriptscriptstyle k \in (K, K+L]}\max} \frac{e^{s_k(x)}}{\sum_{j=1}^{\scriptscriptstyle K+L} e^{s_j(x)}},
\end{multline}
where $\beta$ is a hyperparameter, and we set $\beta=0.25$ in the main results. For more details on the design of $S_\text{EOE}$ and $\beta$, please refer to Appendix~\ref{app: score-detail}.
Based on $S_\text{EOE}$, the OOD detector $G(x; \mathcal{Y}_\text{id}, \mathcal{I}, \mathcal{T})$ can be viewed as the binary classification:
\begin{equation}
    G_{\lambda}(x; \mathcal{Y}_\text{id},\mathcal{Y}_\text{outlier},\mathcal{T},\mathcal{I})=\begin{cases} 
      \text{ID} & S_{\text{EOE}}(x)\ge \lambda \\
      \text{OOD} & S_{\text{EOE}}(x) < \lambda 
   \end{cases},
\end{equation}
where $\lambda$ is a selected threshold such that a high fraction of ID data (typically $95\%$) exceeds this value.  The zero-shot OOD detection procedure is shown in Algorithm~\ref{alg}.

\begin{algorithm}[t]
\begin{algorithmic}[1]
   \STATE\textbf{Input:} ID class labels $\mathcal{Y}_\text{id}$, test sample $x$, text encoder $\mathcal{T}$, image encoder $\mathcal{I}$, LLM, $\beta$, threshold $\lambda$;\\

    \textbf{Envisioning Stage}: 
        \STATE Given $\mathcal{Y}_\text{id}$, 
        $\mathcal{Y}_\text{outlier}=\text{LLM}(\text{prompt}(\mathcal{Y}_\text{id}))$;
   
    \textbf{Testing Stage}: 
        \STATE$K=\text{len}(\mathcal{Y}_\text{id}), L=\text{len}(\mathcal{Y}_\text{outlier})$;
        
        //~Compute label-wise matching score\vspace{-1pt}
        \STATE     \{$s_i(x) = \frac{\mathcal{I}(x) \cdot \mathcal{T}(t_i)}{\lVert \mathcal{I}(x)\rVert \cdot \lVert \mathcal{T}(t_i) \rVert}\}^{\scriptscriptstyle K+L}_{i=1};~~~t_i \in \mathcal{Y}_\text{id} \cup \mathcal{Y}_\text{outlier}$\vspace{+4pt};
        
        //~Compute OOD detection score\vspace{-1pt}
        \STATE \small$ S_{\text{EOE}}(x)\!=\!\underset{\scriptscriptstyle i \in [1,K]}\max \frac{e^{s_i(x)}}{\sum_{j=1}^{K+L} e^{s_j(x)}} 
        - {\beta\! \underset{\scriptscriptstyle k \in (K, K+L]}\max} \frac{e^{s_k(x)}}{\sum_{j=1}^{K+L} e^{s_j(x)}}$;
    \STATE \textbf{Output: } OOD detection decision $\mathbf{1}\{S_{\text{EOE}} \ge \lambda\}$.
\end{algorithmic}
\caption{Zero-shot OOD detection with envisioned outlier class labels}
\label{alg}
\end{algorithm}
\vspace{-.05in}
We summarize the advantages of our approach as follows:
\begin{enumerate}
\vspace{-.05in}
\item \textbf{OOD-Agnostic}: 
EOE does not rely on prior knowledge of unknown OOD data, making it particularly suitable and adaptable to open-world scenarios.
\vspace{-.05in}
\item \textbf{Zero-Shot}: A single pre-trained model efficiently serves various task-specific ID datasets without the need for individual training on each specific ID dataset. EOE can achieve superior OOD detection performance by merely knowing the ID class labels.
\item \textbf{Scalability and Generalizability}: 
In contrast to the existing zero-shot OOD detection method~\citep{esmaeilpour2022zero} that generates candidate OOD class labels, 
EOE can be easily applied to large-scale datasets like ImageNet-1K. Moreover, EOE exhibits generalizability across diverse tasks, including far, near, and fine-grained OOD detection.
\end{enumerate}

\begin{table*}[t]
\caption{Zero-shot \textbf{far} OOD detection results for ImageNet-1K as ID dataset. The \textbf{black bold} indicates the best performance. The \colorbox{gray!22}{gray} indicates that the comparative methods require training or an additional massive auxiliary dataset. Energy (FT) requires fine-tuning, while Energy is post-hoc.} %
\label{tab: far-ood-imagenet}
\centering
\fontsize{7}{8}\selectfont
\setlength\tabcolsep{6pt}
\begin{tabular}{ccccccccccc}
\toprule
\multirow{3}{*}{\textbf{Method}} & \multicolumn{8}{c}{\textbf{OOD Dataset}}                        & \multicolumn{2}{c}{\multirow{2}{*}{\textbf{Average} }} \\
                        & \multicolumn{2}{c}{iNaturalist} & \multicolumn{2}{c}{SUN} & \multicolumn{2}{c}{Places} & \multicolumn{2}{c}{Texture} & \multicolumn{2}{c}{}             \\
                        \cmidrule(lr){2-3}\cmidrule(lr){4-5}\cmidrule(lr){6-7}\cmidrule(lr){8-9}\cmidrule(lr){10-11}
                        & \textbf{FPR95$\downarrow$}         & \textbf{AUROC$\uparrow$}      & \textbf{FPR95$\downarrow$}           & \textbf{AUROC$\uparrow$}         & \textbf{FPR95$\downarrow$}          & \textbf{AUROC$\uparrow$}        & \textbf{FPR95$\downarrow$}         & \textbf{AUROC$\uparrow$}      & \textbf{FPR95$\downarrow$}          & \textbf{AUROC$\uparrow$}        \\  
                        \midrule
\rowcolor{gray!22}
MOS (BiT) & 9.28 & 98.15 & 40.63 & 92.01 & 49.54 & 89.06 & 60.43 & 81.23 & 39.97 & 90.11\\
\rowcolor{gray!22}
 Fort et al. & 15.07&	96.64&	54.12&	86.37&	57.99&	85.24&	53.32&	84.77&	45.12&	88.25\\
\rowcolor{gray!22}
Energy(FT) & 21.59&	95.99& 34.28&	93.15 &36.64&	91.82& 51.18&	88.09& 35.92&	92.26 \\ 
\rowcolor{gray!22}
MSP&40.89&	88.63&	65.81&	81.24&	67.90&	80.14&	64.96&	78.16&	59.89&	82.04\\
\rowcolor{gray!22}
CLIPN 
& 19.13	& 96.20	& 25.69 & 94.18 & 	32.14& 	92.26& 	44.60 & 88.93 & 30.39 & 	92.89\\
Energy
&81.08&85.09&79.02&84.24&75.08&83.38&93.65&65.56&82.21&79.57\\
MaxLogit
& 61.66&89.31&64.39&87.43&63.67&85.95&86.61&71.68&69.08&83.59\\
MCM
& 30.92	& 94.61	& 37.59& 	92.57& 	44.71& 	89.77& 	57.85& 	86.11& 	42.77& 	90.77\\
EOE~(Ours)
& 12.29 & 97.52 & 20.40 & 95.73 & 30.16 & 92.95 & 57.53 & 85.64 & \textbf{30.09} & \textbf{92.96}  \\
Ground Truth  & - & - & - & - & 13.24 & 96.96 & 24.29 & 95.04 & - & - \\
\bottomrule
\end{tabular}
\vspace{-.15in}
\end{table*}

\vspace{-.15in}
\section{Experiments}
\vspace{-.05in}
\label{exp}
\subsection{Setups}
\label{exp_setup}
\textbf{Far OOD Detection.} The ID datasets for far OOD detection encompass \text{CUB-200-2011}~\citep{wah2011caltech}, STANFORD-CARS~\citep{krause20133d}, Food-101~\citep{bossard14}, Oxford-IIIT Pet~\citep{parkhi2012cats} and ImageNet-1K~\citep{deng2009imagenet}. As for the OOD datasets, we use the large-scale OOD datasets iNaturalist~\citep{van2018inaturalist}, SUN~\citep{xiao2010sun}, Places~\citep{zhou2017places}, and Texture~\citep{cimpoi2014describing} curated by MOS~\citep{huang2021mos}. For more details about how each dataset is adapted to be ID-OOD, please refer to Appendix~\ref{app: dataset details}.

\textbf{Near OOD Detection.} We adopt ImageNet-10 and ImageNet-20 alternately as ID and OOD datasets, proposed by MCM~\citep{ming2022delving}, both of which are subsets extracted from ImageNet-1K. The ImageNet-10 dataset curated by MCM mimics the class distribution of CIFAR-10~\citep{krizhevsky2009learning}. The ImageNet-20 dataset consists of 20 classes semantically similar to ImageNet-10
(\textit{e.g.}, "\textit{horse}" (ID) \textit{vs.} "\textit{zebra}" (OOD)).

\textbf{Fine-grained OOD Detection.} We split CUB-200-2011, STANFORD-CARS, Food-101, and Oxford-IIIT Pet. Specifically, half of the classes from each dataset are randomly selected as ID data, while the remaining classes constitute OOD data. Importantly, there is no overlap between the above ID dataset and the corresponding OOD dataset.

\textbf{Evaluation Metrics.} We employ two widely-used metrics for evaluation: (1) FPR95, the false positive rate of OOD data when the true positive rate is at $95\%$ for ID data, where a lower value indicates better performance; (2) AUROC, the area under the receiver operating characteristic curve, with a higher value signifying superior performance. 
In addition, we report results in terms of AUPR in Appendix~\ref{app: aupr}.

\textbf{EOE Setups.}
We employ CLIP~\citep{radford2021learning} as the backbone of our framework. 
Unless otherwise specified, we adopt ViT-B/16 as the image encoder and masked self-attention Transformer~\citep{vaswani2017attention} as the text encoder in our experiments. The pre-trained weights of CLIP are sourced from the official weights provided by OpenAI. In addition, we adopt the GPT-3.5-turbo-16k model as the LLM for our research, with the temperature parameter setting to $0$. To reduce the potential impact of randomness, we instruct LLM to envision the outlier class three times on each ID dataset independently, and the final reported results are the average of these three experiments.

\textbf{Compared Methods.}
We compare our method with state-of-the-art OOD detection methods, including zero-shot and those requiring fine-tuning. For fair comparisons, all compared methods employ CLIP as their backbone, which is consistent with EOE. With respect to fine-tuning methods, we consider MSP~\citep{hendrycks17baseline}, Energy~\citep{liu2020energy}, MOS~\citep{huang2021mos}, and the method proposed by Fort et al.~\citep{fort2021exploring}. As for zero-shot methods, our comparisons are drawn towards MCM~\citep{ming2022delving} and CLIPN~\citep{wang2023clipn}. What's more, We implement post-hoc methods (Energy~\citep{liu2020energy} and MaxLogit~\citep{hendrycks2019scaling}) as additional baselines on CLIP backbone. It is worth noting that CLIPN relies on a large-scale auxiliary dataset~\citep{sharma2018conceptual} to additionally pre-train an encoder. Instead, our EOE does not require any such dataset.

\vspace{-.05in}
\subsection{Main Results}
\vspace{-.05in}
\label{exp_main}

\textbf{Far OOD Detection.}
Table~\ref{tab: far-ood} presents the comparison with the recent state-of-the-art zero-shot OOD detection method across four ID datasets: CUB-200-2011, STANFORD-CARS, Food-50, and Oxford-IIIT Pet. Ground Truth is the result of incorporating the actual OOD class labels, which are in fact unavailable. For each dataset, we guide the LLM to envision $500$ outlier classes, \textit{i.e.}, $L = 500$.  Clearly, EOE achieves superior results on these four ID datasets, with an average FPR95 of $0.21\%$ and AUROC of $99.92\%$. This indicates substantial improvement over the strong baseline. Notably, our EOE outperforms the strong baseline by $6.98\%$ when using CUB-200-2011 as the ID dataset.

\begin{table*}[t]
\centering
\caption{Zero-shot \textbf{near} OOD detection results. The \textbf{bold} indicates the best performance on each dataset, and the \colorbox{gray!22}{gray} indicates methods requiring an additional massive auxiliary dataset.}
\label{tab: near-ood}
\fontsize{7}{8}\selectfont
\setlength\tabcolsep{10pt}
\begin{tabular}{clcccccc}
\toprule

\multirow{2}{*}{\textbf{Method}} & \textbf{ID} & \multicolumn{2}{c}{ImageNet-10} & \multicolumn{2}{c}{ImageNet-20}& \multicolumn{2}{c}{\multirow{2}{*}{\textbf{Average}}} \\
 & \textbf{OOD}        &  \multicolumn{2}{c}{ImageNet-20} & \multicolumn{2}{c}{ImageNet-10} \\

\cmidrule(rl){3-4}
\cmidrule(rl){5-6}
\cmidrule(rl){7-8}
&&  \textbf{FPR95$\downarrow$} & \textbf{AUROC$\uparrow$}               
&  \textbf{FPR95$\downarrow$} & \textbf{AUROC$\uparrow$}   
&  \textbf{FPR95$\downarrow$} & \textbf{AUROC$\uparrow$}   \\
\midrule
\rowcolor{gray!22}
CLIPN    && 7.80 & 98.07  & 13.67 & 97.47 & 10.74 & 97.77    \\
Energy    && 10.30 & 97.94 & 16.40 & 97.37 & 13.35 & 97.66    \\
MaxLogit    && 9.70 & 98.09 & 14.00 & 97.81 & 11.85 & 97.95    \\
MCM    && 5.00 & 98.71  & 17.40 & 97.87 & 11.20 & 98.29    \\
EOE~(Ours)    && 4.20 & 99.09 & 13.93 & 98.10 & \textbf{9.07} & \textbf{98.59}\\
Ground Truth    && 0.20 &  99.80 & 0.20 & 99.93 & 0.20 & 99.87     \\
\bottomrule
\end{tabular}
\vspace{-.05in}
\end{table*}

\begin{table*}
\centering
\caption{Zero-shot \textbf{fine-grained} OOD detection results. The \textbf{bold} indicates the best performance on each dataset, and the \colorbox{gray!22}{gray} indicates methods requiring an additional massive auxiliary dataset.}
\label{tab: fine-grained-ood}
\fontsize{7}{8}\selectfont
\setlength\tabcolsep{6pt}
\begin{tabular}{clcccccccccccccc}
\toprule
\multirow{2}{*}{\textbf{Method}} & \textbf{ID} & \multicolumn{2}{c}{CUB-100} & \multicolumn{2}{c}{Stanford-Cars-98} & \multicolumn{2}{c}{Food-50} & \multicolumn{2}{c}{Oxford-Pet-18} & \multicolumn{2}{c}{\multirow{2}{*}{\textbf{Average}}}\\
 & \textbf{OOD}        &  \multicolumn{2}{c}{CUB-100} & \multicolumn{2}{c}{Stanford-Cars-98} & \multicolumn{2}{c}{Food-51} & \multicolumn{2}{c}{Oxford-Pet-19}  \\

\cmidrule(rl){3-4}
\cmidrule(rl){5-6}
\cmidrule(rl){7-8}
\cmidrule(rl){9-10}
\cmidrule(rl){11-12}
\cmidrule(rl){13-14}
&&  \textbf{FPR95$\downarrow$} & \textbf{AUROC$\uparrow$}               
&  \textbf{FPR95$\downarrow$} & \textbf{AUROC$\uparrow$}
&  \textbf{FPR95$\downarrow$} & \textbf{AUROC$\uparrow$}
&  \textbf{FPR95$\downarrow$} & \textbf{AUROC$\uparrow$}
&  \textbf{FPR95$\downarrow$} & \textbf{AUROC$\uparrow$}   \\
\midrule
\rowcolor{gray!22}
CLIPN   && 73.54        & 74.65    
      & 53.33        & 82.25
      & 43.33       & 88.89
      & 53.90       & 86.92
      & 56.05             & 83.18 \\ 
Energy  && 76.13 & 72.11 & 73.78 & 73.82 & 44.95 & 89.97 & 68.51 & 88.34 & 65.84 & 81.06\\
MaxLogit   && 76.89 & 73.00 & 72.18 & 74.80 & 41.73 & 90.79 & 65.66 & 88.49 & 64.11 & 81.77\\
MCM   &&83.58 & 67.51 & 83.99 & 68.71 & 43.38 & 91.75 & 63.92 & 84.88 & 68.72 & 78.21\\
EOE~(Ours)   
&& 74.74 & 73.41 & 76.83 & 71.60 & 37.95 & 91.96 & 52.55 & 90.33 & \textbf{60.52} & \textbf{81.82}\\
Ground Truth   && 61.23   & 81.42
      & 58.31           & 83.71 
       & 11.34            & 97.79
      & 29.17            & 95.58 
      & 40.01          & 89.63 \\

\bottomrule
\end{tabular}
\vspace{-.15in}
\end{table*}
We then conduct experiments on the large-scale dataset (ImageNet-1K) for far OOD detection. Results are reported in Table~\ref{tab: far-ood-imagenet}. We adopt the results of fine-tuning methods reported by MCM. EOE is comparable to fine-tuning methods and surpasses MCM.
Although CLIPN uses an additional text encoder and large-scale datasets for training the ``no'' prompt, our EOE still outperforms CLIPN. Furthermore, CLIPN utilizes an ensemble strategy for the textual inputs, in which the ensemble and learnable textual inputs are effective in enhancing performance~\citep{zhou2022learning, zhou2022conditional}.

\textbf{Near OOD Detection.}
The results for near OOD detection are presented in Table~\ref{tab: near-ood}. For each ID class label, EOE instructs the LLM to return three outlier class labels. 
EOE outperforms the strong baseline MCM by achieving improvements of $2.13\%$ in average FPR95 and $0.30\%$ in AUROC. And our EOE also surpasses Energy and MaxLogit scores. Compared to CLIPN, which uses extra large datasets for re-training, our EOE clearly outperforms it. 
Note that when using ImageNet-20 as the ID dataset, we filtered out candidate class names with a cosine similarity above $85\%$ to the ID class names to prevent excessive ID samples from being classified as OOD candidates. We also report results on large-scale benchmarks organized by OpenOOD in Table~\ref{app-tab: near-ood}, specifically ImageNet-1K \textit{vs.} SSB-hard~\citep{vaze2021open} and NINCO~\citep{bitterwolf2023or}.

\textbf{Fine-grained OOD Detection.}
Table~\ref{tab: fine-grained-ood} shows the performance of fine-grained OOD detection. EOE guides the LLM to generate 500 outlier class labels for each ID dataset. Compared to MCM, our EOE increases the average OOD performance by $3.59\%$ in FPR95. Despite the unfair comparison, our EOE still outperforms CLIPN on the Food-50 and Oxford-Pet-18 datasets in terms of FPR95.

\vspace{-.05in}
\subsection{Ablation Study}
\begin{figure*}[!t]
\begin{center}
\includegraphics[width=\textwidth]{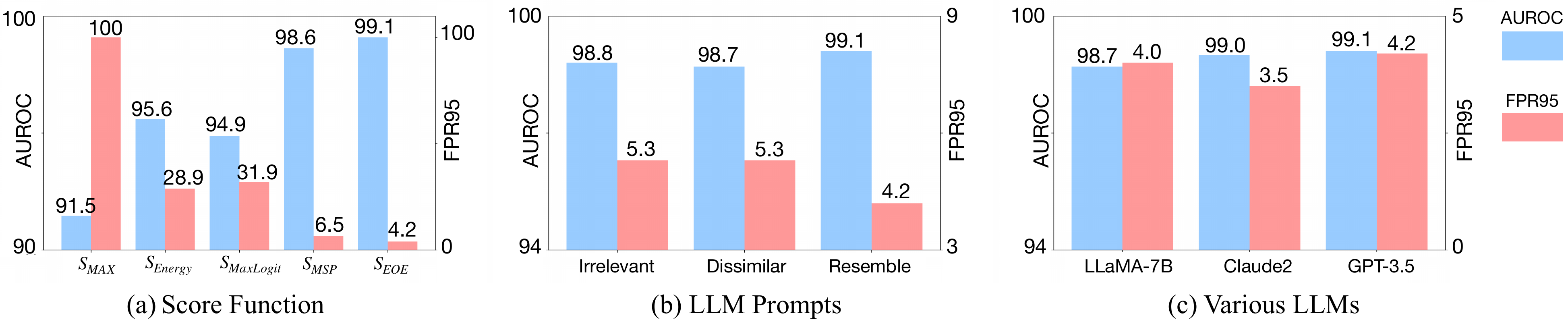}
\end{center}
\vspace{-.05in}
\caption{Ablation study on (a) score function, (b) LLM prompts, and (c) various LLMs. ID dataset: ImageNet-10; OOD dataset: ImageNet-20.}
\label{fig: ablation1}
\vspace{-.05in}
\end{figure*}

\vspace{-.05in}
\label{exp_ablation}
\textbf{Score Functions.}
To demonstrate the superiority of the proposed OOD detection score $S_{\text{EOE}}$, we compare it with the other score functions: $S_{\text{MAX}}, S_{\text{MSP}}, S_{\text{Energy}}$ and $S_{\text{MaxLogit}}$. These score functions are all designed based on potential outlier class labels.
The comparison of these score functions is shown in Figure~\ref{fig: ablation1}~(a). 
Please refer to Appendix~\ref{app: score function} for the specific forms and results on more datasets.
Results show that our $S_\text{EOE}$ achieves the best OOD performance. This verifies the superiority and importance of the proposed OOD detection score.

\textbf{LLM Prompts.} To investigate the effectiveness of the visual similarity rule, we design two additional types of LLM prompts, one termed \textit{visually irrelevant} and the other \textit{visually dissimilar}. The rest of the LLM prompt content remains unchanged. Specifically, the \textit{`irrelevant'} LLM prompt instructs the LLM to generate arbitrary outlier class labels for the ID class without adhering to a \textit{visually resemblance} constraint. Conversely, the \textit{`dissimilar'} LLM prompt asks the LLM to envision outlier class labels for the ID class under a \textit{visually dissimilar} constraint.
As shown in Figure~\ref{fig: ablation1}~(b), without the \textit{visually resemble} constraint proposed in our EOE, the OOD performance degrades on both FPR95 and AUROC metrics, indicating the significance of the proposed constraint.
For detailed LLM prompts 
and results on more datasets, please refer to Appendix~\ref{app: ablation-llm-prompts}.

\textbf{Various LLMs.} We conduct experiments with various LLMs to provide a more comprehensive understanding of EOE’s effectiveness. Specifically, we use LLaMA2-7B~\citep{touvron2023llama} and Claude 2~\citep{anthropic2023claude2} to envision outlier class labels. The results on ImageNet-10 (ID) are shown in Figure~\ref{fig: ablation1}~(c). EOE achieves better results than the baseline MCM across different LLMs. 
Moreover, LLaMA2-7B and Claude 2 outperform GPT-3.5-turbo-16k in terms of FPR95. These results demonstrate the generalizability of our method. For further results on more datasets related to LLM prompts, refer to Appendix~\ref{app:ablaton-llms}.

\begin{figure*}[!t]
\begin{center}
\includegraphics[width=\textwidth]{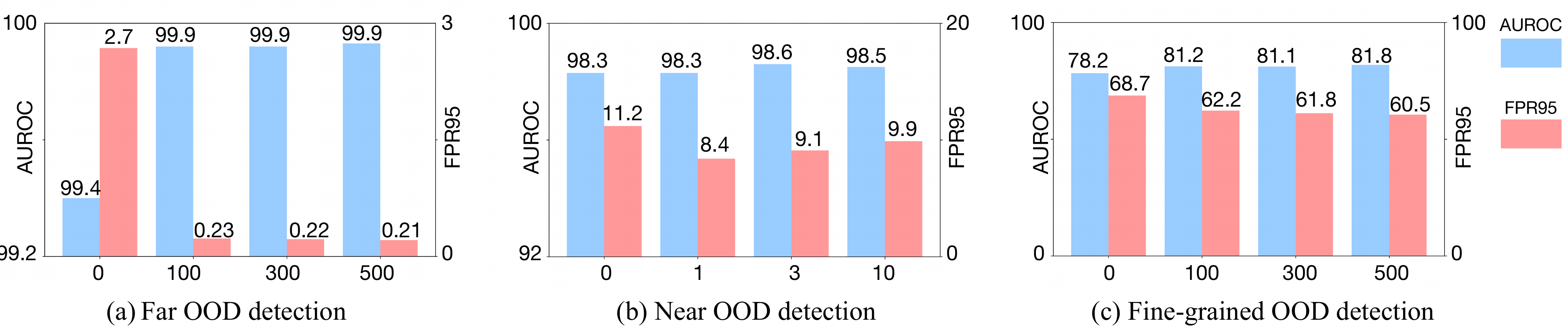}
\end{center}
\vspace{-.05in}
  \caption{Evaluation on the number of outlier class labels. When the number of outlier class labels is zero, the method reduces to the baseline MCM.
  }
  \label{fig: ablation2}
\vspace{-.05in}
\end{figure*}

\textbf{Number of Outlier Class Labels.}
We conduct experiments to investigate the impact of the number of outlier class labels, \textit{i.e.}, $L$, generated by LLM. We instruct the LLM to return $100$, $300$, and $500$ outlier class labels for the far and fine-grained OOD detection for each ID dataset. For the near OOD detection, we ask the LLM to return outlier class labels in quantities of $1$, $3$, and $10$ for each ID class label.
Figure~\ref{fig: ablation2} presents the respective average metrics for varying numbers of outlier class labels across the three tasks. EOE consistently outperforms the baseline MCM and maintains relatively stable performance across different values of $L$.
What's more, we conduct experiments on $\beta \in \{0, 0.25, 0.50, 0.75, 1\}$ with different outlier class number $L$. For detailed results, please refer to Appendix~\ref{app: score-detail}.

\vspace{-.05in}
\subsection{Further Analysis}
\textbf{More Experimental Results}. We perform experiments with CIFAR-10/CIFAR-100~\citep{krizhevsky2009learning} benchmarks to support our method further. Please refer to Appendix~\ref{app: cifar benchmark} for the results. We also explore the performance of using different CLIP vision encoders and different VLMs in Appendix~\ref{app: differ-vlm}. 
When using different backbones, our EOE’s performance is significantly better than the other methods. For example, when employing GroupViT~\citep{xu2022groupvit} as the backbone, our EOE method outperforms MCM significantly, achieving a $23.22\%$ improvement in FPR95 when using ImageNet-1K as the ID dataset. These results indicate that our EOE is more generalizable to different VLMs. To explore the robustness of EOE, we conduct experiments on more datasets, including ImageNet-Sketch~\citep{wang2019learning} and ImageNet-C~\citep{hendrycks2019benchmarking}. Compared to MCM, EOE demonstrates a $27.19\%$ improvement in terms of FPR95 when using ImageNet-Sketch as the ID dataset.
The detailed results are in Appendix~\ref{app: robustness}. What's more, We also perform prompt ensembling strategy for text input using different CLIP backbones. Interestingly, only our EOE achieves gains with the prompt ensembling strategy across different vision encoders. See Table~\ref{tab: prompt-engineering} in Appendix~\ref{app: prompt-engineering} for details.



\begin{figure}[t]
\begin{center}
\includegraphics[width=\columnwidth]{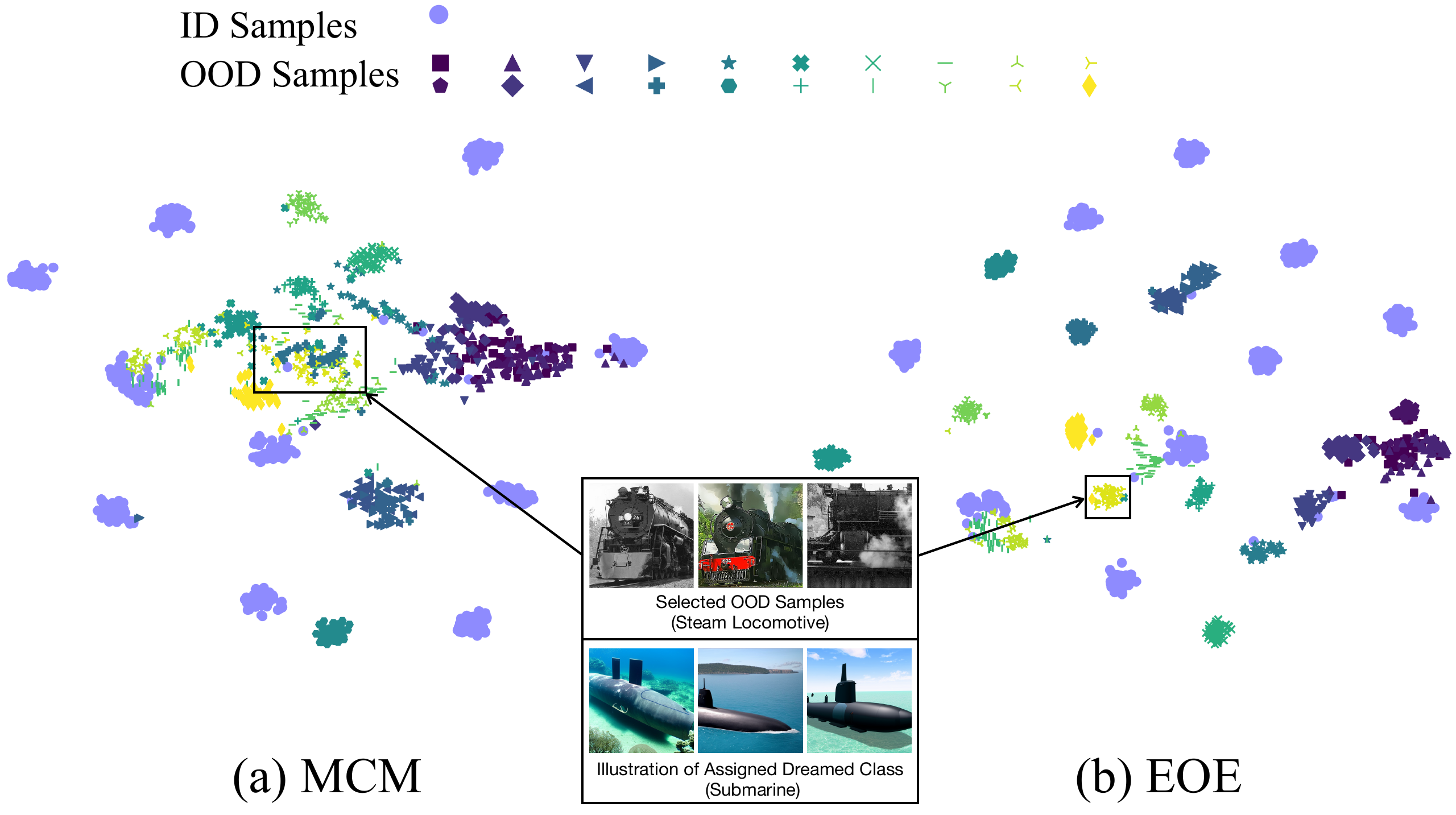}
\end{center}
  \caption{T-SNE visualizations obtained by the classifier output. ID set: ImageNet-10; OOD set: ImageNet-20. We use distinct colors to represent different OOD classes. The illustrated envisioned OOD name is the class assigned with the corresponding cluster, and its examples are generated by Stable Diffusion~\citep{rombach2022high}. 
  }
  \label{fig: visual_ImageNet10}
  \vspace{-0.3in}
\end{figure}

\textbf{Understanding of EOE.} In fact, it is impractical for the generated outlier class labels to have a high probability of hitting the ground truth (GT) OOD class. This is because the OOD data encountered are diverse and unpredictable in the actual deployment of the model. However, it is for this reason that our EOE contributes to the OOD detection community. Based on the visual similarity rule, EOE asks the LLM to generate potential outlier class labels. \textit{Even without hitting the GT OOD, these potential outlier classes can still enhance our performance in OOD detection.} To better understand this argument, we display the visualizations derived from the softmax output for the label-wise matching score via T-SNE~\citep{van2014accelerating}. Results compared between our EOE and the baseline MCM are shown in Figure~\ref{fig: visual_ImageNet10}. Based on the ID class labels of ImageNet-10, LLM generates the potential outlier label `submarine' anchored on the visual similarity rule. When encountering the OOD class `\textit{Steam Locomotive}'(class in ImageNet-20), `\textit{Steam Locomotive}' exhibits the highest similarity with `\textit{submarine}' among $\mathcal{Y}_\text{id}$ and $\mathcal{Y}_\text{outlier}$. Therefore, EOE will cluster it as `\textit{submarine}', thus detecting it as an OOD class. Without the potential outlier class labels, we can find that MCM tends to cluster all the OOD class labels together. This could lead to identifying hard OOD samples as ID classes. 
We also present the visualization when using ImageNet-20 as the ID dataset, as shown in Figure~\ref{fig: visual_ImageNet20}. EOE will cluster actual OOD class `\textit{antelope}' into potential outlier class `\textit{Giraffe}'. In summary, within our EOE framework, 1) OOD samples belonging to the same class tend to be clustered together, and 2) samples in the same group are classified into the envisioned outlier class that is visually similar to them (Steam Locomotive \textit{vs} Submarine). These observations indicate that our EOE can enhance OOD detection without hitting actual OOD classes and is more semantically explainable.

\textbf{Limitation I.} Our method assumes that the LLM understands the ID categories. However, when the ID categories fall outside the expertise of the LLM, it may not provide appropriate candidate OOD categories. This situation can arise when the ID categories require extreme expertise or when the ID categories are entirely new, and the LLM has never encountered them before. To handle this issue, we can adopt a preprocessing strategy. Specifically, before asking the LLM to provide candidate OOD categories, it should be queried whether it understands the given ID categories. If the LLM demonstrates understanding, our algorithm can proceed as designed; otherwise, it should fall back to the MCM approach. Furthermore, thanks to the generality of linguistic descriptions and the LLM's remarkable capacity for in-context learning, we can describe ID categories that are beyond the LLM's expert knowledge. By teaching the LLM about the visual features associated with these categories, we can then prompt the LLM to generate appropriate candidate OOD categories. This approach can also somewhat mitigate the impact of this failure case.


\textbf{Limitation II.} Another minor limitation of our approach is the use of different prompts, although our experiments demonstrate that the three proposed prompts are universal and can be applied across various datasets for each task. This implicitly assumes that we are aware of the specific type of OOD task at hand (far, near, or fine-grained).
Despite this, the restriction imposed by this assumption is minimal and does not undermine the core principles of OOD detection.
Additionally, we conducted experiments without this assumption, where we used the far OOD detection prompt~(Figure~\ref{fig: far ood LLM prompts}) in all tasks. The results are shown in Table~\ref{app: limitation-tab}, our EOE method still surpasses MCM in most cases. 
In this paper, we hope to propose a new perspective using LLMs to explore potential OOD candidates for solving the zero-shot OOD detection task.

\vspace{-.05in}
\section{Related Works}
\vspace{-.05in}
\textbf{OOD Detection.} Methods adopted for previous OOD detection can be broadly categorized into classification-based~\citep{hendrycks17baseline}, density-based~\citep{ren2019likelihood, xiao2020likelihood}, and reconstruction-based~\citep{denouden2018improving, zhou2022rethinking, liu2023unsupervised}. Classification-based methods leverage a well-trained ID classifier and formulate a score function to recognize OOD samples. The score function can be formulated from input~\citep{liang2017enhancing}, hidden~\citep{sun2022out, lee2018simple, sun2022dice, zhu2023unleashing, wang2022watermarking}, output~\citep{hendrycks17baseline, hendrycks2019scaling, liu2020energy}, and gradient space~\citep{huang2021importance}. 
When the label space of the test ID data differs from the training data, the model needs to be re-trained from scratch or fine-tuned in previous OOD detection scenarios, which require significant computational overhead. 

\textbf{Zero-Shot OOD Detection.}
Owing to the powerful capabilities of the VLMs, zero-shot OOD detection methods have shown promising results. 
Typically, the similarity between the feature representations of input images and the textual input is measured to identify OOD samples.
MCM~\citep{ming2022delving} is entirely dependent on closed-set ID class labels and does not effectively harness the potent abilities of CLIP for handling open-world inputs.
Although ZOC~\citep{esmaeilpour2022zero} and CLIPN~\citep{wang2023clipn} take into account the open-world setting, ZOC requires an additional auxiliary dataset to train a text-based image description generator to generate candidate unknown class names for each input sample. This makes ZOC ill-suited for handling large-scale datasets. Similarly, CLIPN requires an extra auxiliary dataset for training the text encoder.
By contrast, EOE not only considers the open-world scenario but also foregoes the need for any auxiliary datasets for extra training and can be easily scaled to large datasets.

\textbf{LLM for Visual Classification.}
Drawing upon the expert knowledge embedded in LLMs has emerged as a novel trend in vision tasks and remains under-explored. \cite{menon2022visual, maniparambil2023enhancing} employ the expertise within LLMs to extract information contained in ID class labels, thereby enhancing the performance of image classification~(classification by description). Differing from this, EOE leverages the expert knowledge in LLMs to envision outlier OOD classes based on the visual similarity rule, harnessing the capabilities of VLMs more effectively, thus improving the performance of identifying OOD samples.

\vspace{-0.05in}
\section{Conclusion}
\vspace{-0.05in}
In this paper, we propose a new paradigm for zero-shot OOD detection, called EOE, by harnessing the expert knowledge embedded in LLMs to envision outlier exposure without relying on actual or auxiliary OOD data. Based on the designed visual similarity rule, the proposed three LLM prompts are applicable across various datasets for far, near, and fine-grained OOD detection tasks. We introduce a new score function based on potential outlier penalty, enabling us to recognize OOD samples effectively. 
Extensive experiments show that EOE achieves new state-of-the-art performance 
and can be effectively scaled to the ImageNet-1K dataset.
We hope this work could open a new door in future research in the OOD detection field.

\vspace{-.05in}
\section*{Acknowledgements}
\vspace{-.05in}
CTC, ZKZ, and BH were supported by the NSFC General Program No. 62376235, Guangdong Basic and Applied Basic Research Foundation Nos.~2024A1515012399 and 2022A1515011652, HKBU Faculty Niche Research Areas No.~RC-FNRA-IG/22-23/SCI/04, ByteDance Faculty Research Award, and HKBU CSD Departmental Incentive Grant. YL was partially supported by the National Science Foundation (NSF) under grants IIS-2007951 and IIS-2143895. TLL was partially supported by the following Australian Research Council projects: FT220100318, DP220102121, LP220100527, LP220200949, and IC190100031.

\vspace{-0.05in}
\section*{Impact Statement}
\vspace{-0.05in}
This paper presents work whose goal is to advance the field of machine learning. There are many potential societal consequences of our work, none of which we feel must be specifically highlighted here.

\bibliography{draft}

\begin{thebibliography}{66}
\providecommand{\natexlab}[1]{#1}
\providecommand{\url}[1]{\texttt{#1}}
\expandafter\ifx\csname urlstyle\endcsname\relax
  \providecommand{\doi}[1]{doi: #1}\else
  \providecommand{\doi}{doi: \begingroup \urlstyle{rm}\Url}\fi

\bibitem[Anthropic(2023)]{anthropic2023claude2}
Anthropic.
\newblock Claude 2.
\newblock Anthropic Blog, July 11 2023.
\newblock URL \url{https://www.anthropic.com/index/claude-2}.

\bibitem[Anthropic(2024)]{anthropic2024claude}
Anthropic, A.
\newblock The claude 3 model family: Opus, sonnet, haiku.
\newblock \emph{Claude-3 Model Card}, 2024.

\bibitem[Bitterwolf et~al.(2023)Bitterwolf, Mueller, and Hein]{bitterwolf2023or}
Bitterwolf, J., Mueller, M., and Hein, M.
\newblock In or out? fixing imagenet out-of-distribution detection evaluation.
\newblock \emph{arXiv preprint arXiv:2306.00826}, 2023.

\bibitem[Bossard et~al.(2014)Bossard, Guillaumin, and Van~Gool]{bossard14}
Bossard, L., Guillaumin, M., and Van~Gool, L.
\newblock Food-101 -- mining discriminative components with random forests.
\newblock In \emph{ECCV}, 2014.

\bibitem[Brown et~al.(2020)Brown, Mann, Ryder, Subbiah, Kaplan, Dhariwal, Neelakantan, Shyam, Sastry, Askell, et~al.]{brown2020language}
Brown, T., Mann, B., Ryder, N., Subbiah, M., Kaplan, J.~D., Dhariwal, P., Neelakantan, A., Shyam, P., Sastry, G., Askell, A., et~al.
\newblock Language models are few-shot learners.
\newblock In \emph{NeurIPS}, 2020.

\bibitem[Chen et~al.(2022)Chen, Liu, Zhang, Ye, Yang, and Wu]{chen2022altclip}
Chen, Z., Liu, G., Zhang, B.-W., Ye, F., Yang, Q., and Wu, L.
\newblock Altclip: Altering the language encoder in clip for extended language capabilities.
\newblock \emph{arXiv preprint arXiv:2211.06679}, 2022.

\bibitem[Chowdhery et~al.(2022)Chowdhery, Narang, Devlin, Bosma, Mishra, Roberts, Barham, Chung, Sutton, Gehrmann, et~al.]{chowdhery2022palm}
Chowdhery, A., Narang, S., Devlin, J., Bosma, M., Mishra, G., Roberts, A., Barham, P., Chung, H.~W., Sutton, C., Gehrmann, S., et~al.
\newblock Palm: Scaling language modeling with pathways.
\newblock \emph{arXiv preprint arXiv:2204.02311}, 2022.

\bibitem[Cimpoi et~al.(2014)Cimpoi, Maji, Kokkinos, Mohamed, and Vedaldi]{cimpoi2014describing}
Cimpoi, M., Maji, S., Kokkinos, I., Mohamed, S., and Vedaldi, A.
\newblock Describing textures in the wild.
\newblock In \emph{CVPR}, 2014.

\bibitem[Deng et~al.(2009)Deng, Dong, Socher, Li, Li, and Fei-Fei]{deng2009imagenet}
Deng, J., Dong, W., Socher, R., Li, L.-J., Li, K., and Fei-Fei, L.
\newblock Imagenet: A large-scale hierarchical image database.
\newblock In \emph{CVPR}, 2009.

\bibitem[Denouden et~al.(2018)Denouden, Salay, Czarnecki, Abdelzad, Phan, and Vernekar]{denouden2018improving}
Denouden, T., Salay, R., Czarnecki, K., Abdelzad, V., Phan, B., and Vernekar, S.
\newblock Improving reconstruction autoencoder out-of-distribution detection with mahalanobis distance.
\newblock \emph{arXiv preprint arXiv:1812.02765}, 2018.

\bibitem[Dosovitskiy et~al.(2021)Dosovitskiy, Beyer, Kolesnikov, Weissenborn, Zhai, Unterthiner, Dehghani, Minderer, Heigold, Gelly, Uszkoreit, and Houlsby]{dosovitskiy2020vit}
Dosovitskiy, A., Beyer, L., Kolesnikov, A., Weissenborn, D., Zhai, X., Unterthiner, T., Dehghani, M., Minderer, M., Heigold, G., Gelly, S., Uszkoreit, J., and Houlsby, N.
\newblock An image is worth 16x16 words: Transformers for image recognition at scale.
\newblock In \emph{ICLR}, 2021.

\bibitem[Esmaeilpour et~al.(2022)Esmaeilpour, Liu, Robertson, and Shu]{esmaeilpour2022zero}
Esmaeilpour, S., Liu, B., Robertson, E., and Shu, L.
\newblock Zero-shot out-of-distribution detection based on the pre-trained model clip.
\newblock In \emph{AAAI}, 2022.

\bibitem[Fort et~al.(2021)Fort, Ren, and Lakshminarayanan]{fort2021exploring}
Fort, S., Ren, J., and Lakshminarayanan, B.
\newblock Exploring the limits of out-of-distribution detection.
\newblock In \emph{NeurIPS}, 2021.

\bibitem[He et~al.(2016)He, Zhang, Ren, and Sun]{he2016deep}
He, K., Zhang, X., Ren, S., and Sun, J.
\newblock Deep residual learning for image recognition.
\newblock In \emph{CVPR}, 2016.

\bibitem[Hendrycks \& Dietterich(2019)Hendrycks and Dietterich]{hendrycks2019benchmarking}
Hendrycks, D. and Dietterich, T.
\newblock Benchmarking neural network robustness to common corruptions and perturbations.
\newblock \emph{arXiv preprint arXiv:1903.12261}, 2019.

\bibitem[Hendrycks \& Gimpel(2017)Hendrycks and Gimpel]{hendrycks17baseline}
Hendrycks, D. and Gimpel, K.
\newblock A baseline for detecting misclassified and out-of-distribution examples in neural networks.
\newblock In \emph{ICLR}, 2017.

\bibitem[Hendrycks et~al.(2019{\natexlab{a}})Hendrycks, Basart, Mazeika, Zou, Kwon, Mostajabi, Steinhardt, and Song]{hendrycks2019scaling}
Hendrycks, D., Basart, S., Mazeika, M., Zou, A., Kwon, J., Mostajabi, M., Steinhardt, J., and Song, D.
\newblock Scaling out-of-distribution detection for real-world settings.
\newblock \emph{arXiv preprint arXiv:1911.11132}, 2019{\natexlab{a}}.

\bibitem[Hendrycks et~al.(2019{\natexlab{b}})Hendrycks, Mazeika, and Dietterich]{hendrycks2019oe}
Hendrycks, D., Mazeika, M., and Dietterich, T.
\newblock Deep anomaly detection with outlier exposure.
\newblock In \emph{ICLR}, 2019{\natexlab{b}}.

\bibitem[Huang \& Li(2021)Huang and Li]{huang2021mos}
Huang, R. and Li, Y.
\newblock Mos: Towards scaling out-of-distribution detection for large semantic space.
\newblock In \emph{CVPR}, 2021.

\bibitem[Huang et~al.(2021)Huang, Geng, and Li]{huang2021importance}
Huang, R., Geng, A., and Li, Y.
\newblock On the importance of gradients for detecting distributional shifts in the wild.
\newblock In \emph{NeurIPS}, 2021.

\bibitem[Jia et~al.(2021)Jia, Yang, Xia, Chen, Parekh, Pham, Le, Sung, Li, and Duerig]{jia2021scaling}
Jia, C., Yang, Y., Xia, Y., Chen, Y.-T., Parekh, Z., Pham, H., Le, Q., Sung, Y.-H., Li, Z., and Duerig, T.
\newblock Scaling up visual and vision-language representation learning with noisy text supervision.
\newblock In \emph{ICML}, 2021.

\bibitem[Jiang et~al.(2024)Jiang, Sablayrolles, Roux, Mensch, Savary, Bamford, Chaplot, Casas, Hanna, Bressand, et~al.]{jiang2024mixtral}
Jiang, A.~Q., Sablayrolles, A., Roux, A., Mensch, A., Savary, B., Bamford, C., Chaplot, D.~S., Casas, D. d.~l., Hanna, E.~B., Bressand, F., et~al.
\newblock Mixtral of experts.
\newblock \emph{arXiv preprint arXiv:2401.04088}, 2024.

\bibitem[Krause et~al.(2013)Krause, Stark, Deng, and Fei-Fei]{krause20133d}
Krause, J., Stark, M., Deng, J., and Fei-Fei, L.
\newblock 3d object representations for fine-grained categorization.
\newblock In \emph{ICCV Workshops}, 2013.

\bibitem[Krizhevsky et~al.(2009)Krizhevsky, Hinton, et~al.]{krizhevsky2009learning}
Krizhevsky, A., Hinton, G., et~al.
\newblock Learning multiple layers of features from tiny images.
\newblock 2009.

\bibitem[Lee et~al.(2018)Lee, Lee, Lee, and Shin]{lee2018simple}
Lee, K., Lee, K., Lee, H., and Shin, J.
\newblock A simple unified framework for detecting out-of-distribution samples and adversarial attacks.
\newblock In \emph{NeurIPS}, 2018.

\bibitem[Liang et~al.(2017)Liang, Li, and Srikant]{liang2017enhancing}
Liang, S., Li, Y., and Srikant, R.
\newblock Enhancing the reliability of out-of-distribution image detection in neural networks.
\newblock \emph{arXiv preprint arXiv:1706.02690}, 2017.

\bibitem[Liu et~al.(2020)Liu, Wang, Owens, and Li]{liu2020energy}
Liu, W., Wang, X., Owens, J., and Li, Y.
\newblock Energy-based out-of-distribution detection.
\newblock In \emph{NeurIPS}, 2020.

\bibitem[Liu et~al.(2023)Liu, Zhou, Wang, and Weinberger]{liu2023unsupervised}
Liu, Z., Zhou, J.~P., Wang, Y., and Weinberger, K.~Q.
\newblock Unsupervised out-of-distribution detection with diffusion inpainting.
\newblock \emph{arXiv preprint arXiv:2302.10326}, 2023.

\bibitem[Maniparambil et~al.(2023)Maniparambil, Vorster, Molloy, Murphy, McGuinness, and O'Connor]{maniparambil2023enhancing}
Maniparambil, M., Vorster, C., Molloy, D., Murphy, N., McGuinness, K., and O'Connor, N.~E.
\newblock Enhancing clip with gpt-4: Harnessing visual descriptions as prompts.
\newblock \emph{arXiv preprint arXiv:2307.11661}, 2023.

\bibitem[Menon \& Vondrick(2023)Menon and Vondrick]{menon2022visual}
Menon, S. and Vondrick, C.
\newblock Visual classification via description from large language models.
\newblock In \emph{ICLR}, 2023.

\bibitem[Ming et~al.(2022)Ming, Cai, Gu, Sun, Li, and Li]{ming2022delving}
Ming, Y., Cai, Z., Gu, J., Sun, Y., Li, W., and Li, Y.
\newblock Delving into out-of-distribution detection with vision-language representations.
\newblock In \emph{NeurIPS}, 2022.

\bibitem[Netzer et~al.(2011)Netzer, Wang, Coates, Bissacco, Wu, and Ng]{netzer2011reading}
Netzer, Y., Wang, T., Coates, A., Bissacco, A., Wu, B., and Ng, A.~Y.
\newblock Reading digits in natural images with unsupervised feature learning.
\newblock In \emph{NeurIPS Workshop}, 2011.

\bibitem[OpenAI(2023)]{gpt4}
OpenAI.
\newblock Gpt-4 technical report.
\newblock \emph{arXiv preprint arXiv:2303.08774}, 2023.

\bibitem[Parkhi et~al.(2012)Parkhi, Vedaldi, Zisserman, and Jawahar]{parkhi2012cats}
Parkhi, O.~M., Vedaldi, A., Zisserman, A., and Jawahar, C.
\newblock Cats and dogs.
\newblock pp.\  3498--3505, 2012.

\bibitem[Paszke et~al.(2019)Paszke, Gross, Massa, Lerer, Bradbury, Chanan, Killeen, Lin, Gimelshein, Antiga, et~al.]{paszke2019pytorch}
Paszke, A., Gross, S., Massa, F., Lerer, A., Bradbury, J., Chanan, G., Killeen, T., Lin, Z., Gimelshein, N., Antiga, L., et~al.
\newblock Pytorch: An imperative style, high-performance deep learning library.
\newblock In \emph{NeurIPS}, 2019.

\bibitem[Radford et~al.(2021)Radford, Kim, Hallacy, Ramesh, Goh, Agarwal, Sastry, Askell, Mishkin, Clark, et~al.]{radford2021learning}
Radford, A., Kim, J.~W., Hallacy, C., Ramesh, A., Goh, G., Agarwal, S., Sastry, G., Askell, A., Mishkin, P., Clark, J., et~al.
\newblock Learning transferable visual models from natural language supervision.
\newblock In \emph{ICML}, 2021.

\bibitem[Reid et~al.(2024)Reid, Savinov, Teplyashin, Lepikhin, Lillicrap, Alayrac, Soricut, Lazaridou, Firat, Schrittwieser, et~al.]{reid2024gemini}
Reid, M., Savinov, N., Teplyashin, D., Lepikhin, D., Lillicrap, T., Alayrac, J.-b., Soricut, R., Lazaridou, A., Firat, O., Schrittwieser, J., et~al.
\newblock Gemini 1.5: Unlocking multimodal understanding across millions of tokens of context.
\newblock \emph{arXiv preprint arXiv:2403.05530}, 2024.

\bibitem[Ren et~al.(2019)Ren, Liu, Fertig, Snoek, Poplin, Depristo, Dillon, and Lakshminarayanan]{ren2019likelihood}
Ren, J., Liu, P.~J., Fertig, E., Snoek, J., Poplin, R., Depristo, M., Dillon, J., and Lakshminarayanan, B.
\newblock Likelihood ratios for out-of-distribution detection.
\newblock In \emph{NeurIPS}, 2019.

\bibitem[Rombach et~al.(2022)Rombach, Blattmann, Lorenz, Esser, and Ommer]{rombach2022high}
Rombach, R., Blattmann, A., Lorenz, D., Esser, P., and Ommer, B.
\newblock High-resolution image synthesis with latent diffusion models.
\newblock In \emph{CVPR}, 2022.

\bibitem[Salehi et~al.(2021)Salehi, Mirzaei, Hendrycks, Li, Rohban, and Sabokrou]{salehi2021unified}
Salehi, M., Mirzaei, H., Hendrycks, D., Li, Y., Rohban, M.~H., and Sabokrou, M.
\newblock A unified survey on anomaly, novelty, open-set, and out-of-distribution detection: Solutions and future challenges.
\newblock \emph{arXiv preprint arXiv:2110.14051}, 2021.

\bibitem[Sehwag et~al.(2021)Sehwag, Chiang, and Mittal]{sehwag2021ssd}
Sehwag, V., Chiang, M., and Mittal, P.
\newblock Ssd: A unified framework for self-supervised outlier detection.
\newblock In \emph{ICLR}, 2021.

\bibitem[Sharma et~al.(2018)Sharma, Ding, Goodman, and Soricut]{sharma2018conceptual}
Sharma, P., Ding, N., Goodman, S., and Soricut, R.
\newblock Conceptual captions: A cleaned, hypernymed, image alt-text dataset for automatic image captioning.
\newblock In \emph{ACL}, 2018.

\bibitem[Sun \& Li(2022)Sun and Li]{sun2022dice}
Sun, Y. and Li, Y.
\newblock Dice: Leveraging sparsification for out-of-distribution detection.
\newblock In \emph{European Conference on Computer Vision}, pp.\  691--708. Springer, 2022.

\bibitem[Sun et~al.(2022)Sun, Ming, Zhu, and Li]{sun2022out}
Sun, Y., Ming, Y., Zhu, X., and Li, Y.
\newblock Out-of-distribution detection with deep nearest neighbors.
\newblock In \emph{ICML}, 2022.

\bibitem[Tian et~al.(2020)Tian, Krishnan, and Isola]{tian2020contrastive}
Tian, Y., Krishnan, D., and Isola, P.
\newblock Contrastive multiview coding.
\newblock In \emph{ECCV}, 2020.

\bibitem[Touvron et~al.(2023)Touvron, Lavril, Izacard, Martinet, Lachaux, Lacroix, Rozi{\`e}re, Goyal, Hambro, Azhar, et~al.]{touvron2023llama}
Touvron, H., Lavril, T., Izacard, G., Martinet, X., Lachaux, M.-A., Lacroix, T., Rozi{\`e}re, B., Goyal, N., Hambro, E., Azhar, F., et~al.
\newblock Llama: Open and efficient foundation language models.
\newblock \emph{arXiv preprint arXiv:2302.13971}, 2023.

\bibitem[Van Der~Maaten(2014)]{van2014accelerating}
Van Der~Maaten, L.
\newblock Accelerating t-sne using tree-based algorithms.
\newblock \emph{JMLR}, 15\penalty0 (1):\penalty0 3221--3245, 2014.

\bibitem[Van~Horn et~al.(2018)Van~Horn, Mac~Aodha, Song, Cui, Sun, Shepard, Adam, Perona, and Belongie]{van2018inaturalist}
Van~Horn, G., Mac~Aodha, O., Song, Y., Cui, Y., Sun, C., Shepard, A., Adam, H., Perona, P., and Belongie, S.
\newblock The inaturalist species classification and detection dataset.
\newblock In \emph{CVPR}, 2018.

\bibitem[Vaswani et~al.(2017)Vaswani, Shazeer, Parmar, Uszkoreit, Jones, Gomez, Kaiser, and Polosukhin]{vaswani2017attention}
Vaswani, A., Shazeer, N., Parmar, N., Uszkoreit, J., Jones, L., Gomez, A.~N., Kaiser, {\L}., and Polosukhin, I.
\newblock Attention is all you need.
\newblock In \emph{NeurIPS}, 2017.

\bibitem[Vaze et~al.(2022)Vaze, Han, Vedaldi, and Zisserman]{vaze2021open}
Vaze, S., Han, K., Vedaldi, A., and Zisserman, A.
\newblock Open-set recognition: A good closed-set classifier is all you need?
\newblock 2022.

\bibitem[Wah et~al.(2011)Wah, Branson, Welinder, Perona, and Belongie]{wah2011caltech}
Wah, C., Branson, S., Welinder, P., Perona, P., and Belongie, S.
\newblock The caltech-ucsd birds-200-2011 dataset.
\newblock 2011.

\bibitem[Wang et~al.(2019)Wang, Ge, Lipton, and Xing]{wang2019learning}
Wang, H., Ge, S., Lipton, Z., and Xing, E.~P.
\newblock Learning robust global representations by penalizing local predictive power.
\newblock In \emph{NeurIPS}, 2019.

\bibitem[Wang et~al.(2023)Wang, Li, Yao, and Li]{wang2023clipn}
Wang, H., Li, Y., Yao, H., and Li, X.
\newblock Clipn for zero-shot ood detection: Teaching clip to say no.
\newblock \emph{arXiv preprint arXiv:2308.12213}, 2023.

\bibitem[Wang et~al.(2022)Wang, Liu, Zhang, Zhang, Gong, Liu, and Han]{wang2022watermarking}
Wang, Q., Liu, F., Zhang, Y., Zhang, J., Gong, C., Liu, T., and Han, B.
\newblock Watermarking for out-of-distribution detection.
\newblock In \emph{NeurIPS}, 2022.

\bibitem[Xiao et~al.(2010)Xiao, Hays, Ehinger, Oliva, and Torralba]{xiao2010sun}
Xiao, J., Hays, J., Ehinger, K.~A., Oliva, A., and Torralba, A.
\newblock Sun database: Large-scale scene recognition from abbey to zoo.
\newblock In \emph{CVPR}, 2010.

\bibitem[Xiao et~al.(2020)Xiao, Yan, and Amit]{xiao2020likelihood}
Xiao, Z., Yan, Q., and Amit, Y.
\newblock Likelihood regret: An out-of-distribution detection score for variational auto-encoder.
\newblock In \emph{NeurIPS}, 2020.

\bibitem[Xu et~al.(2022)Xu, De~Mello, Liu, Byeon, Breuel, Kautz, and Wang]{xu2022groupvit}
Xu, J., De~Mello, S., Liu, S., Byeon, W., Breuel, T., Kautz, J., and Wang, X.
\newblock Groupvit: Semantic segmentation emerges from text supervision.
\newblock In \emph{CVPR}, 2022.

\bibitem[Yang et~al.(2021)Yang, Zhou, Li, and Liu]{yang2021oodsurvey}
Yang, J., Zhou, K., Li, Y., and Liu, Z.
\newblock Generalized out-of-distribution detection: A survey.
\newblock \emph{arXiv preprint arXiv:2110.11334}, 2021.

\bibitem[Yang et~al.(2022)Yang, Xu, Bao, He, Cao, and Huang]{yang2022optimizing}
Yang, Z., Xu, Q., Bao, S., He, Y., Cao, X., and Huang, Q.
\newblock Optimizing two-way partial auc with an end-to-end framework.
\newblock \emph{IEEE TPAMI}, 2022.

\bibitem[Yu et~al.(2015)Yu, Seff, Zhang, Song, Funkhouser, and Xiao]{yu2015lsun}
Yu, F., Seff, A., Zhang, Y., Song, S., Funkhouser, T., and Xiao, J.
\newblock Lsun: construction of a large-scale image dataset using deep learning with humans in the loop.
\newblock \emph{arXiv preprint arXiv:1506.03365}, 2015.

\bibitem[Zhang et~al.(2023)Zhang, Yang, Wang, Wang, Lin, Zhang, Sun, Du, Zhou, Zhang, Li, Liu, Chen, and Li]{zhang2023openood}
Zhang, J., Yang, J., Wang, P., Wang, H., Lin, Y., Zhang, H., Sun, Y., Du, X., Zhou, K., Zhang, W., Li, Y., Liu, Z., Chen, Y., and Li, H.
\newblock Openood v1.5: Enhanced benchmark for out-of-distribution detection.
\newblock \emph{arXiv preprint arXiv:2306.09301}, 2023.

\bibitem[Zhou et~al.(2017)Zhou, Lapedriza, Khosla, Oliva, and Torralba]{zhou2017places}
Zhou, B., Lapedriza, A., Khosla, A., Oliva, A., and Torralba, A.
\newblock Places: A 10 million image database for scene recognition.
\newblock \emph{IEEE TPAMI}, 40\penalty0 (6):\penalty0 1452--1464, 2017.

\bibitem[Zhou et~al.(2022{\natexlab{a}})Zhou, Yang, Loy, and Liu]{zhou2022conditional}
Zhou, K., Yang, J., Loy, C.~C., and Liu, Z.
\newblock Conditional prompt learning for vision-language models.
\newblock In \emph{CVPR}, 2022{\natexlab{a}}.

\bibitem[Zhou et~al.(2022{\natexlab{b}})Zhou, Yang, Loy, and Liu]{zhou2022learning}
Zhou, K., Yang, J., Loy, C.~C., and Liu, Z.
\newblock Learning to prompt for vision-language models.
\newblock \emph{IJCV}, 130\penalty0 (9):\penalty0 2337--2348, 2022{\natexlab{b}}.

\bibitem[Zhou(2022)]{zhou2022rethinking}
Zhou, Y.
\newblock Rethinking reconstruction autoencoder-based out-of-distribution detection.
\newblock In \emph{CVPR}, 2022.

\bibitem[Zhu et~al.(2023)Zhu, Li, Yao, Liu, Xu, and Han]{zhu2023unleashing}
Zhu, J., Li, H., Yao, J., Liu, T., Xu, J., and Han, B.
\newblock Unleashing mask: Explore the intrinsic out-of-distribution detection capability.
\newblock In \emph{ICML}, 2023.

\end{thebibliography}
\bibliographystyle{icml2024}

\newpage
\onecolumn
\appendix
	
\etocdepthtag.toc{mtappendix}
\etocsettagdepth{mtchapter}{none}
\etocsettagdepth{mtappendix}{subsection}
	
\renewcommand{\contentsname}{Appendix}
\tableofcontents

\section{Further Analysis}
\subsection{Understanding EOE's Effectiveness: Without Hitting Actual OOD Classes
}
\begin{figure*}[h]
\begin{center}
\includegraphics[width=0.7\textwidth]{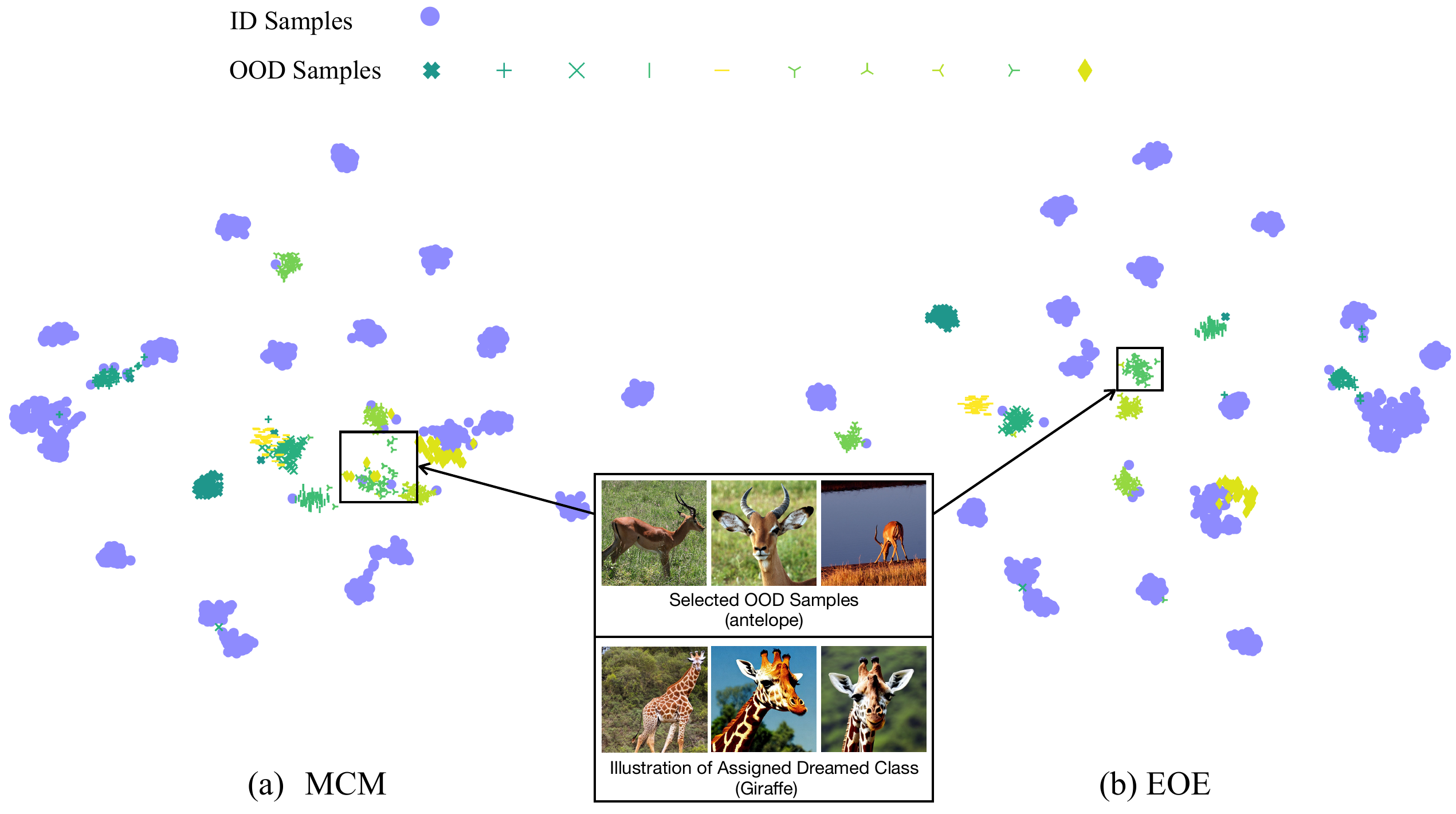}
\end{center}
  \caption{T-SNE visualizations obtained by the classifier output. ID set: ImageNet-20; OOD set: ImageNet-10. We use distinct colors to represent different OOD classes. The illustrated envisioned OOD name is the class assigned with the corresponding cluster, and its examples are generated by Stable Diffusion~\citep{rombach2022high}. 
  }
  \label{fig: visual_ImageNet20}
  \vspace{-0.25in}
\end{figure*}

Figure~\ref{fig: visual_ImageNet20} presents the visualization when using ImageNet-20 as the ID dataset. This visualization indicates that OOD samples belonging to the same class tend to be clustered together under our EOE framework. 
OOD samples within the same group are classified into the envisioned outlier class that is visually similar to them. This demonstrates that our EOE method can improve OOD detection performance without hitting actual OOD class labels.

\subsection{Further Analysis on the Design of $S_\text{EOE}$}
\label{app: score-detail}
How to utilize available outlier labels to enhance the ability of the score function to distinguish ID and OOD samples is crucial. The intuitive idea is to incorporate the envisioned class into the denominator of the MSP score:
\begin{equation}
    \label{eq: s_msp}
    S_{\text{MSP}}(x;\mathcal{Y}_\text{id},\mathcal{Y}_\text{outlier},\mathcal{T},\mathcal{I}) = \max_{i \in [1, K]} \frac{e^{s_i(x)}}{\sum_{j=1}^{K+L} e^{s_j(x)}},
\end{equation}
However, in this case, the envisioned class only functions in the denominator, which doesn't significantly impact the final score distribution, implying that the envisioned class is not fully utilized. To amplify the role of the envisioned class, EOE further subtracts the second item
($- {\beta \cdot \underset{\scriptscriptstyle k \in [K+1, K+L]}\max} \frac{e^{s_k(x)}}{\sum_{j=1}^{\scriptscriptstyle K+L} e^{s_j(x)}}$), \textit{i.e.},
\begin{equation}
    S_{\text{EOE}}(x;\mathcal{Y}_\text{id},\mathcal{Y}_\text{outlier},\mathcal{T},\mathcal{I}) = \max_{\scriptscriptstyle i \in [1, K]} \frac{e^{s_i(x)}}{\sum_{j=1}^{\scriptscriptstyle K+L} e^{s_j(x)}} - {\beta \cdot \underset{\scriptscriptstyle k \in (K, K+L]}\max} \frac{e^{s_k(x)}}{\sum_{j=1}^{\scriptscriptstyle K+L} e^{s_j(x)}},
\end{equation}
It stems from an intuition: samples visually similar to the envisioned class should have lower scores, thus making it easier to distinguish between the ID and OOD score distribution.
We conduct experiments on $\beta \in \{0, 0.25, 0.50, 0.75, 1\}$ with different outlier class number $L$, and the results are shown in Tables.~\ref{tab: number-far-ood1} to \ref{tab: number-fine-grained-ood}. Empirically, $\beta \in [0.25, 0.5]$ is a suitable choice and EOE outperforms the strong baseline MCM even if $L$ changes significantly.

\begin{table}[htbp!]
\caption{Zero-shot far OOD detection results with different outlier class label number $L$ and $\beta$.
}
\label{tab: number-far-ood1}
\centering
\fontsize{7}{8}\selectfont
\setlength\tabcolsep{3pt}
\begin{tabular}{cccccccccccccc}
\toprule
\multirow{3}{*}{\textbf{ID Dataset}} & \multirow{3}{*}{\shortstack{\textbf{Outlier Class} \\ \textbf{Label Number}}} &  \multirow{2}{*}{$\beta$} & \multicolumn{8}{c}{\textbf{OOD Dataset}} & \multicolumn{2}{c}{\multirow{2}{*}{\textbf{Average}}} \\
&&& \multicolumn{2}{c}{iNaturalist} & \multicolumn{2}{c}{SUN} & \multicolumn{2}{c}{Places} & \multicolumn{2}{c}{Texture} & \multicolumn{2}{c}{} \\            
\cmidrule(lr){4-5}\cmidrule(lr){6-7}\cmidrule(lr){8-9}\cmidrule(lr){10-11}\cmidrule(lr){12-13} 
&&& \textbf{FPR95$\downarrow$} & \textbf{AUROC$\uparrow$} & \textbf{FPR95$\downarrow$} & \textbf{AUROC$\uparrow$} & \textbf{FPR95$\downarrow$} & \textbf{AUROC$\uparrow$} & \textbf{FPR95$\downarrow$} & \textbf{AUROC$\uparrow$} & \textbf{FPR95$\downarrow$} & \textbf{AUROC$\uparrow$} \\    
\midrule
\multirow{18}{*}{\textbf{CUB-200-2011}}
& 0~(MCM) && 9.83 & 98.24 & 4.93 & 99.10 & 6.66 & 98.57 & 6.99 & 98.75 & 7.10 & 98.66 \\
\cmidrule(lr){2-13}
&\multirow{5}{*}{$L=100$} 
& 0& 0.37 & 99.92 & 0.09 & 99.97 & 0.70 & 99.80 & 0.15 & 99.96 & 0.33 & 99.91\\
&& 0.25& 0.06 & 99.98 & 0.04 & 99.99 & 0.37 & 99.88 & 0.01 & 99.99 & 0.12 & 99.96\\
&& 0.50& 0.06 & 99.98 & 0.01 & 100.00 & 0.31 & 99.90 & 0.00 & 100.00 & 0.10 & 99.97\\
&& 0.75& 0.07 & 99.98 & 0.01 & 100.00 & 0.28 & 99.92 & 0.00 & 100.00 & \textbf{0.09} & \textbf{99.98}\\
&& 1.00& 0.09 & 99.98 & 0.01 & 100.00 & 0.29 & 99.92 & 0.00 & 100.00 & 0.10 & \textbf{99.98}\\
\cmidrule(lr){2-13}
&\multirow{5}{*}{$L=300$} 
& 0 & 0.18 & 99.95 & 0.07 & 99.98 & 0.52 & 99.84 & 0.07 & 99.98 & 0.21 & 99.93\\
&& 0.25& 0.06 & 99.98 & 0.03 & 99.99 & 0.39 & 99.88 & 0.01 & 99.99 & 0.12 & \textbf{99.96} \\
&& 0.50 & 0.06 & 99.98 & 0.02 & 100.00 & 0.36 & 99.89 & 0.00 & 100.00 & \textbf{0.11} & \textbf{99.96}\\
&& 0.75 & 0.07 & 99.97 & 0.02 & 100.00 & 0.38 & 99.89 & 0.01 & 99.98 & 0.12 & \textbf{99.96}\\
&& 1.00 & 0.11 & 99.95 & 0.03 & 99.99 & 0.43 & 99.88 & 0.02 & 99.95 & 0.14 & 99.94\\
\cmidrule(lr){2-13}
&\multirow{5}{*}{$L=500$} 
& 0 & 0.13 & 99.96 & 0.05 & 99.98 & 0.54 & 99.84 & 0.07 & 99.98 & 0.20 & 99.94\\
&& 0.25 & 0.06 & 99.98 & 0.03 & 100.00 & 0.37 & 99.88 & 0.01 & 100.00 & 0.12 & 99.96\\
&& 0.50 & 0.06 & 99.98 & 0.02 & 100.00 & 0.33 & 99.90 & 0.00 & 100.00 & \textbf{0.10} & \textbf{99.97}\\
&& 0.75 & 0.08 & 99.97 & 0.02 & 100.00 & 0.32 & 99.90 & 0.00 & 99.99 & 0.11 & \textbf{99.97}\\
&& 1.00 & 0.10 & 99.96 & 0.02 & 100.00 & 0.37 & 99.90 & 0.02 & 99.97 & 0.13 & 99.96\\
\midrule
\multirow{18}{*}{\textbf{STANFORD-CARS}}
& 0~(MCM) &&0.05&	99.77&	0.02&	99.95&	0.24&	99.89&	0.02&	99.96&	0.08 &	99.89 \\
\cmidrule(lr){2-13}
&\multirow{5}{*}{$L=100$} 
& 0& 0.00 & 99.99 & 0.01 & 99.99 & 0.13 & 99.97 & 0.00 & 100.00 & 0.04 & \textbf{99.99}\\
&& 0.25& 0.00 & 99.99 & 0.01 & 99.99 & 0.11 & 99.97 & 0.00 & 100.00 & \textbf{0.03} & \textbf{99.99}\\
&& 0.50& 0.00 & 99.99 & 0.01 & 99.99 & 0.12 & 99.97 & 0.00 & 100.00 & \textbf{0.03} & \textbf{99.99}\\
&& 0.75& 0.01 & 99.97 & 0.07 & 99.97 & 0.14 & 99.95 & 0.01 & 100.00 & 0.06 & 99.97\\
&& 1.00& 0.18 & 99.89 & 0.47 & 99.86 & 0.32 & 99.89 & 0.04 & 99.97 & 0.25 & 99.90\\
\cmidrule(lr){2-13}
&\multirow{5}{*}{$L=300$} 
& 0 & 0.00 & 100.00 & 0.01 & 100.00 & 0.11 & 99.98 & 0.00 & 100.00 & \textbf{0.03} & \textbf{100.00}\\
&& 0.25 & 0.00 & 100.00 & 0.01 & 100.00 & 0.09 & 99.98 & 0.00 & 100.00 & \textbf{0.03} & \textbf{100.00}\\
&& 0.50& 0.00 & 100.00 & 0.01 & 99.99 & 0.10 & 99.98 & 0.00 & 100.00 & \textbf{0.03} & 99.99 \\
&& 0.75 & 0.00 & 99.99 & 0.03 & 99.99 & 0.12 & 99.96 & 0.01 & 100.00 & 0.04 & 99.98\\
&& 1.00 & 0.03 & 99.97 & 0.18 & 99.94 & 0.22 & 99.93 & 0.01 & 99.99 & 0.11 & 99.96\\
\cmidrule(lr){2-13}
&\multirow{5}{*}{$L=500$} 
& 0& 0.00 & 100.00 & 0.01 & 100.00 & 0.05 & 99.99 & 0.00 & 100.00 & \textbf{0.02} & \textbf{100.00}\\
&& 0.25& 0.00 & 100.00 & 0.01 & 100.00 & 0.06 & 99.99 & 0.00 & 100.00 & \textbf{0.02} & \textbf{100.00}\\
&& 0.50& 0.00 & 100.00 & 0.01 & 100.00 & 0.07 & 99.98 & 0.00 & 100.00 & \textbf{0.02} & \textbf{100.00}\\
&& 0.75& 0.00 & 100.00 & 0.01 & 100.00 & 0.11 & 99.98 & 0.00 & 100.00 & 0.03 & \textbf{100.00}\\
&& 1.00& 0.00 & 100.00 & 0.03 & 99.99 & 0.15 & 99.96 & 0.00 & 100.00 & 0.04 & 99.99\\
\midrule
\multirow{18}{*}{\textbf{Average}}
& 0~(MCM) && 4.94 & 99.01 & 2.48 & 99.53 & 3.45 & 99.23 & 3.51 & 99.35 & 3.59 & 99.28 \\
\cmidrule(lr){2-13}
&\multirow{5}{*}{$L=100$} 
& 0& 0.19 & 99.95 & 0.05 & 99.98 & 0.42 & 99.88 & 0.07 & 99.98 & 0.19 & 99.95\\
&& 0.25& 0.03 & 99.98 & 0.02 & 99.99 & 0.24 & 99.92 & 0.01 & 100.00 & 0.07 & 99.97\\
&& 0.50& 0.03 & 99.98 & 0.01 & 99.99 & 0.21 & 99.94 & 0.00 & 100.00 & \textbf{0.06} & \textbf{99.98}\\
&& 0.75& 0.04 & 99.98 & 0.04 & 99.98 & 0.21 & 99.94 & 0.01 & 100.00 & 0.08 & \textbf{99.98}\\
&& 1.00& 0.13 & 99.94 & 0.24 & 99.93 & 0.31 & 99.90 & 0.02 & 99.98 & 0.17 & 99.94\\
\cmidrule(lr){2-13}
&\multirow{5}{*}{$L=300$} 
& 0& 0.09 & 99.97 & 0.04 & 99.99 & 0.31 & 99.91 & 0.03 & 99.99 & 0.12 & 99.96 \\
&& 0.25& 0.03 & 99.99 & 0.02 & 100.00 & 0.24 & 99.93 & 0.00 & 100.00 & \textbf{0.07} & \textbf{99.98} \\
&& 0.50 & 0.03 & 99.99 & 0.01 & 100.00 & 0.23 & 99.93 & 0.00 & 100.00 & \textbf{0.07} & \textbf{99.98}\\
&& 0.75& 0.04 & 99.98 & 0.02 & 99.99 & 0.25 & 99.93 & 0.01 & 99.99 & 0.08 & 99.97 \\
&& 1.00 & 0.07 & 99.96 & 0.10 & 99.97 & 0.32 & 99.90 & 0.02 & 99.97 & 0.12 & 99.95\\
\cmidrule(lr){2-13}
&\multirow{5}{*}{$L=500$} 
& 0& 0.07 & 99.98 & 0.03 & 99.99 & 0.29 & 99.92 & 0.03 & 99.99 & 0.11 & 99.97\\
&& 0.25& 0.03 & 99.99 & 0.02 & 100.00 & 0.21 & 99.94 & 0.00 & 100.00 & 0.07 & \textbf{99.98}\\
&& 0.50& 0.03 & 99.99 & 0.01 & 100.00 & 0.20 & 99.94 & 0.00 & 100.00 & \textbf{0.06} & \textbf{99.98}\\
&& 0.75& 0.04 & 99.99 & 0.01 & 100.00 & 0.22 & 99.94 & 0.00 & 100.00 & 0.07 & \textbf{99.98}\\
&& 1.00& 0.05 & 99.98 & 0.02 & 100.00 & 0.26 & 99.93 & 0.01 & 99.99 & 0.09 & 99.97\\
\bottomrule
\end{tabular}
\end{table}

\begin{table*}
\caption{Zero-shot far OOD detection results with different outlier class label number $L$ and $\beta$. 
}
\label{tab: number-far-ood2}
\centering
\fontsize{7}{8}\selectfont
\setlength\tabcolsep{3pt}
\begin{tabular}{cccccccccccccc}
\toprule
\multirow{3}{*}{\textbf{ID Dataset}} & \multirow{3}{*}{\shortstack{\textbf{Outlier Class} \\ \textbf{Label Number}}} &  \multirow{2}{*}{$\beta$} & \multicolumn{8}{c}{\textbf{OOD Dataset}} & \multicolumn{2}{c}{\multirow{2}{*}{\textbf{Average}}} \\
&&& \multicolumn{2}{c}{iNaturalist} & \multicolumn{2}{c}{SUN} & \multicolumn{2}{c}{Places} & \multicolumn{2}{c}{Texture} & \multicolumn{2}{c}{} \\            
\cmidrule(lr){4-5}\cmidrule(lr){6-7}\cmidrule(lr){8-9}\cmidrule(lr){10-11}\cmidrule(lr){12-13} 
&&& \textbf{FPR95$\downarrow$} & \textbf{AUROC$\uparrow$} & \textbf{FPR95$\downarrow$} & \textbf{AUROC$\uparrow$} & \textbf{FPR95$\downarrow$} & \textbf{AUROC$\uparrow$} & \textbf{FPR95$\downarrow$} & \textbf{AUROC$\uparrow$} & \textbf{FPR95$\downarrow$} & \textbf{AUROC$\uparrow$} \\    
\midrule
\multirow{18}{*}{\textbf{Food-101}}
& 0~(MCM) && 0.64& 	99.78& 	0.90& 	99.75& 	1.86& 	99.58& 	4.04& 	98.62& 	1.86& 	99.43\\
\cmidrule(lr){2-13}
&\multirow{5}{*}{$L=100$} 
& 0& 0.12 & 99.97 & 0.02 & 99.99 & 0.29 & 99.94 & 2.73 & 98.97 & 0.79 & 99.72\\
&& 0.25& 0.08 & 99.98 & 0.00 & 100.00 & 0.18 & 99.96 & 2.58 & 99.00 & 0.71 & 99.73\\
&& 0.50& 0.06 & 99.98 & 0.00 & 100.00 & 0.14 & 99.97 & 2.61 & 99.01 & \textbf{0.70} & \textbf{99.74}\\
&& 0.75& 0.06 & 99.98 & 0.01 & 99.99 & 0.14 & 99.96 & 2.74 & 98.98 & 0.74 & 99.73\\
&& 1.00& 0.07 & 99.98 & 0.01 & 99.98 & 0.18 & 99.94 & 3.06 & 98.89 & 0.83 & 99.70\\
\cmidrule(lr){2-13}
&\multirow{5}{*}{$L=300$} 
& 0 & 0.09 & 99.98 & 0.00 & 100.00 & 0.11 & 99.98 & 2.55 & 99.04 & 0.69 & \textbf{99.75}\\
&& 0.25 & 0.08 & 99.98 & 0.00 & 100.00 & 0.07 & 99.98 & 2.52 & 99.01 & \textbf{0.67} & 99.74\\
&& 0.50 & 0.07 & 99.98 & 0.00 & 100.00 & 0.08 & 99.98 & 2.64 & 98.96 & 0.70 & 99.73\\
&& 0.75 & 0.07 & 99.98 & 0.00 & 99.99 & 0.09 & 99.97 & 2.91 & 98.87 & 0.77 & 99.70\\
&& 1.00 & 0.08 & 99.97 & 0.01 & 99.99 & 0.12 & 99.95 & 3.23 & 98.71 & 0.86 & 99.65\\
\cmidrule(lr){2-13}
&\multirow{5}{*}{$L=500$} 
& 0& 0.07 & 99.99 & 0.00 & 100.00 & 0.10 & 99.98 & 2.43 & 99.08 & \textbf{0.65} & \textbf{99.76}\\
&& 0.25& 0.06 & 99.99 & 0.00 & 100.00 & 0.10 & 99.98 & 2.45 & 99.04 & \textbf{0.65} & \textbf{99.76}\\
&& 0.50& 0.04 & 99.99 & 0.00 & 100.00 & 0.09 & 99.98 & 2.58 & 98.98 & 0.68 & 99.74\\
&& 0.75& 0.04 & 99.99 & 0.01 & 99.99 & 0.13 & 99.97 & 2.83 & 98.86 & 0.75 & 99.70\\
&& 1.00& 0.08 & 99.98 & 0.02 & 99.98 & 0.22 & 99.94 & 3.39 & 98.64 & 0.93 & 99.63\\
\midrule
\multirow{18}{*}{\textbf{Oxford-IIIT Pet}}
& 0~(MCM) && 2.80&	99.38&	1.05&	99.73&	2.11&	99.56&	0.80&	99.81&	1.69&	99.62\\
\cmidrule(lr){2-13}
&\multirow{5}{*}{$L=100$} 
& 0& 0.00 & 100.00 & 0.02 & 99.99 & 0.16 & 99.96 & 0.14 & 99.97 & 0.08 & 99.98\\
&& 0.25& 0.00 & 100.00 & 0.01 & 99.99 & 0.14 & 99.97 & 0.11 & 99.97 & \textbf{0.07} & \textbf{99.98}\\
&& 0.50& 0.00 & 100.00 & 0.01 & 99.99 & 0.14 & 99.97 & 0.14 & 99.97 & 0.08 & \textbf{99.98}\\
&& 0.75& 0.00 & 100.00 & 0.01 & 99.99 & 0.14 & 99.97 & 0.14 & 99.96 & 0.08 & \textbf{99.98}\\
&& 1.00& 0.00 & 100.00 & 0.01 & 99.99 & 0.14 & 99.96 & 0.16 & 99.95 & 0.08 & 99.97\\
\cmidrule(lr){2-13}
&\multirow{5}{*}{$L=300$} 
& 0& 0.00 & 100.00 & 0.01 & 99.99 & 0.15 & 99.97 & 0.15 & 99.97 & 0.08 & \textbf{99.98} \\
&& 0.25& 0.00 & 100.00 & 0.01 & 99.99 & 0.13 & 99.97 & 0.14 & 99.97 & \textbf{0.07} & \textbf{99.98} \\
&& 0.50 & 0.00 & 100.00 & 0.01 & 99.99 & 0.13 & 99.97 & 0.15 & 99.96 & 0.08 & \textbf{99.98}\\
&& 0.75 & 0.00 & 100.00 & 0.02 & 99.99 & 0.13 & 99.97 & 0.17 & 99.96 & 0.08 & \textbf{99.98}\\
&& 1.00 & 0.00 & 99.99 & 0.02 & 99.99 & 0.14 & 99.96 & 0.20 & 99.94 & 0.09 & 99.97\\
\cmidrule(lr){2-13}
&\multirow{5}{*}{$L=500$} 
& 0& 0.00 & 100.00 & 0.01 & 99.99 & 0.16 & 99.95 & 0.12 & 99.97 & \textbf{0.07} & 99.97\\
&& 0.25& 0.00 & 100.00 & 0.01 & 99.99 & 0.15 & 99.96 & 0.12 & 99.97 & \textbf{0.07} & \textbf{99.98}\\
&& 0.50& 0.00 & 100.00 & 0.01 & 99.99 & 0.15 & 99.96 & 0.12 & 99.97 & \textbf{0.07} & \textbf{99.98}\\
&& 0.75& 0.00 & 100.00 & 0.02 & 99.99 & 0.15 & 99.95 & 0.13 & 99.96 & 0.08 & 99.97\\
&& 1.00& 0.00 & 99.99 & 0.03 & 99.99 & 0.19 & 99.94 & 0.15 & 99.96 & 0.09 & 99.97\\
\midrule
\multirow{18}{*}{\textbf{Average}}
& 0~(MCM) && 1.72& 	99.58& 	0.98& 	99.74& 	1.98& 	99.57& 	2.42&   99.21& 	1.78& 	99.53\\
\cmidrule(lr){2-13}
&\multirow{5}{*}{$L=100$} 
& 0& 0.06 & 99.99 & 0.02 & 99.99 & 0.23 & 99.95 & 1.43 & 99.47 & 0.44 & 99.85\\
&& 0.25& 0.04 & 99.99 & 0.01 & 99.99 & 0.16 & 99.96 & 1.34 & 99.49 & \textbf{0.39} & \textbf{99.86}\\
&& 0.50& 0.03 & 99.99 & 0.01 & 100.00 & 0.14 & 99.97 & 1.38 & 99.49 & \textbf{0.39} & \textbf{99.86}\\
&& 0.75& 0.03 & 99.99 & 0.01 & 99.99 & 0.14 & 99.96 & 1.44 & 99.47 & 0.41 & 99.85\\
&& 1.00& 0.04 & 99.99 & 0.01 & 99.99 & 0.16 & 99.95 & 1.61 & 99.42 & 0.46 & 99.83\\
\cmidrule(lr){2-13}
&\multirow{5}{*}{$L=300$} 
& 0 & 0.05 & 99.99 & 0.01 & 100.00 & 0.13 & 99.98 & 1.35 & 99.51 & 0.38 & 99.87\\
&& 0.25 & 0.04 & 99.99 & 0.01 & 100.00 & 0.10 & 99.98 & 1.33 & 99.49 & \textbf{0.37} & \textbf{99.86}\\
&& 0.50& 0.04 & 99.99 & 0.01 & 100.00 & 0.11 & 99.98 & 1.40 & 99.46 & 0.39 & \textbf{99.86} \\
&& 0.75 & 0.04 & 99.99 & 0.01 & 99.99 & 0.11 & 99.97 & 1.54 & 99.41 & 0.43 & 99.84\\
&& 1.00 & 0.04 & 99.98 & 0.01 & 99.99 & 0.13 & 99.95 & 1.72 & 99.33 & 0.47 & 99.81\\
\cmidrule(lr){2-13}
&\multirow{5}{*}{$L=500$} 
& 0& 0.03 & 99.99 & 0.01 & 100.00 & 0.13 & 99.96 & 1.27 & 99.52 & \textbf{0.36} & \textbf{99.87}\\
&& 0.25& 0.03 & 100.00 & 0.01 & 100.00 & 0.12 & 99.97 & 1.28 & 99.51 & \textbf{0.36} & \textbf{99.87}\\
&& 0.50& 0.02 & 99.99 & 0.01 & 99.99 & 0.12 & 99.97 & 1.35 & 99.47 & 0.38 & 99.86\\
&& 0.75& 0.02 & 99.99 & 0.01 & 99.99 & 0.14 & 99.96 & 1.48 & 99.41 & 0.41 & 99.84\\
&& 1.00& 0.04 & 99.98 & 0.02 & 99.98 & 0.20 & 99.94 & 1.77 & 99.30 & 0.51 & 99.80\\
\bottomrule
\end{tabular}
\end{table*}

\begin{table}[htbp!]
\caption{Zero-shot far OOD detection results for ImageNet-1K as the ID dataset with different outlier class label number $L$ and $\beta$. 
} %
\label{tab: beta-number-imagenet-1k}
\centering
\fontsize{7}{8}\selectfont
\setlength\tabcolsep{4pt}
\begin{tabular}{cccccccccccc}
\toprule
\multirow{3}{*}{\shortstack{\textbf{Outlier Class} \\ \textbf{Label Number}}} &  \multirow{2}{*}{$\beta$} & \multicolumn{8}{c}{\textbf{OOD Dataset}}  & \multicolumn{2}{c}{\multirow{2}{*}{\textbf{Average} }} \\
                        && \multicolumn{2}{c}{iNaturalist} & \multicolumn{2}{c}{SUN} & \multicolumn{2}{c}{Places} & \multicolumn{2}{c}{Texture} & \multicolumn{2}{c}{}             \\
                        \cmidrule(lr){3-4}\cmidrule(lr){5-6}\cmidrule(lr){7-8}\cmidrule(lr){9-10}\cmidrule(lr){11-12}
                        && \textbf{FPR95$\downarrow$}         & \textbf{AUROC$\uparrow$}      & \textbf{FPR95$\downarrow$}           & \textbf{AUROC$\uparrow$}         & \textbf{FPR95$\downarrow$}          & \textbf{AUROC$\uparrow$}        & \textbf{FPR95$\downarrow$}         & \textbf{AUROC$\uparrow$}      & \textbf{FPR95$\downarrow$}          & \textbf{AUROC$\uparrow$}        \\  
                        \midrule
0~(MCM)
&& 30.92	& 94.61	& 37.59& 	92.57& 	44.71& 	89.77& 	57.85& 	86.11& 	42.77& 	90.77\\
\midrule
\multirow{5}{*}{$L=100$} 
& 0 & 28.91 & 94.89 & 34.90 & 93.06 & 42.23 & 90.29 & 57.22 & 86.24 & 40.81 & 91.12\\
& 0.25 & 20.40 & 96.23 & 22.65 & 95.27 & 31.82 & 92.54 & 58.54 & 85.49 & 33.36 & 92.39 \\
& 0.50 & 18.24 & 96.58 & 18.63 & 96.09 & 27.69 & 93.53 & 62.16 & 83.97 & \textbf{31.68} & \textbf{92.54}\\
& 0.75 & 19.93 & 96.38 & 18.01 & 96.28 & 27.18 & 93.82 & 67.60 & 81.96 & 33.18 & 92.11\\
& 1.00 & 23.94 & 95.91 & 19.10 & 96.16 & 28.54 & 93.72 & 72.89 & 79.71 & 36.12 & 91.37\\
\midrule
\multirow{5}{*}{$L=300$} 
& 0 & 29.42 & 94.78 & 33.83 & 93.28 & 41.11 & 90.52 & 57.03 & 86.26 & 40.35 & 91.21\\
& 0.25 & 20.19 & 96.17 & 22.64 & 95.25 & 31.77 & 92.65 & 57.19 & 85.69 & 32.94 & 92.44\\
& 0.50 & 18.17 & 96.52 & 19.57 & 95.90 & 28.50 & 93.50 & 60.84 & 84.16 & \textbf{31.77} & \textbf{92.52}\\
& 0.75 & 19.95 & 96.24 & 19.85 & 95.89 & 28.49 & 93.59 & 66.09 & 81.87 & 33.59 & 91.90\\
& 1.00 & 24.25 & 95.56 & 22.03 & 95.53 & 30.77 & 93.22 & 73.04 & 79.15 & 37.52 & 90.86\\
\midrule
\multirow{5}{*}{$L=500$} 
& 0 &  24.34 & 95.62 & 31.04 & 93.77 & 38.92 & 90.94 & 55.87 & 86.56 & 37.54 & 91.72\\
& 0.25 & 12.29 & 97.52 & 20.40 & 95.73 & 30.16 & 92.95 & 57.53 & 85.64 & 30.09 & 92.96  \\
& 0.50 & 8.20 & 98.19 & 16.96 & 96.42 & 26.69 & 93.82 & 62.23 & 83.80 & \textbf{28.52} & \textbf{93.06} \\
& 0.75 & 7.14 & 98.33 & 16.55 & 96.52 & 26.31 & 94.01 & 68.18 & 81.27 & 29.55 & 92.53 \\
& 1.00 & 7.47 & 98.20 & 17.92 & 96.30 & 28.04 & 93.81 & 75.04 & 78.36 & 32.12 & 91.66\\
\bottomrule
\end{tabular}
\end{table}

\begin{table}[htbp!]
\centering
\caption{Zero-shot near OOD detection results with different outlier class label number $L$ and $\beta$. $K$ is the number of classes in the corresponding ID dataset. 
}
\label{tab: beta-number-near-ood}
\fontsize{8}{10}\selectfont
\setlength\tabcolsep{7pt}
\begin{tabular}{cclcccccc}
\toprule
\multirow{2}{*}{\shortstack{\textbf{Outlier Class} \\ \textbf{Label Number}}} & \multirow{2}{*}{$\beta$} & \textbf{ID} & \multicolumn{2}{c}{ImageNet-10} & \multicolumn{2}{c}{ImageNet-20}& \multicolumn{2}{c}{\multirow{2}{*}{\textbf{Average}}} \\
 && \textbf{OOD}        &  \multicolumn{2}{c}{ImageNet-20} & \multicolumn{2}{c}{ImageNet-10} \\

\cmidrule(rl){4-5}
\cmidrule(rl){6-7}
\cmidrule(rl){8-9}
&&&  \textbf{FPR95$\downarrow$} & \textbf{AUROC$\uparrow$}               
&  \textbf{FPR95$\downarrow$} & \textbf{AUROC$\uparrow$}   
&  \textbf{FPR95$\downarrow$} & \textbf{AUROC$\uparrow$}   \\
\midrule
$L=0$~(MCM)    & && 5.00 & 98.71  & 17.40 & 97.87 & 11.20 & 98.29    \\
\midrule
\multirow{5}{*}{$L=1\times K$} 
& 0 && 6.43 & 98.49 & 14.07 & 98.12 & 10.25 & \textbf{98.30}\\
& 0.25 && 7.40 & 97.95 & 9.40 & 98.63 & \textbf{8.40} & 98.29\\
& 0.50 && 12.73 & 96.64 & 10.20 & 98.26 & 11.47 & 97.45\\
& 0.75 && 24.40 & 94.10 & 19.27 & 97.04 & 21.83 & 95.57\\
& 1.00 && 60.63 & 89.24 & 33.40 & 95.07 & 47.02 & 92.15\\
\midrule
\multirow{5}{*}{$L=3\times K$} 
& 0 && 6.50 & 98.62 & 16.40 & 97.95 & 11.45 & 98.28\\
& 0.25 && 4.20 & 99.09 & 13.93 & 98.10 & \textbf{9.07} & \textbf{98.59}\\
& 0.50 && 4.00 & 98.84 & 15.47 & 97.83 & 9.73 & 98.33\\
& 0.75 && 11.10 & 97.66 & 22.13 & 96.95 & 16.62 & 97.30 \\
& 1.00 && 32.60 & 94.79 & 30.00 & 95.45 & 31.30 & 95.12\\
\midrule
\multirow{5}{*}{$L=10\times K$} 
& 0 && 6.10 & 98.49 & 19.27 & 97.74 & 12.68 & 98.12   \\
& 0.25 && 3.63 & 99.11 & 16.20 & 97.92 & 9.92 & \textbf{98.52}\\
& 0.50 && 1.97 & 99.18 & 15.60 & 97.68 & \textbf{8.78} & 98.43\\
& 0.75 && 4.30 & 98.72 & 19.07 & 96.81 & 11.68 & 97.77\\
& 1.00 && 18.37 & 97.26 & 30.67 & 95.16 & 24.52 & 96.21\\
\bottomrule
\end{tabular}
\end{table}

\begin{table}[htbp!]
\centering
\caption{Zero-shot fine-grained OOD detection results with different outlier class label number $L$ and $\beta$. 
}
\label{tab: number-fine-grained-ood}
\fontsize{7}{8}\selectfont
\setlength\tabcolsep{4pt}
\begin{tabular}{cclcccccccccccccc}
\toprule
\multirow{2}{*}{\shortstack{\textbf{Outlier Class} \\ \textbf{Label Number}}} & \multirow{2}{*}{$\beta$} & \textbf{ID} & \multicolumn{2}{c}{CUB-100} & \multicolumn{2}{c}{Stanford-Cars-98} & \multicolumn{2}{c}{Food-50} & \multicolumn{2}{c}{Oxford-Pet-18} & \multicolumn{2}{c}{\multirow{2}{*}{\textbf{Average}}}\\
 && \textbf{OOD}        &  \multicolumn{2}{c}{CUB-100} & \multicolumn{2}{c}{Stanford-Cars-98} & \multicolumn{2}{c}{Food-51} & \multicolumn{2}{c}{Oxford-Pet-19}  \\

\cmidrule(rl){4-5}
\cmidrule(rl){6-7}
\cmidrule(rl){8-9}
\cmidrule(rl){10-11}
\cmidrule(rl){12-13}
\cmidrule(rl){14-15}
&&&  \textbf{FPR95$\downarrow$} & \textbf{AUROC$\uparrow$}               
&  \textbf{FPR95$\downarrow$} & \textbf{AUROC$\uparrow$}
&  \textbf{FPR95$\downarrow$} & \textbf{AUROC$\uparrow$}
&  \textbf{FPR95$\downarrow$} & \textbf{AUROC$\uparrow$}
&  \textbf{FPR95$\downarrow$} & \textbf{AUROC$\uparrow$}   \\
\midrule
$L=0$~(MCM)  &&&83.58 & 67.51 & 83.99 & 68.71 & 43.38 & 91.75 & 63.92 & 84.88 & 68.72 & 78.21\\
\midrule
\multirow{5}{*}{$L=100$} 
& 0 && 79.58 & 71.23 & 81.89 & 68.99 & 40.74 & 91.63 & 64.89 & 88.16 & 66.77 & 80.00 \\
& 0.25 && 76.79 & 72.31 & 77.73 & 70.92 & 38.36 & 91.83 & 56.03 & 89.63 & 62.23 & \textbf{81.17}  \\
& 0.50 && 75.58 & 71.92 & 74.69 & 71.52 & 40.79 & 90.93 & 47.63 & 90.03 & 59.67 & 81.10 \\
& 0.75 && 75.54 & 70.14 & 72.89 & 70.85 & 45.39 & 89.29 & 44.52 & 89.46 & \textbf{59.59} & 79.93 \\
& 1.00 && 77.10 & 67.80 & 73.94 & 69.14 & 50.26 & 87.30 & 44.06 & 88.16 & 61.34 & 78.10 \\
\midrule
\multirow{5}{*}{$L=300$} 
& 0 && 80.55 & 70.72 & 80.90 & 69.26 & 46.90 & 90.03 & 61.89 & 87.52 & 67.56 & 79.38  \\
& 0.25 && 76.01 & 72.75 & 76.35 & 71.45 & 42.63 & 90.13 & 52.32 & 89.97 & 61.83 & 81.07 \\
& 0.50 && 72.62 & 73.51 & 72.64 & 72.71 & 42.30 & 89.13 & 44.60 & 91.51 & 58.04 & \textbf{81.71} \\
& 0.75 && 71.53 & 72.66 & 70.01 & 72.70 & 44.34 & 87.38 & 36.88 & 92.11 & \textbf{55.69} & 81.21 \\
& 1.00 && 71.99 & 70.63 & 70.90 & 71.35 & 47.51 & 85.29 & 35.84 & 91.80 & 56.56 & 79.77 \\
\midrule
\multirow{5}{*}{$L=500$} 
& 0 && 79.65 & 71.29 & 80.90 & 69.55 & 41.13 & 91.61 & 59.90 & 87.94 & 65.39 & 80.10\\
& 0.25 && 74.74 & 73.41 & 76.83 & 71.60 & 37.95 & 91.96 & 52.55 & 90.33 & 60.52 & 81.82\\
& 0.50 && 70.57 & 74.18 & 73.42 & 72.77 & 37.50 & 91.35 & 44.53 & 91.88 & 56.51 & \textbf{82.54}\\
& 0.75 && 69.35 & 73.27 & 70.12 & 72.77 & 40.60 & 89.92 & 38.03 & 92.48 & \textbf{54.52} & 82.11\\
& 1.00 && 70.14 & 71.12 & 70.10 & 71.33 & 45.39 & 87.92 & 36.57 & 92.05 & 55.55 & 80.61\\
\bottomrule
\end{tabular}
\end{table}

\section{Experimental Details}
\subsection{Dataset Details}
\label{app: dataset details}
In this section, we explain in detail how the ID dataset and OOD dataset are divided. 

\textbf{Far OOD Detection.} 
Following the settings of MCM~\citep{ming2022delving}, we conduct our experiments on large-scale datasets\footnote{\url{https://github.com/deeplearning-wisc/large_scale_ood}}. The ID datasets include CUB-200-2011, STANFORD-CARS, Food-101, Oxford-IIIT Pet and ImageNet-1K. The OOD datasets are curated by MOS~\citep{huang2021mos}, including subsets of iNaturalist, SUN, Places, and Textures. The OOD datasets are constructed based on the concept of the group, \textit{i.e.}, decomposing a large semantic space into smaller groups with similar concepts, thereby simplifying the decision boundaries of OOD class labels and ID class labels. All OOD datasets exclude class labels that are present in the ID datasets. Based on the selected concepts, 10,000 images are sampled from each of the iNaturalist, SUN, and Places datasets. For the Textures dataset, the entire dataset is used.

\textbf{Near OOD Detection.} 
We use ImageNet-10 and ImageNet-20 curated by MCM, which are extracted from ImageNet-1K. Specifically, the class labels of ImageNet-10 and ImageNet-20 are shown in Figure~\ref{app dataset: imagene10and20}

\textbf{Fine-grained OOD Detection.}
We perform random partitioning of classes in the datasets CUB-200-2011, STANFORD-CARS, Food-101, and Oxford-IIIT Pet. We designate half of these classes as ID classes and the other half as OOD classes. This partitioning ensures that there is no overlap between the above ID dataset and the corresponding OOD dataset. After random partitioning, the ID/OOD datasets are CUB-100/CUB100, Stanford-Cars-98/Stanford-Cars-98, Food-50/Food-51, and Oxford-Pet-18/Oxford-Pet-19. In order to illustrate our random division more clearly, we take CUB-100(ID) and CUB-100(OOD) as examples, as shown in Figure~\ref{app dataset: cub100}.

\begin{figure}[htbp!]
\centering
\begin{tikzpicture}
\small
\definecolor{chatcolor1}{HTML}{5fedb7} 
\definecolor{shadecolor}{gray}{0.9}
\fontfamily{cmss}\selectfont

\node[align=left, text width=0.31\textwidth, fill=shadecolor, rounded corners=1mm, anchor=north west] (node1) at ([yshift=-0.4cm]node1.south west |- node2.south) {
\textbf{ImageNet-10:}\\
\begin{tabular}{@{}l}
- "n04552348": "warplane"\\
- "n04285008": "sports car"\\
- "n01530575": "brambling bird"\\
- "n02123597": "Siamese cat"\\
- "n02422699": "antelope"\\
- "n02107574": "swiss mountain dog"\\
- "n01641577": "bull frog"\\
- "n03417042": "garbage truck"\\
- "n02389026": "horse"\\
- "n03095699": "container ship"\\
\end{tabular}
};
\node[align=left, text width=0.61\textwidth, fill=chatcolor1, rounded corners=1mm, anchor=north east] (node2) at (node1.north -| {$(0,0)+(0.95\textwidth,0)$}) {\textbf{ImageNet-20:}\\
\begin{tabular}{@{}l l}
- "n04147183": "sailboat"& - "n02951358": "canoe"\\
- "n02782093": "balloon"& - "n04389033": "tank"\\ 
- "n03773504": "missile" & - "n02917067": "bullet train"\\
- "n02317335": "starfish" & - "n01632458": "spotted salamander"\\
- "n01630670": "common newt" & - "n01631663": "eft"\\
- "n02391049": "zebra" & - "n01693334": "green lizard"\\
- "n01697457": "African crocodile" & - "n02120079": "Arctic fox"\\
- "n02114367": "timber wolf"& - "n02132136": "brown bear"\\
- "n03785016": "moped" & - "n04310018": "steam locomotive"\\
- "n04266014": "space shuttle" & - "n04252077": "snowmobile"\\
\end{tabular}
};

\node[draw, black, thick, rounded corners=3mm, inner sep=7pt, fit=(node1) (node2)] {};

\end{tikzpicture}
\caption{ImageNet-10 and ImageNet-20 class labels. To illustrate how each dataset is adapted to be ID-OOD.}
\label{app dataset: imagene10and20}
\end{figure}

\begin{figure}[htbp!]
\centering
\begin{tikzpicture}
\small
\definecolor{chatcolor1}{HTML}{5fedb7} 
\definecolor{shadecolor}{gray}{0.9}
\fontfamily{cmss}\selectfont

\node[align=left, text width=0.45\textwidth, fill=shadecolor, rounded corners=1mm, anchor=north west] (node1) at ([yshift=-0.4cm]node1.south west |- node2.south) {
\textbf{CUB-100~(ID):}\\
\begin{tabular}{@{}l l}
- "Laysan Albatross" & - "Groove billed Ani"\\
- "Crested Auklet" & - "Parakeet Auklet" \\
- "Rusty Blackbird" & - "Red faced Cormorant"\\
- "Indigo Bunting" & - "Painted Bunting"\\
- "Spotted Catbird" & - "Yellow breasted Chat"\\ 
- "Eastern Towhee" & - "Chuck will Widow"\\
- "Bobolink" & - "Pelagic Cormorant"\\ 
- "Bronzed Cowbird" & - "Shiny Cowbird"\\ 
- "American Crow" & - "Golden winged Warbler"\\
- "Acadian Flycatcher" & - "Olive sided Flycatcher"\\
- "Gadwall"& - "Yellow bellied Flycatcher"\\
- "Frigatebird" & - "Scissor tailed Flycatcher"\\
- "American Goldfinch" & - "Boat tailed Grackle"\\
- "Eared Grebe"& - "Glaucous winged Gull"\\
- "Pied billed Grebe"& - "Blue Grosbeak"\\
- "Evening Grosbeak"& - "Pine Grosbeak"\\
- "Herring Gull"& - "Pigeon Guillemot"\\
- "Horned Grebe"& - "Rose breasted Grosbeak"\\
- "Ivory Gull"& - "Ring billed Gull"\\
- "Anna Hummingbird"& - "Pomarine Jaeger"\\
- "Blue Jay"& - "Green Jay"\\
- "Gray Kingbird"& - "Ringed Kingfisher"\\
- "Horned Lark"& - "Red legged Kittiwake"\\
- "Western Meadowlark"& - "Red breasted Merganser"\\
- "Ovenbird"& - "White breasted Nuthatch"\\
- "Orchard Oriole"& - "Hooded Oriole"\\
- "Sayornis"& - "Western Wood Pewee"\\
- "Whip poor Will" & -"White necked Raven"\\
- "Fox Sparrow"& - "Brewer Sparrow"\\
- "Chipping Sparrow"& - "Clay colored Sparrow"\\
- "House Sparrow"& - "Great Grey Shrike"\\
- "Grasshopper Sparrow"& - "Henslow Sparrow"\\
- "Le Conte Sparrow"& - "Savannah Sparrow"\\
- "Vesper Sparrow"& - "Bank Swallow"\\
- "Cliff Swallow"& - "Scarlet Tanager"\\
- "Black Tern"& - "Golden winged Warbler"\\
- "Brown Thrasher"& - "Sage Thrasher"\\
- "Black capped Vireo"& - "Blue headed Vireo"\\
- "Philadelphia Vireo"& - "Warbling Vireo"\\
- "Caspian Tern"& - "Common Yellowthroat"\\
- "Bewick Wren"& - "Cape May Warbler"\\
- "Purple Finch"& - "Hooded Warbler"\\
- "Magnolia Warbler"& - "Myrtle Warbler"\\
- "Palm Warbler" & - "Orange crowned Warbler"\\
- "Pine Warbler"& - "Prothonotary Warbler"\\
- "Swainson Warbler"& - "Tennessee Warbler"\\
- "Worm eating Warbler"& - "Northern Waterthrush"\\
- "Cedar Waxwing"& - "Red bellied Woodpecker"\\
- "Downy Woodpecker"& - "Blue winged Warbler"\\
- "House Wren" & - "Black throated"\\ & \phantom{-} "Blue Warbler" \\
\end{tabular}
};
\node[align=left, text width=0.45\textwidth, fill=chatcolor1, rounded corners=1mm, anchor=north east] (node2) at (node1.north -| {$(0,0)+(0.95\textwidth,0)$}) {\textbf{CUB-100~(OOD):}\\
\begin{tabular}{@{}l l}
- "Sooty Albatross"& -"Black footed Albatross"\\
- "Least Auklet"& - "Rhinoceros Auklet"\\
- "Brewer Blackbird"& -"Red winged Blackbird"\\
- "Gray Catbird"& - "Lazuli Bunting"\\
- "Cardinal"& - "Yellow headed Blackbird"\\
- "Brandt Cormorant"& -"Brown Creeper"\\
- "Fish Crow"& - "Black billed Cuckoo"\\
- "Mangrove Cuckoo"& - "Yellow billed Cuckoo"\\
- "Northern Flicker"& - "Gray crowned Rosy Finch"\\
- "Northern Fulmar"& -"Least Flycatcher"\\
- "Heermann Gull"& - "Great Crested Flycatcher"\\
- "Western Grebe"& - "European Goldfinch"\\
- "California Gull"& - "Vermilion Flycatcher"\\
- "Nighthawk"& -"Ruby throated Hummingbird"\\
- "Western Gull"& -"Rufous Hummingbird"\\
- "Green Violetear"& - "Long tailed Jaeger"\\
- "Florida Jay"& - "Dark eyed Junco"\\
- "Common Raven"& - "Belted Kingfisher"\\
- "Green Kingfisher"& - "Pied Kingfisher"\\
- "Pacific Loon" & -"White breasted Kingfisher"\\
- "Mallard"& -"Hooded Merganser"\\
- "Mockingbird"& -"Slaty backed Gull"\\
- "Brown Pelican"& -"Baltimore Oriole"\\
- "Scott Oriole"& - "Clark Nutcracker"\\
- "White Pelican"& -"American Pipit"\\
- "Horned Puffin"& - "Tropical Kingbird"\\
- "American Redstart"& -"Black throated Sparrow"\\
- "Loggerhead Shrike"& -"Baird Sparrow"\\
- "Geococcyx"& -"Field Sparrow"\\
- "Harris Sparrow"& - "Lincoln Sparrow"\\
- "Tree Sparrow"& - "White crowned Sparrow"\\
- "Song Sparrow"& - "Nelson Sharp tailed Sparrow"\\
- "Seaside Sparrow"& -"White throated Sparrow"\\
- "Summer Tanager"& - "Barn Swallow"\\
- "Tree Swallow"& -"Cape Glossy Starling"\\
- "Artic Tern"& -"Bay breasted Warbler"\\
- "Elegant Tern"& - "Chestnut sided Warbler"\\
- "Least Tern"& -"Green tailed Towhee"\\
- "Red eyed Vireo"& - "White eyed Vireo"\\
- "Common Tern"& -"Black and white Warbler"\\
- "Canada Warbler"& -"Cerulean Warbler"\\
- "Forsters Tern"& -"Kentucky Warbler"\\
- "Mourning Warbler"& -"Nashville Warbler"\\
- "Prairie Warbler"& -"Wilson Warbler"\\
- "Yellow Warbler"& - "Louisiana Waterthrush"\\
- "Marsh Wren"& -"Pileated Woodpecker"\\
- "Winter Wren"& -"Red cockaded Woodpecker"\\
- "Cactus Wren"& - "Red headed Woodpecker"\\
- "Carolina Wren"& -"Bohemian Waxwing"\\
- "Rock Wren" & - "American Three toed"\\
              & \phantom{-} "Woodpecker" \\
\end{tabular}
};
\node[draw, black, thick, rounded corners=3mm, inner sep=7pt, fit=(node1) (node2)] {};

\end{tikzpicture}
\caption{CUB-100~(ID) and CUB-100~(OOD) class labels. To illustrate how each dataset is adapted to be ID-OOD.}
\label{app dataset: cub100}
\end{figure}

\subsection{Implement Details}
\subsubsection{Environment}
EOE does not require any training. All experiments are conducted under the zero-shot setting without training samples. All experiments are performed using the PyTorch 1.13 library~\citep{paszke2019pytorch} and Python 3.10.8, running on an NVIDIA A100 80GB PCIe GPU and AMD EPYC 7H12 CPU.

\subsubsection{Evaluation Metrics}
\label{app: aupr}
We use the most commonly used metrics~(FPR95 and AUROC)~\citep{yang2022optimizing} in the OOD detection community on the main pages. To further explore the performance of EOE under other metrics, we report results in terms of AUPR on CLIP(ViT-B/16) backbone in Table~\ref{tab: AUPR-far-ood-imagenet}. Our EOE achieves the best results on AUPR.

\begin{table}[htbp!]
\caption{Performance in terms of AUPR. ID dataset: The experiments are zero-shot OOD detection results with ImageNet-1K as the ID dataset. The \textbf{black bold} indicates the best performance. The \colorbox{gray!22}{gray} indicates that the comparative methods require training or an additional massive auxiliary dataset.} %
\label{tab: AUPR-far-ood-imagenet}
\centering
\fontsize{7}{8}\selectfont
\setlength\tabcolsep{10pt}
\begin{tabular}{cccccc}
\toprule
\multirow{3}{*}{\textbf{Method}} & \multicolumn{4}{c}{\textbf{OOD Dataset}}  & \multirow{2}{*}{\textbf{Average}} \\
& iNaturalist & SUN & Places & Texture &              \\                        
\midrule
\rowcolor{gray!22}
CLIPN & 99.15 & 98.59 & 98.22 & 98.38 & 98.59\\
Energy & 96.84 & 96.50 & 96.16 & 94.66 & 96.04\\
MaxLogit & 97.74 & 97.12 & 96.65 & 95.61 & 96.78\\
MCM
&98.86 & 98.28 & 97.49 & 98.04 & 98.17\\
EOE~(Ours)
& 99.47& 99.01 & 98.27 & 97.97 & \textbf{98.68}\\

\bottomrule
\end{tabular}
\end{table}

\begin{table}[htbp!]
\caption{Additional empirical results with CIFAR-10 and CIFAR-100 as ID datasets. The \textbf{bold} indicates the best performance on each dataset. The \colorbox{gray!22}{gray} indicates that the comparative methods require training or an additional massive auxiliary dataset.}
\label{tab: cifar far-ood}
\centering
\fontsize{7}{8}\selectfont
\setlength\tabcolsep{6pt}
\begin{tabular}{ccccccccccccc}
\toprule
\multirow{3}{*}{\textbf{ID Dataset}} & \multirow{3}{*}{\textbf{Method}} & \multicolumn{8}{c}{\textbf{OOD Dataset}} & \multicolumn{2}{c}{\multirow{2}{*}{\textbf{Average}}} \\
& & \multicolumn{2}{c}{SVHN} & \multicolumn{2}{c}{LSUN} & \multicolumn{2}{c}{Texture} & \multicolumn{2}{c}{Places} & \multicolumn{2}{c}{} \\            
\cmidrule(lr){3-4}\cmidrule(lr){5-6}\cmidrule(lr){7-8}\cmidrule(lr){9-10}\cmidrule(lr){11-12} 
& & \textbf{FPR95$\downarrow$} & \textbf{AUROC$\uparrow$} & \textbf{FPR95$\downarrow$} & \textbf{AUROC$\uparrow$} & \textbf{FPR95$\downarrow$} & \textbf{AUROC$\uparrow$} & \textbf{FPR95$\downarrow$} & \textbf{AUROC$\uparrow$} & \textbf{FPR95$\downarrow$} & \textbf{AUROC$\uparrow$} \\    
\midrule

\rowcolor{gray!22}
\cellcolor{white}
\multirow{5}{*}{\textbf{CIFAR-10}}
& CLIPN & 53.28 & 74.20 & 27.89 & 92.72 & 3.58 & 98.93 & 9.82 & 96.98 & 23.64 & 90.63 \\
& Energy &  18.97 & 96.67 & 60.60 & 88.81 & 16.13 & 96.59 & 17.48 & 95.42 & 28.29 & 94.37\\
& MaxLogit & 6.50 & 98.27 & 36.54 & 94.20 & 11.37 & 97.61 & 16.67 & 95.56 & 17.77 & 96.41\\
& MCM & 3.98 & 99.03 & 5.12 & 98.72 & 16.35 & 96.44 & 36.55 & 90.79 & 15.50 & 96.25\\
& EOE~(Ours) & 5.78 & 98.20 & 4.69 & 98.64 & 5.61 & 98.66 & 11.94 & 96.36 & \textbf{7.01} & \textbf{97.97}\\
\midrule
\rowcolor{gray!22}
\cellcolor{white}
\multirow{5}{*}{\textbf{CIFAR-100}} 
& CLIPN & 71.72 &	68.20 &	84.42&	80.90&	37.74&	90.92&	51.06&	87.25&	61.24 &	81.82 \\
& Energy& 72.54 & 88.20 & 93.64 & 73.08 & 65.55 & 80.43 & 59.86 & 83.47 & 72.90 & 81.30\\
& MaxLogit & 59.05 & 91.01 & 82.48 & 83.06 & 62.82 & 82.08 & 65.58 & 80.88 & 67.48 & 84.26\\
& MCM & 64.45 & 89.96 & 47.26 & 91.69 & 90.30 & 73.61 & 98.42 & 61.37 & 75.11 & 79.16\\
& EOE~(Ours) & 68.47 & 88.78 & 36.85 & 93.29 & 66.89 & 82.64 & 77.60 & 78.06 & \textbf{62.45} & \textbf{85.69}\\
\midrule
\rowcolor{gray!22}
\cellcolor{white}
\multirow{5}{*}{\textbf{Average}} 
& CLIPN & 62.50 & 71.20 & 56.16 & 86.81 & 20.66 & 94.93 & 30.44 & 92.12 & 42.44 & 86.23 \\
& Energy & 45.76 & 92.44 & 77.12 & 80.95 & 40.84 & 88.51 & 38.67 & 89.45 & 50.60 & 87.84 \\
& MaxLogit & 32.78 & 94.64 & 59.51 & 88.63 & 37.10 & 89.85 & 41.13 & 88.22 & 42.63 & 90.34 \\
& MCM & 34.22 & 94.50 & 26.19 & 95.21 & 53.33 & 85.03 & 67.49 & 76.08 & 45.31 & 87.71 \\
& EOE~(Ours) & 37.31 & 93.43 & 22.02 & 95.75 & 37.45 & 90.31 & 45.79 & 86.77 & \textbf{35.64} & \textbf{91.56}\\
\bottomrule
\end{tabular}
\end{table}

\begin{table}[t]
\centering
\caption{Zero-shot \textbf{near} OOD detection results on large-scale datasets. The \textbf{bold} indicates the best performance on each dataset.}
\label{app-tab: near-ood}
\fontsize{7}{8}\selectfont
\setlength\tabcolsep{6pt}
\begin{tabular}{clcccccc}
\toprule

\multirow{2}{*}{\textbf{Method}} & \textbf{ID} & \multicolumn{2}{c}{ImageNet} & \multicolumn{2}{c}{ImageNet}& \multicolumn{2}{c}{\multirow{2}{*}{\textbf{Average}}} \\
 & \textbf{OOD}        &  \multicolumn{2}{c}{SSB-hard} & \multicolumn{2}{c}{NINCO} \\

\cmidrule(rl){3-4}
\cmidrule(rl){5-6}
\cmidrule(rl){7-8}
&&  \textbf{FPR95$\downarrow$} & \textbf{AUROC$\uparrow$}               
&  \textbf{FPR95$\downarrow$} & \textbf{AUROC$\uparrow$}   
&  \textbf{FPR95$\downarrow$} & \textbf{AUROC$\uparrow$}   \\
\midrule
Energy    && 90.51 & 61.66 & 86.10 & 69.94 & 88.31 & 65.80\\
MaxLogit    && 88.25 & 62.68 & 80.35 & 72.81 & 84.30 & 67.75\\
MCM    && 88.71 & 64.69 & 79.40 & 74.27 & 84.06 & 69.48    \\
EOE~(Ours)    && \textbf{85.99} & \textbf{66.36} & \textbf{73.63} & \textbf{76.93} & \textbf{79.81} & \textbf{71.65}\\
\bottomrule
\end{tabular}
\end{table}

\section{Additional Empirical Results}
\subsection{Other OOD Detection Benchmarks}\label{app: cifar benchmark}
We conduct experiments on the CIFAR-10/CIFAR-100~(as ID datasets)~\citep{krizhevsky2009learning} benchmarks. The test OOD datasets include SVHN~\citep{netzer2011reading}, LSUN~\citep{yu2015lsun}, Texture~\citep{cimpoi2014describing}, Places~\citep{zhou2017places}. The results are shown in Table~\ref{tab: cifar far-ood}. Clearly, EOE achieves superior results on CIFAR-10/CIFAR-100. 
We additionally conduct experiments on the setting of large-scale near-OOD benchmarks organized by OpenOOD~\citep{zhang2023openood}. The ID dataset used is ImageNet~\citep{deng2009imagenet}, while the OOD datasets are SSB-hard~\citep{vaze2021open} and NINCO~\citep{bitterwolf2023or}. Results in the Table~\ref{app-tab: near-ood} show that EOE achieves improvements of $4.25\%$ on average FPR95 and $2.17\%$ on average AUROC compared to MCM.

\subsection{Robustness}
\label{app: robustness}
To explore the OOD detection performance of data from different domains under the same ID class label, we conducted experiments on ImageNet-Sketch\footnote{\url{https://github.com/HaohanWang/ImageNet-Sketch}}~\citep{wang2019learning} and ImageNet-C\footnote{\url{https://zenodo.org/records/2235448}}~\citep{hendrycks2019benchmarking}. Both ImageNet-Sketch and ImageNet-C contain the same 1000 classes as ImageNet. ImageNet-Sketch was constructed through Google Image queries `sketch of \textit{class\_name}', and the images are all black and white. ImageNet-C includes five major types of corruption (Noise, Blur, Weather, Digital, Extra) with a total of 19 types, and it has five levels of severity. We report the results for level-1. Table~\ref{tab: ImageNet-Sketch} shows the experimental results with ImageNet-Sketch as the ID dataset. 

Since ImageNet-Sketch shares the same ID class labels as ImageNet, EOE uses the same outlier class labels as ImageNet. Experiments show that the performance of MCM on ImageNet-Sketch deteriorates significantly, while our method still maintains good performance. \textit{Compared to MCM, EOE achieves improvements of $27.19\%$ in terms of FPR95.} Table~\ref{tab: ImageNet-C-Noise} to \ref{tab: ImageNet-C-Extra} shows the performance of EOE under different types of corruption. Our EOE still significantly outperforms the compared methods. These results on ImageNet-Sketch and ImageNet-C indicate that our EOE exhibits stronger robustness against domain shifts compared to other methods.

\begin{table}[htbp!]
\caption{Robustness results on ImageNet-Sketch dataset. The ID class labels are the same as ImageNet. The \textbf{black bold} indicates the best performance.} %
\label{tab: ImageNet-Sketch}
\centering
\fontsize{7}{8}\selectfont
\setlength\tabcolsep{4pt}
\begin{tabular}{ccccccccccc}
\toprule
\multirow{3}{*}{\textbf{Method}} & \multicolumn{8}{c}{\textbf{OOD Dataset}}                        & \multicolumn{2}{c}{\multirow{2}{*}{\textbf{Average} }} \\
                        & \multicolumn{2}{c}{iNaturalist} & \multicolumn{2}{c}{SUN} & \multicolumn{2}{c}{Places} & \multicolumn{2}{c}{Texture} & \multicolumn{2}{c}{}             \\
                        \cmidrule(lr){2-3}\cmidrule(lr){4-5}\cmidrule(lr){6-7}\cmidrule(lr){8-9}\cmidrule(lr){10-11}
                        & \textbf{FPR95$\downarrow$}         & \textbf{AUROC$\uparrow$}      & \textbf{FPR95$\downarrow$}           & \textbf{AUROC$\uparrow$}         & \textbf{FPR95$\downarrow$}          & \textbf{AUROC$\uparrow$}        & \textbf{FPR95$\downarrow$}         & \textbf{AUROC$\uparrow$}      & \textbf{FPR95$\downarrow$}          & \textbf{AUROC$\uparrow$}        \\  
                        \midrule
Energy
& 82.16 & 81.62 & 79.72 & 80.51 & 75.71 & 79.32 & 93.95 & 56.32 & 82.88 & 74.44\\
MaxLogit
& 70.90 & 84.44 & 72.90 & 81.91 & 70.69 & 79.91 & 89.80 & 60.77 & 76.07 & 76.76\\
MCM 
& 62.88 & 88.08 & 68.55 & 85.01 & 71.25 & 81.01 & 77.84 & 75.35 & 70.13 & 82.36\\
EOE~(Ours)
& 23.97 & 95.10 & 33.18 & 92.45 & 43.62 & 88.49 & 71.00 & 77.82 & \textbf{42.94} & \textbf{88.47}\\
\bottomrule
\end{tabular}
\end{table}

\begin{table}[htbp!]
\caption{Robustness results on ImageNet-C dataset~(corruption type: Noise). The ID class labels are the same as ImageNet. The \textbf{black bold} indicates the best performance.}
\label{tab: ImageNet-C-Noise}
\centering
\fontsize{7}{8}\selectfont
\setlength\tabcolsep{4pt}
\begin{tabular}{ccccccccccccc}
\toprule
\multirow{3}{*}{\textbf{ID Dataset}} & \multirow{3}{*}{\textbf{Method}} & \multicolumn{8}{c}{\textbf{OOD Dataset}} & \multicolumn{2}{c}{\multirow{2}{*}{\textbf{Average}}} \\
& & \multicolumn{2}{c}{iNaturalist} & \multicolumn{2}{c}{SUN} & \multicolumn{2}{c}{Places} & \multicolumn{2}{c}{Texture} & \multicolumn{2}{c}{} \\            
\cmidrule(lr){3-4}\cmidrule(lr){5-6}\cmidrule(lr){7-8}\cmidrule(lr){9-10}\cmidrule(lr){11-12} 
& & \textbf{FPR95$\downarrow$} & \textbf{AUROC$\uparrow$} & \textbf{FPR95$\downarrow$} & \textbf{AUROC$\uparrow$} & \textbf{FPR95$\downarrow$} & \textbf{AUROC$\uparrow$} & \textbf{FPR95$\downarrow$} & \textbf{AUROC$\uparrow$} & \textbf{FPR95$\downarrow$} & \textbf{AUROC$\uparrow$} \\    
\midrule
\multirow{4}{*}{\textbf{Gaussian}} 
& Energy & 88.40 & 81.17 & 84.96 & 80.24 & 80.83 & 79.27 & 95.62 & 58.28 & 87.45 & 74.74 \\
& MaxLogit & 75.17 & 85.06 & 76.64 & 82.80 & 73.77 & 81.06 & 91.17 & 63.76 & 79.19 & 78.17 \\
& MCM & 64.26 & 88.04 & 69.88 & 85.12 & 72.31 & 81.42 & 78.74 & 76.11 & 71.30 & 82.67 \\
& EOE~(Ours) & 27.16 & 94.91 & 35.87 & 92.27 & 46.16 & 88.30 & 73.17 & 77.63 & \textbf{45.59} & \textbf{88.28} \\
\midrule
\multirow{4}{*}{\textbf{Shot}} 
& Energy & 87.40 & 81.18 & 84.14 & 80.20 & 80.03 & 79.18 & 95.32 & 57.70 & 86.72 & 74.56 \\
& MaxLogit & 74.53 & 85.00 & 76.08 & 82.70 & 73.32 & 80.89 & 90.98 & 63.22 & 78.73 & 77.95 \\
& MCM & 66.25 & 87.54 & 71.72 & 84.55 & 73.69 & 80.78 & 80.00 & 75.36 & 72.91 & 82.06 \\
& EOE~(Ours) & 28.99 & 94.62 & 37.47 & 91.89 & 47.88 & 87.81 & 74.38 & 76.83 & \textbf{47.18} & \textbf{87.79} \\
\midrule
\multirow{4}{*}{\textbf{Impulse}} 
& Energy & 87.47 & 78.07 & 84.18 & 76.95 & 80.10 & 75.86 & 95.35 & 51.57 & 86.78 & 70.61 \\
& MaxLogit & 78.82 & 81.37 & 79.54 & 78.74 & 76.36 & 76.79 & 92.02 & 56.79 & 81.69 & 73.42 \\
& MCM & 76.85 & 83.71 & 80.38 & 80.31 & 80.94 & 76.19 & 85.07 & 70.04 & 80.81 & 77.56 \\
& EOE~(Ours) & 44.44 & 91.53 & 51.27 & 88.08 & 60.64 & 83.00 & 83.27 & 69.31 & \textbf{59.90} & \textbf{82.98} \\
\midrule
\multirow{4}{*}{\textbf{Average}} 
& Energy  & 87.76 & 80.14 & 84.43 & 79.13 & 80.32 & 78.10 & 95.43 & 55.85 & 86.98 & 73.30 \\
& MaxLogit  & 76.17 & 83.81 & 77.42 & 81.41 & 74.48 & 79.58 & 91.39 & 61.26 & 79.87 & 76.51 \\
& MCM  & 69.12 & 86.43 & 73.99 & 83.33 & 75.65 & 79.46 & 81.27 & 73.84 & 75.01 & 80.76 \\
& EOE~(Ours)  & 33.53 & 93.69 & 41.54 & 90.74 & 51.56 & 86.37 & 76.94 & 74.59 & \textbf{50.89} & \textbf{86.35} \\
\bottomrule
\end{tabular}
\end{table}

\begin{table}[htbp!]
\caption{Robustness results on ImageNet-C dataset~(corruption type: Blur). The ID class labels are the same as ImageNet. The \textbf{black bold} indicates the best performance.}
\label{tab: ImageNet-C-Blur}
\centering
\fontsize{7}{8}\selectfont
\setlength\tabcolsep{4pt}
\begin{tabular}{ccccccccccccc}
\toprule
\multirow{3}{*}{\textbf{ID Dataset}} & \multirow{3}{*}{\textbf{Method}} & \multicolumn{8}{c}{\textbf{OOD Dataset}} & \multicolumn{2}{c}{\multirow{2}{*}{\textbf{Average}}} \\
& & \multicolumn{2}{c}{iNaturalist} & \multicolumn{2}{c}{SUN} & \multicolumn{2}{c}{Places} & \multicolumn{2}{c}{Texture} & \multicolumn{2}{c}{} \\            
\cmidrule(lr){3-4}\cmidrule(lr){5-6}\cmidrule(lr){7-8}\cmidrule(lr){9-10}\cmidrule(lr){11-12} 
& & \textbf{FPR95$\downarrow$} & \textbf{AUROC$\uparrow$} & \textbf{FPR95$\downarrow$} & \textbf{AUROC$\uparrow$} & \textbf{FPR95$\downarrow$} & \textbf{AUROC$\uparrow$} & \textbf{FPR95$\downarrow$} & \textbf{AUROC$\uparrow$} & \textbf{FPR95$\downarrow$} & \textbf{AUROC$\uparrow$} \\    
\midrule
\multirow{4}{*}{\textbf{Defocus}} 
& Energy & 69.35 & 85.40 & 69.09 & 84.28 & 66.54 & 83.02 & 90.07 & 61.20 & 73.76 & 78.47 \\
& MaxLogit & 63.92 & 87.03 & 66.39 & 84.68 & 65.32 & 82.75 & 87.36 & 64.89 & 70.75 & 79.84 \\
& MCM & 84.88 & 81.53 & 87.26 & 78.01 & 87.12 & 73.73 & 89.13 & 67.32 & 87.10 & 75.15 \\
& EOE~(Ours) & 43.67 & 91.96 & 50.61 & 88.58 & 60.05 & 83.60 & 82.97 & 70.16 & \textbf{59.33} & \textbf{83.58} \\
\midrule
\multirow{4}{*}{\textbf{Glass}} 
& Energy & 66.72 & 86.25 & 67.13 & 85.13 & 64.68 & 83.80 & 89.38 & 61.95 & 71.98 & 79.28 \\
& MaxLogit & 61.03 & 87.72 & 63.81 & 85.35 & 63.19 & 83.36 & 86.40 & 65.39 & 68.61 & 80.45 \\
& MCM & 83.00 & 82.02 & 85.79 & 78.54 & 85.84 & 74.32 & 88.32 & 68.00 & 85.74 & 75.72 \\
& EOE~(Ours) & 42.28 & 92.04 & 49.61 & 88.60 & 59.16 & 83.60 & 82.34 & 70.10 & \textbf{58.35} & \textbf{83.58} \\
\midrule
\multirow{4}{*}{\textbf{Motion}} 
& Energy & 70.88 & 85.52 & 70.69 & 84.45 & 67.60 & 83.25 & 90.73 & 62.16 & 74.97 & 78.84 \\
& MaxLogit & 61.13 & 88.08 & 63.90 & 85.84 & 63.29 & 83.99 & 86.49 & 67.02 & 68.70 & 81.23 \\
& MCM & 70.65 & 86.67 & 75.45 & 83.52 & 76.53 & 79.52 & 82.20 & 73.77 & 76.21 & 80.87 \\
& EOE~(Ours) & 31.66 & 94.18 & 39.82 & 91.30 & 50.20 & 86.98 & 76.34 & 75.35 & \textbf{49.50} & \textbf{86.95} \\
\midrule
\multirow{4}{*}{\textbf{Zoom}} 
& Energy & 51.18 & 89.51 & 53.74 & 88.40 & 52.86 & 86.97 & 84.08 & 66.05 & \textbf{60.47} & \textbf{82.73} \\
& MaxLogit & 53.25 & 89.02 & 57.36 & 86.65 & 57.51 & 84.57 & 83.48 & 66.58 & 62.90 & 81.70 \\
& MCM & 92.87 & 75.13 & 93.61 & 71.21 & 93.20 & 66.70 & 93.33 & 59.63 & 93.25 & 68.17 \\
& EOE~(Ours) & 57.37 & 88.59 & 61.96 & 84.58 & 70.10 & 78.76 & 88.85 & 62.96 & 69.57 & 78.72 \\
\midrule
\multirow{4}{*}{\textbf{Average}} 
& Energy  & 64.53 & 86.67 & 65.16 & 85.56 & 62.92 & 84.26 & 88.56 & 62.84 & 70.30 & 79.83 \\
& MaxLogit  & 59.83 & 87.96 & 62.86 & 85.63 & 62.33 & 83.67 & 85.93 & 65.97 & 67.74 & 80.81 \\
& MCM  & 82.85 & 81.34 & 85.53 & 77.82 & 85.67 & 73.57 & 88.24 & 67.18 & 85.57 & 74.98 \\
& EOE~(Ours)  & 43.74 & 91.69 & 50.50 & 88.27 & 59.88 & 83.23 & 82.63 & 69.64 & \textbf{59.19} & \textbf{83.21} \\
\bottomrule
\end{tabular}
\end{table}

\begin{table}[htbp!]
\caption{Robustness results on ImageNet-C dataset~(corruption type: Weather). The ID class labels are the same as ImageNet. The \textbf{black bold} indicates the best performance.}
\label{tab: ImageNet-C-Weather}
\centering
\fontsize{7}{8}\selectfont
\setlength\tabcolsep{4pt}
\begin{tabular}{ccccccccccccc}
\toprule
\multirow{3}{*}{\textbf{ID Dataset}} & \multirow{3}{*}{\textbf{Method}} & \multicolumn{8}{c}{\textbf{OOD Dataset}} & \multicolumn{2}{c}{\multirow{2}{*}{\textbf{Average}}} \\
& & \multicolumn{2}{c}{iNaturalist} & \multicolumn{2}{c}{SUN} & \multicolumn{2}{c}{Places} & \multicolumn{2}{c}{Texture} & \multicolumn{2}{c}{} \\            
\cmidrule(lr){3-4}\cmidrule(lr){5-6}\cmidrule(lr){7-8}\cmidrule(lr){9-10}\cmidrule(lr){11-12} 
& & \textbf{FPR95$\downarrow$} & \textbf{AUROC$\uparrow$} & \textbf{FPR95$\downarrow$} & \textbf{AUROC$\uparrow$} & \textbf{FPR95$\downarrow$} & \textbf{AUROC$\uparrow$} & \textbf{FPR95$\downarrow$} & \textbf{AUROC$\uparrow$} & \textbf{FPR95$\downarrow$} & \textbf{AUROC$\uparrow$} \\    
\midrule
\multirow{4}{*}{\textbf{Snow}} 
& Energy & 88.06 & 78.76 & 84.71 & 77.69 & 80.60 & 76.67 & 95.55 & 53.58 & 87.23 & 71.67 \\
& MaxLogit & 79.73 & 82.16 & 80.26 & 79.67 & 77.16 & 77.84 & 92.20 & 58.83 & 82.34 & 74.62 \\
& MCM & 77.45 & 84.17 & 80.91 & 80.87 & 81.42 & 76.80 & 85.46 & 70.81 & 81.31 & 78.16 \\
& EOE~(Ours) & 52.85 & 90.16 & 58.33 & 86.76 & 66.92 & 81.63 & 86.93 & 67.72 & \textbf{66.26} & \textbf{81.57} \\
\midrule
\multirow{4}{*}{\textbf{Frost}} 
& Energy & 75.03 & 85.72 & 74.20 & 84.75 & 70.57 & 83.69 & 92.02 & 64.13 & 77.95 & 79.57 \\
& MaxLogit & 63.27 & 88.10 & 65.84 & 85.97 & 64.76 & 84.26 & 87.13 & 68.16 & 70.25 & 81.62 \\
& MCM & 67.69 & 87.56 & 72.84 & 84.65 & 74.60 & 80.96 & 80.73 & 75.67 & 73.97 & 82.21 \\
& EOE~(Ours) & 33.72 & 93.93 & 41.72 & 91.14 & 51.91 & 86.98 & 77.69 & 75.82 & \textbf{51.26} & \textbf{86.97} \\
\midrule
\multirow{4}{*}{\textbf{Fog}} 
& Energy & 72.37 & 85.41 & 71.74 & 84.35 & 68.54 & 83.18 & 91.24 & 62.35 & 75.97 & 78.82 \\
& MaxLogit & 64.20 & 87.67 & 66.70 & 85.45 & 65.51 & 83.63 & 87.41 & 66.74 & 70.96 & 80.87 \\
& MCM & 71.89 & 86.20 & 76.22 & 83.03 & 77.37 & 79.01 & 82.73 & 73.22 & 77.05 & 80.37 \\
& EOE~(Ours) & 36.54 & 93.31 & 44.19 & 90.31 & 54.23 & 85.80 & 79.26 & 73.64 & \textbf{53.56} & \textbf{85.77} \\
\midrule
\multirow{4}{*}{\textbf{Brightness}} 
& Energy & 85.99 & 82.73 & 82.96 & 81.83 & 78.77 & 80.91 & 94.89 & 61.24 & 85.65 & 76.68 \\
& MaxLogit & 68.47 & 87.35 & 70.72 & 85.28 & 69.10 & 83.65 & 88.94 & 67.83 & 74.31 & 81.03 \\
& MCM & 41.83 & 92.69 & 47.88 & 90.34 & 54.15 & 87.18 & 65.48 & 82.93 & 52.34 & 88.29 \\
& EOE~(Ours) & 16.72 & 96.68 & 25.48 & 94.61 & 35.70 & 91.42 & 63.57 & 82.96 & \textbf{35.37} & \textbf{91.42} \\
\midrule
\multirow{4}{*}{\textbf{Average}} 
& Energy  & 80.36 & 83.16 & 78.40 & 82.16 & 74.62 & 81.11 & 93.42 & 60.33 & 81.70 & 76.69 \\
& MaxLogit  & 68.92 & 86.32 & 70.88 & 84.09 & 69.13 & 82.34 & 88.92 & 65.39 & 74.47 & 79.53 \\
& MCM  & 64.72 & 87.66 & 69.46 & 84.72 & 71.89 & 80.99 & 78.60 & 75.66 & 71.17 & 82.26 \\
& EOE~(Ours)  & 34.96 & 93.52 & 42.43 & 90.71 & 52.19 & 86.46 & 76.86 & 75.03 & \textbf{51.61} & \textbf{86.43} \\
\bottomrule
\end{tabular}
\end{table}

\begin{table}[htbp!]
\caption{Robustness results on ImageNet-C dataset~(corruption type: Digital). The ID class labels are the same as ImageNet-1K. The \textbf{black bold} indicates the best performance.}
\label{tab: ImageNet-C-Digital}
\centering
\fontsize{7}{8}\selectfont
\setlength\tabcolsep{4pt}
\begin{tabular}{ccccccccccccc}
\toprule
\multirow{3}{*}{\textbf{ID Dataset}} & \multirow{3}{*}{\textbf{Method}} & \multicolumn{8}{c}{\textbf{OOD Dataset}} & \multicolumn{2}{c}{\multirow{2}{*}{\textbf{Average}}} \\
& & \multicolumn{2}{c}{iNaturalist} & \multicolumn{2}{c}{SUN} & \multicolumn{2}{c}{Places} & \multicolumn{2}{c}{Texture} & \multicolumn{2}{c}{} \\            
\cmidrule(lr){3-4}\cmidrule(lr){5-6}\cmidrule(lr){7-8}\cmidrule(lr){9-10}\cmidrule(lr){11-12} 
& & \textbf{FPR95$\downarrow$} & \textbf{AUROC$\uparrow$} & \textbf{FPR95$\downarrow$} & \textbf{AUROC$\uparrow$} & \textbf{FPR95$\downarrow$} & \textbf{AUROC$\uparrow$} & \textbf{FPR95$\downarrow$} & \textbf{AUROC$\uparrow$} & \textbf{FPR95$\downarrow$} & \textbf{AUROC$\uparrow$} \\    
\midrule
\multirow{4}{*}{\textbf{Contrast}} 
& Energy & 79.17 & 83.57 & 77.40 & 82.54 & 73.43 & 81.45 & 93.09 & 60.43 & 80.77 & 77.00 \\
& MaxLogit & 68.30 & 86.74 & 70.42 & 84.51 & 68.97 & 82.74 & 88.78 & 65.80 & 74.12 & 79.95 \\
& MCM & 71.34 & 86.64 & 75.87 & 83.57 & 76.97 & 79.67 & 82.52 & 74.06 & 76.67 & 80.98 \\
& EOE~(Ours) & 34.05 & 93.83 & 42.05 & 90.94 & 52.17 & 86.60 & 77.87 & 74.93 & \textbf{51.54} & \textbf{86.57} \\
\midrule
\multirow{4}{*}{\textbf{Elastic Transformation}} 
& Energy & 83.45 & 81.09 & 80.84 & 80.02 & 76.80 & 78.95 & 94.33 & 56.59 & 83.86 & 74.16 \\
& MaxLogit & 71.72 & 84.91 & 73.58 & 82.52 & 71.31 & 80.66 & 90.04 & 62.49 & 76.66 & 77.64 \\
& MCM & 68.27 & 87.02 & 73.42 & 83.94 & 75.05 & 80.04 & 81.06 & 74.44 & 74.45 & 81.36 \\
& EOE~(Ours) & 30.94 & 94.24 & 39.27 & 91.36 & 49.68 & 87.06 & 75.80 & 75.51 & \textbf{48.92} & \textbf{87.04} \\
\midrule
\multirow{4}{*}{\textbf{Pixelate}} 
& Energy & 82.32 & 83.51 & 79.97 & 82.54 & 75.90 & 81.50 & 94.04 & 61.07 & 83.06 & 77.16 \\
& MaxLogit & 68.89 & 86.86 & 70.93 & 84.67 & 69.30 & 82.93 & 89.06 & 66.23 & 74.55 & 80.17 \\
& MCM & 63.25 & 88.16 & 68.93 & 85.16 & 71.63 & 81.30 & 78.07 & 75.82 & 70.47 & 82.61 \\
& EOE~(Ours) & 27.72 & 94.78 & 36.40 & 92.06 & 46.74 & 87.97 & 73.63 & 76.97 & \textbf{46.12} & \textbf{87.94} \\
\midrule
\multirow{4}{*}{\textbf{JPEG}} 
& Energy & 87.33 & 82.47 & 84.09 & 81.61 & 79.97 & 80.76 & 95.27 & 61.64 & 86.67 & 76.62 \\
& MaxLogit & 73.30 & 86.47 & 74.87 & 84.42 & 72.17 & 82.86 & 90.55 & 67.18 & 77.72 & 80.23 \\
& MCM & 51.52 & 91.13 & 58.09 & 88.61 & 62.43 & 85.33 & 71.63 & 80.78 & 60.92 & 86.46 \\
& EOE~(Ours) & 20.13 & 96.19 & 28.97 & 94.01 & 39.55 & 90.67 & 67.12 & 81.78 & \textbf{38.94} & \textbf{90.66} \\
\midrule
\multirow{4}{*}{\textbf{Average}} 
& Energy  & 83.07 & 82.66 & 80.58 & 81.68 & 76.53 & 80.67 & 94.18 & 59.93 & 83.59 & 76.23 \\
& MaxLogit  & 70.55 & 86.25 & 72.45 & 84.03 & 70.44 & 82.30 & 89.61 & 65.42 & 75.76 & 79.50 \\
& MCM  & 63.60 & 88.24 & 69.08 & 85.32 & 71.52 & 81.59 & 78.32 & 76.28 & 70.63 & 82.85 \\
& EOE~(Ours)  & 28.21 & 94.76 & 36.67 & 92.09 & 47.03 & 88.07 & 73.61 & 77.30 & \textbf{46.38} & \textbf{88.06} \\
\bottomrule
\end{tabular}
\end{table}

\begin{table}[htbp!]
\caption{Robustness results on ImageNet-C dataset~(corruption type: Extra). The ID class labels are the same as ImageNet-1K. The \textbf{black bold} indicates the best performance.}
\label{tab: ImageNet-C-Extra}
\centering
\fontsize{7}{8}\selectfont
\setlength\tabcolsep{4pt}
\begin{tabular}{ccccccccccccc}
\toprule
\multirow{3}{*}{\textbf{ID Dataset}} & \multirow{3}{*}{\textbf{Method}} & \multicolumn{8}{c}{\textbf{OOD Dataset}} & \multicolumn{2}{c}{\multirow{2}{*}{\textbf{Average}}} \\
& & \multicolumn{2}{c}{iNaturalist} & \multicolumn{2}{c}{SUN} & \multicolumn{2}{c}{Places} & \multicolumn{2}{c}{Texture} & \multicolumn{2}{c}{} \\            
\cmidrule(lr){3-4}\cmidrule(lr){5-6}\cmidrule(lr){7-8}\cmidrule(lr){9-10}\cmidrule(lr){11-12} 
& & \textbf{FPR95$\downarrow$} & \textbf{AUROC$\uparrow$} & \textbf{FPR95$\downarrow$} & \textbf{AUROC$\uparrow$} & \textbf{FPR95$\downarrow$} & \textbf{AUROC$\uparrow$} & \textbf{FPR95$\downarrow$} & \textbf{AUROC$\uparrow$} & \textbf{FPR95$\downarrow$} & \textbf{AUROC$\uparrow$} \\    
\midrule
\multirow{4}{*}{\textbf{Speckle}} 
& Energy & 83.33 & 82.83 & 80.70 & 81.85 & 76.56 & 80.83 & 94.27 & 60.08 & 83.72 & 76.40 \\
& MaxLogit & 69.52 & 86.36 & 71.49 & 84.13 & 69.68 & 82.37 & 89.38 & 65.36 & 75.02 & 79.56 \\
& MCM & 63.90 & 88.09 & 69.46 & 85.15 & 72.05 & 81.41 & 78.51 & 76.08 & 70.98 & 82.68 \\
& EOE~(Ours) & 28.10 & 94.76 & 36.78 & 92.06 & 47.22 & 88.03 & 73.97 & 77.20 & \textbf{46.51} & \textbf{88.02} \\
\midrule
\multirow{4}{*}{\textbf{Gaussian}} 
& Energy & 82.96 & 82.31 & 80.38 & 81.30 & 76.35 & 80.24 & 94.17 & 58.77 & 83.46 & 75.66 \\
& MaxLogit & 70.06 & 86.26 & 72.15 & 83.99 & 70.13 & 82.19 & 89.56 & 64.86 & 75.47 & 79.33 \\
& MCM & 63.56 & 88.30 & 69.20 & 85.32 & 71.85 & 81.48 & 78.28 & 76.05 & 70.72 & 82.79 \\
& EOE~(Ours) & 27.06 & 94.88 & 35.85 & 92.18 & 46.19 & 88.12 & 73.19 & 77.22 & \textbf{45.57} & \textbf{88.10} \\
\midrule
\multirow{4}{*}{\textbf{Spatter}} 
& Energy & 82.96 & 83.08 & 80.40 & 82.13 & 76.39 & 81.16 & 94.17 & 61.06 & 83.48 & 76.86 \\
& MaxLogit & 67.98 & 87.20 & 70.04 & 85.07 & 68.70 & 83.40 & 88.65 & 67.21 & 73.84 & 80.72 \\
& MCM & 48.49 & 91.36 & 55.08 & 88.77 & 59.96 & 85.33 & 70.07 & 80.61 & 58.40 & 86.52 \\
& EOE~(Ours) & 22.03 & 95.77 & 30.79 & 93.40 & 41.33 & 89.77 & 68.85 & 80.07 & \textbf{40.75} & \textbf{89.76} \\
\midrule
\multirow{4}{*}{\textbf{Saturate}} 
& Energy & 75.42 & 85.64 & 74.64 & 84.69 & 70.84 & 83.64 & 92.15 & 64.26 & 78.26 & 79.56 \\
& MaxLogit & 62.18 & 88.80 & 64.92 & 86.76 & 64.00 & 85.09 & 86.81 & 69.53 & 69.48 & 82.54 \\
& MCM & 52.30 & 90.68 & 59.00 & 88.00 & 63.21 & 84.45 & 72.16 & 79.57 & 61.67 & 85.67 \\
& EOE~(Ours) & 22.27 & 95.75 & 31.26 & 93.36 & 41.81 & 89.70 & 69.21 & 79.96 & \textbf{41.14} & \textbf{89.69} \\
\midrule
\multirow{4}{*}{\textbf{Average}} 
& Energy  & 81.17 & 83.46 & 79.03 & 82.49 & 75.03 & 81.47 & 93.69 & 61.04 & 82.23 & 77.12 \\
& MaxLogit  & 67.44 & 87.16 & 69.65 & 84.99 & 68.13 & 83.26 & 88.60 & 66.74 & 73.45 & 80.54 \\
& MCM  & 57.06 & 89.61 & 63.18 & 86.81 & 66.77 & 83.17 & 74.75 & 78.08 & 65.44 & 84.42 \\
& EOE~(Ours)  & 24.86 & 95.29 & 33.67 & 92.75 & 44.14 & 88.91 & 71.30 & 78.61 & \textbf{43.49} & \textbf{88.89} \\
\bottomrule
\end{tabular}
\end{table}

\subsection{Prompt Ensembling for Text Input}
\label{app: prompt-engineering}
To investigate the effect of prompt ensembling for CLIP text input on EOE, we conduct experiments using five types of prompts for CLIP text input from MCM under different vision encoders. The five prompts are shown in Figure~\ref{app dataset: prompt-engineering-set}. Table~\ref{tab: prompt-engineering} presents the results of prompt ensembling under different vision encoders. \textbf{Interestingly, only our EOE achieves gains with the prompt ensembling strategy across different vision encoders.} In contrast, MCM exhibits varying degrees of performance degradation under ViT-B/16, ViT-B/32, RN50, and RN50x16.

\begin{figure}[htbp!]
\centering
\begin{tikzpicture}
\small
\definecolor{chatcolor1}{HTML}{5fedb7} 
\definecolor{shadecolor}{gray}{0.9}
\fontfamily{cmss}\selectfont

\node[align=left, text width=0.31\textwidth, fill=shadecolor, rounded corners=1mm, anchor=north west] (node1) at ([yshift=-0.4cm]node1.south west |- node2.south) {
\textbf{Prompt set}\\
\begin{tabular}{@{}l}
- a photo of a <label>.\\
- a blurry photo of a <label>.\\
- a photo of many <label>.\\
- a photo of the large <label>.\\
- a photo of the small <label>.\\
\end{tabular}
};
\node[draw, black, thick, rounded corners=3mm, inner sep=5pt, fit=(node1)] {};
\end{tikzpicture}
\caption{Prompt set for text encoder input. <label> is the ID class name, such as "\textit{horse}". The prompt set is extracted from CLIP~\citep{radford2021learning} by MCM~\citep{ming2022delving}.}
\label{app dataset: prompt-engineering-set}
\end{figure}

\begin{table}[htbp!]
\caption{Prompt ensembling for text input using different backbones. The ID dataset is ImageNet-1K. The "(+/- $\times \times$)" in the Average column indicates the difference compared to the text prompt of "\textit{a photo of a <label>.}". Enhancements are marked in \textcolor{teal}{green}, while performance degradation is indicated in \textcolor{red}{red}. Only our EOE achieves gains using prompt ensembling strategies across different backbones.} %
\label{tab: prompt-engineering}
\centering
\fontsize{7}{8}\selectfont
\setlength\tabcolsep{4pt}
\begin{tabular}{ccccccccccc}
\toprule
\multirow{3}{*}{\textbf{Method}} & \multicolumn{8}{c}{\textbf{OOD Dataset}}                        & \multicolumn{2}{c}{\multirow{2}{*}{\textbf{Average} }} \\
                        & \multicolumn{2}{c}{iNaturalist} & \multicolumn{2}{c}{SUN} & \multicolumn{2}{c}{Places} & \multicolumn{2}{c}{Texture} & \multicolumn{2}{c}{}             \\
                        \cmidrule(lr){2-3}\cmidrule(lr){4-5}\cmidrule(lr){6-7}\cmidrule(lr){8-9}\cmidrule(lr){10-11}
                        & \textbf{FPR95$\downarrow$}         & \textbf{AUROC$\uparrow$}      & \textbf{FPR95$\downarrow$}           & \textbf{AUROC$\uparrow$}         & \textbf{FPR95$\downarrow$}          & \textbf{AUROC$\uparrow$}        & \textbf{FPR95$\downarrow$}         & \textbf{AUROC$\uparrow$}      & \textbf{FPR95$\downarrow$}          & \textbf{AUROC$\uparrow$}        \\  
                        \midrule
Energy (ViT-B/16)
& 79.75 & 83.75 & 79.81 & 83.21 & 70.28 & 83.95 & 88.23 & 71.51 & 79.52~(\textcolor{teal}{+2.69}) & 80.60~(\textcolor{teal}{+1.03})\\
MaxLogit (ViT-B/16)
& 67.24 & 87.31 & 66.14 & 86.36 & 61.09 & 85.96 & 80.83 & 76.01 & 68.83~(\textcolor{teal}{+0.25}) & 83.91~(\textcolor{teal}{+0.32})\\
MCM (ViT-B/16)
&40.33&92.75&35.43&92.78&44.08&89.60&54.41&87.10&43.56~(\textcolor{red}{-0.79})&90.56~(\textcolor{red}{-0.21})\\
EOE~(Ours) (ViT-B/16)
& 15.24 & 96.86 & 18.62 & 95.99 & 28.75 & 93.03 & 53.95 & 87.15 &\textbf{29.14}~(\textcolor{teal}{+0.95})&  \textbf{93.26}~(\textcolor{teal}{+0.30})\\
\midrule
Energy (ViT-B/32)
& 89.22 & 79.15 & 81.01 & 81.62 & 61.22 & 87.20 & 87.64 & 71.36 & 79.77~(\textcolor{red}{-1.87}) & 79.83~(\textcolor{red}{-0.31})\\
MaxLogit (ViT-B/32)
& 79.45 & 83.75 & 68.89 & 84.85 & 52.30 & 88.60 & 79.88 & 75.29 & 70.13~(\textcolor{red}{-3.17}) & 83.12~(\textcolor{red}{-0.42})\\
MCM (ViT-B/32)
& 49.81 & 91.37 & 40.31 & 91.80 & 42.94 & 90.08 & 59.33 & 85.32 & 48.10~(\textcolor{red}{-2.33}) & 89.64~(\textcolor{red}{-0.28})\\
EOE~(Ours) (ViT-B/32)
& 20.33 & 96.16 & 20.52 & 95.64 & 28.25 & 93.31 & 57.71 & 85.49 & \textbf{31.71}~(\textcolor{teal}{+0.01}) & \textbf{92.65}~(\textcolor{teal}{+0.15})\\
\midrule
Energy (ViT-L/14)
& 79.20 & 85.29 & 76.83 & 84.68 & 65.62 & 87.59 & 87.23 & 70.14 & 77.22~(\textcolor{teal}{+3.26}) & 81.93~(\textcolor{teal}{+2.04})\\
MaxLogit (ViT-L/14)
& 63.06 & 89.02 & 60.26 & 88.29 & 52.51 & 89.65 & 80.66 & 73.96 & 64.12~(\textcolor{teal}{+3.08})&85.23~(\textcolor{teal}{+1.15})\\
MCM (ViT-L/14) 
& 31.63~ & 94.43 & 23.64 & 94.99 & 30.99 & 92.79 & 57.77 & 85.19 & 36.01(\textcolor{teal}{+2.09}) & 91.85~(\textcolor{teal}{+0.36})\\
EOE~(Ours) (ViT-L/14)
& 13.26 & 97.43 & 14.51 & 97.01 & 22.35 & 94.79 & 56.77 & 85.27 & \textbf{26.72}~(\textcolor{teal}{+1.63}) & \textbf{93.62}~(\textcolor{teal}{+0.37})\\
\midrule
Energy (RN50) & 94.75 & 75.56 & 86.24 & 81.39 & 86.42 & 78.68 & 92.98 & 69.87 & 90.10(\textcolor{red}{-1.26}) & 76.38~(\textcolor{red}{-0.40}) \\
MaxLogit (RN50) & 86.45 & 81.21 & 74.56 & 84.31 & 78.15 & 81.10 & 86.45 & 74.61 & 81.40~(\textcolor{red}{-2.22}) & 80.31~(\textcolor{red}{-0.66})\\
MCM (RN50) & 45.42 & 91.50 & 43.33 & 91.40 & 55.92 & 86.73 & 55.92 & 86.68 & 50.15~(\textcolor{red}{-0.43}) & 89.08~(\textcolor{teal}{+0.09}) \\
EOE~(Ours) (RN50) & 20.45 & 96.05 & 22.58 & 95.31 & 35.32 & 91.33 & 54.82 & 86.81 & \textbf{33.30}~(\textcolor{teal}{+1.18}) & \textbf{92.38}~(\textcolor{teal}{+0.39}) \\
\midrule
Energy (RN50x4) & 85.55 & 81.25 & 80.13 & 84.81 & 68.84 & 85.40 & 92.09 & 69.28 & 81.65~(\textcolor{teal}{+1.23}) & 80.19~(\textcolor{teal}{+1.35}) \\
MaxLogit (RN50x4) & 74.51 & 85.14 & 65.51 & 87.61 & 58.86 & 87.26 & 84.47 & 74.81 & 70.84~(\textcolor{teal}{+0.03}) & 83.70~(\textcolor{teal}{+0.49}) \\
MCM (RN50x4) & 48.00 & 90.86 & 33.81 & 93.14 & 42.90 & 89.93 & 52.16 & 87.44 & 44.22~(\textcolor{teal}{+1.12}) & 90.34~(\textcolor{teal}{+0.35}) \\
EOE~(Ours) (RN50x4) & 25.65 & 95.19 & 20.13 & 95.90 & 30.18 & 92.77 & 52.03 & 87.55 & \textbf{32.00}~(\textcolor{teal}{+2.56}) & \textbf{92.85}~(\textcolor{teal}{+0.60}) \\
\midrule
Energy (RN50x16) & 73.44 & 86.95 & 65.15 & 88.97 & 73.74 & 83.97 & 84.43 & 76.11 & 74.19(\textcolor{red}{-0.50}) & 84.00~(\textcolor{teal}{+0.64}) \\
MaxLogit (RN50x16) & 62.10 & 89.05 & 52.35 & 90.45 & 64.74 & 85.69 & 75.66 & 79.37 & 63.71~(\textcolor{red}{-2.06}) & 86.14~(\textcolor{red}{-0.05}) \\
MCM (RN50x16) & 43.02 & 91.69 & 34.24 & 93.27 & 46.96 & 89.27 & 51.93 & 87.94 & 44.04~(\textcolor{teal}{+1.14}) & 90.54~(\textcolor{teal}{+0.53}) \\
EOE~(Ours) (RN50x16) & 23.03 & 95.45 & 20.00 & 95.92 & 31.94 & 92.52 & 52.54 & 87.65 & \textbf{31.88}~(\textcolor{teal}{+2.52}) & \textbf{92.89}~(\textcolor{teal}{+0.73}) \\
\midrule
Energy (RN101) & 97.82 & 71.11 & 87.81 & 81.10 & 85.43 & 77.92 & 95.96 & 62.32 & 91.75~(\textcolor{red}{-0.00}) & 73.11(\textcolor{red}{-0.16}) \\
MaxLogit (RN101) & 92.65 & 77.38 & 74.77 & 84.67 & 75.96 & 81.30 & 90.90 & 68.66 & 83.57~(\textcolor{red}{-1.72}) & 78.00~(\textcolor{red}{-0.61}) \\
MCM (RN101) & 60.90 & 88.14 & 39.37 & 91.96 & 48.62 & 88.08 & 59.49 & 85.34 & 52.09~(\textcolor{red}{-1.44}) & 88.38~(\textcolor{red}{-0.11}) \\
EOE~(Ours) (RN101) & 33.46 & 93.86 & 23.25 & 95.19 & 33.48 & 91.62 & 58.53 & 85.50 & \textbf{37.18}~(\textcolor{teal}{+1.10}) & \textbf{91.54}~(\textcolor{teal}{+0.36}) \\
\bottomrule
\end{tabular}
\end{table}

\section{Additional Ablation Studies}
\begin{table}[htbp!]
\caption{Additional empirical results with different CLIP vision encoders on ImageNet-1K benchmark. The \textbf{black bold} indicates the best performance. The \colorbox{gray!22}{gray} indicates that the comparative methods require training or an additional massive auxiliary dataset. Energy (FT) requires fine-tuning, while Energy is post-hoc.}%
\label{tab: differ-vision-encoder}
\centering
\fontsize{7}{8}\selectfont
\setlength\tabcolsep{4.5pt}
\begin{tabular}{ccccccccccc}
\toprule
\multirow{3}{*}{\textbf{Method}} & \multicolumn{8}{c}{\textbf{OOD Dataset}}                        & \multicolumn{2}{c}{\multirow{2}{*}{\textbf{Average} }} \\
                        & \multicolumn{2}{c}{iNaturalist} & \multicolumn{2}{c}{SUN} & \multicolumn{2}{c}{Places} & \multicolumn{2}{c}{Texture} & \multicolumn{2}{c}{}             \\
                        \cmidrule(lr){2-3}\cmidrule(lr){4-5}\cmidrule(lr){6-7}\cmidrule(lr){8-9}\cmidrule(lr){10-11}
                        & \textbf{FPR95$\downarrow$}         & \textbf{AUROC$\uparrow$}      & \textbf{FPR95$\downarrow$}           & \textbf{AUROC$\uparrow$}         & \textbf{FPR95$\downarrow$}          & \textbf{AUROC$\uparrow$}        & \textbf{FPR95$\downarrow$}         & \textbf{AUROC$\uparrow$}      & \textbf{FPR95$\downarrow$}          & \textbf{AUROC$\uparrow$}        \\  
                        \midrule
\rowcolor{gray!22}
 Fort et al. (ViT-B/16) & 15.07&	96.64&	54.12&	86.37&	57.99&	85.24&	53.32&	84.77&	45.12&	88.25\\
\rowcolor{gray!22}
Energy (FT) (ViT-B/16) & 21.59&	95.99& 34.28&	93.15 &36.64&	91.82& 51.18&	88.09& 35.92&	92.26 \\ 
\rowcolor{gray!22}
MSP (ViT-B/16)&40.89&	88.63&	65.81&	81.24&	67.90&	80.14&	64.96&	78.16&	59.89&	82.04\\
\rowcolor{gray!22}
CLIPN (ViT-B/16)
& 19.13	& 96.20	& 25.69 & 94.18 & 	32.14& 	92.26& 	44.60 & 88.93 & 30.39 & 	92.89\\
\midrule
\rowcolor{gray!22}
Fort et al. (ViT-L/14)&15.74&	96.51	&52.34&	87.32&	55.14&	86.48&	51.38&	85.54&	43.65&	88.96 \\
\rowcolor{gray!22}
Energy (FT) (ViT-L/14) & 10.62&	97.52 &30.46&	93.83 &32.25&	93.01& 44.35&	89.64 & 29.42&	93.50\\
\rowcolor{gray!22}
MSP (ViT-L/14)&34.54&	92.62&	61.18&	83.68&	59.86&	84.10&	59.27&	82.31&	53.71&	85.68\\
\rowcolor{gray!22}
CLIPN (ViT-L/14) 
& 25.09	& 94.59	& 24.76 & 	94.93& 	30.89 & 93.14 & 48.97 & 87.01 & 32.43 & 92.42\\
\midrule
Energy (ViT-B/16) 
&81.08&85.09&79.02&84.24&75.08&83.38&93.65&65.56&82.21&79.57\\
MaxLogit (ViT-B/16)
& 61.66&89.31&64.39&87.43&63.67&85.95&86.61&71.68&69.08&83.59\\
MCM (ViT-B/16)
& 30.92	& 94.61	& 37.59& 	92.57& 	44.71& 	89.77& 	57.85& 	86.11& 	42.77& 	90.77\\
EOE~(Ours) (ViT-B/16)
& 12.29 & 97.52 & 20.40 & 95.73 & 30.16 & 92.95 & 57.53 & 85.64 & \textbf{30.09} & \textbf{92.96}  \\
\midrule
Energy (ViT-B/32)
& 80.16 & 83.75 & 77.21 & 83.85 & 61.28 & 87.77 & 92.96 & 65.19 & 77.90 & 80.14 \\
MaxLogit (ViT-B/32)
& 64.13 & 87.86 & 65.16 & 86.64 & 51.92 & 89.10 & 86.63 & 70.57 & 66.96 & 83.54\\
MCM (ViT-B/32)
& 33.91 & 93.61 & 41.82 & 91.42 & 45.66 & 89.56 & 61.67 & 84.67 & 45.77 & 89.82\\
EOE~(Ours) (ViT-B/32)
& 14.97 & 97.03 & 21.88 & 95.40 & 29.91 & 92.97 & 60.10 & 84.61 & \textbf{31.72} & \textbf{92.50}\\
\midrule
Energy (ViT-L/14)
& 78.84 & 85.87 & 78.87 & 83.51 & 70.30 & 86.44 & 93.90 & 63.74 & 80.48 & 79.89\\
MaxLogit (ViT-L/14) 
& 58.96 & 90.13 & 63.97 & 87.75 & 57.00 & 89.05 & 88.90 & 69.37 & 67.21 & 84.08\\
MCM (ViT-L/14) 
& 28.35 & 94.95 & 28.93 & 94.14 & 35.34 & 92.00 & 59.79 & 84.88 & 38.10 & 91.49\\
EOE~(Ours) (ViT-L/14)
& 11.79 & 97.63 & 17.14 & 96.46 & 24.93 & 94.23 & 59.55 & 84.69 & \textbf{28.35} & \textbf{93.25}\\
\midrule
Energy (RN50) & 91.53 & 80.26 & 83.70 & 82.78 & 84.64 & 81.30 & 95.51 & 62.76 & 88.84 & 76.78 \\
MaxLogit (RN50) & 76.17 & 85.78 & 73.09 & 85.46 & 76.43 & 83.26 & 91.03 & 69.37 & 79.18 & 80.97 \\
MCM (RN50) & 32.06 & 93.86 & 46.14 & 90.74 & 60.60 & 85.67 & 60.09 & 85.71 & 49.72 & 88.99 \\
EOE~(Ours) (RN50) & 14.44 & 97.05 & 24.85 & 94.82 & 38.89 & 90.64 & 59.73 & 85.44 & \textbf{34.47} & \textbf{91.98} \\
\midrule
Energy (RN50x4) & 83.73 & 83.14 & 79.73 & 84.70 & 71.95 & 85.21 & 96.10 & 62.29 & 82.88 & 78.84 \\
MaxLogit (RN50x4) & 69.80 & 87.02 & 63.44 & 88.27 & 58.87 & 87.66 & 91.37 & 69.89 & 70.87 & 83.21 \\
MCM (RN50x4) & 44.87 & 91.42 & 35.22 & 92.85 & 44.15 & 89.49 & 57.11 & 86.21 & 45.34 & 89.99 \\
EOE~(Ours) (RN50x4) & 26.48 & 95.01 & 21.82 & 95.47 & 32.71 & 92.21 & 57.22 & 86.32 & \textbf{34.56} & \textbf{92.25} \\
\midrule
Energy (RN50x16) & 68.41 & 88.91 & 63.01 & 89.49 & 71.03 & 85.61 & 92.29 & 69.43 & 73.69 & 83.36 \\
MaxLogit (RN50x16) & 50.80 & 91.32 & 49.84 & 91.13 & 61.08 & 87.23 & 84.89 & 75.08 & 61.65 & 86.19 \\
MCM (RN50x16) & 35.98 & 93.05 & 38.20 & 92.30 & 51.33 & 87.89 & 55.20 & 86.80 & 45.18 & 90.01 \\
EOE~(Ours) (RN50x16) & 21.19 & 95.81 & 23.42 & 95.19 & 35.86 & 91.59 & 57.13 & 86.05 & \textbf{34.40} & \textbf{92.16} \\
\midrule
Energy (RN101) & 95.66 & 74.97 & 87.35 & 81.14 & 85.97 & 78.65 & 98.03 & 58.33 & 91.75 & 73.27 \\
MaxLogit (RN101) & 85.97 & 80.62 & 72.23 & 85.35 & 75.19 & 82.37 & 94.02 & 66.09 & 81.85 & 78.61 \\
MCM (RN101) & 51.90 & 89.78 & 40.09 & 91.68 & 50.69 & 87.50 & 59.93 & 85.00 & 50.65 & 88.49 \\
EOE~(Ours) (RN101) & 32.57 & 93.97 & 24.60 & 94.73 & 36.43 & 90.98 & 59.53 & 85.07 & \textbf{38.29} & \textbf{91.19} \\
\bottomrule
\end{tabular}
\end{table}

\begin{table}[htbp!]
\caption{Additional empirical results with different vision-language models with ImageNet-1K as the ID dataset. The \textbf{black bold} indicates the best performance.} %
\label{tab: differ-vlm}
\centering
\fontsize{7}{8}\selectfont
\setlength\tabcolsep{4.5pt}
\begin{tabular}{ccccccccccc}
\toprule
\multirow{3}{*}{\textbf{Method}} & \multicolumn{8}{c}{\textbf{OOD Dataset}}                        & \multicolumn{2}{c}{\multirow{2}{*}{\textbf{Average} }} \\
                        & \multicolumn{2}{c}{iNaturalist} & \multicolumn{2}{c}{SUN} & \multicolumn{2}{c}{Places} & \multicolumn{2}{c}{Texture} & \multicolumn{2}{c}{}             \\
                        \cmidrule(lr){2-3}\cmidrule(lr){4-5}\cmidrule(lr){6-7}\cmidrule(lr){8-9}\cmidrule(lr){10-11}
                        & \textbf{FPR95$\downarrow$}         & \textbf{AUROC$\uparrow$}      & \textbf{FPR95$\downarrow$}           & \textbf{AUROC$\uparrow$}         & \textbf{FPR95$\downarrow$}          & \textbf{AUROC$\uparrow$}        & \textbf{FPR95$\downarrow$}         & \textbf{AUROC$\uparrow$}      & \textbf{FPR95$\downarrow$}          & \textbf{AUROC$\uparrow$}        \\  
                        \midrule
Energy (GroupViT)
& 85.33 & 79.29 & 83.49 & 79.78 & 80.05 & 77.79 & 93.72 & 59.34 & 85.65 & 74.05\\
MaxLogit (GroupViT)
& 81.80 & 79.60 & 83.01 & 79.03 & 80.42 & 76.91 & 93.49 & 59.78 & 84.68 & 73.83\\
MCM (GroupViT)  
& 53.20 & 88.33 & 60.30 & 84.77 & 66.48 & 80.93 & 61.52 & 81.50 & 60.38 & 83.88\\
EOE~(Ours) (GroupViT)
& 20.48 & 95.75 & 28.52 & 93.13 & 38.26 & 89.27 & 61.37 & 81.44 & \textbf{37.16} & \textbf{89.90}\\
\midrule
Energy (AltCLIP) 
& 58.47 & 91.57 & 50.29 & 92.29 & 51.29 & 90.51 & 94.73 & 64.36 & 63.69 & 84.68\\
MaxLogit (AltCLIP)  
& 49.81 & 92.35 & 41.29 & 93.11 & 45.73 & 91.04 & 89.50 & 68.98 & 56.58 & 86.37\\
MCM (AltCLIP)  
& 43.55 & 92.88 & 28.39 & 94.56 & 38.22 & 91.57 & 53.58 & 87.00 & 40.94 & 91.50\\
EOE~(Ours) (AltCLIP)
& 18.35 & 96.75 & 15.50 & 97.05 & 25.38 & 94.34 & 54.68 & 86.36 & \textbf{28.48} & \textbf{93.63}\\
\midrule
Energy (ALIGN) 
& 92.18 & 83.62 & 74.14 & 86.28 & 70.48 & 84.64 & 68.65 & 81.76 & 76.36 & 84.08\\
MaxLogit (ALIGN)  
& 83.47 & 84.95 & 68.46 & 86.63 & 67.08 & 84.77 & 65.32 & 81.89 & 71.08 & 84.56\\
MCM (ALIGN)  
& 60.63 & 89.39 & 53.20 & 89.17 & 61.05 & 85.58 & 59.63 & 83.94 & 58.63 & 87.02\\
EOE~(Ours) (ALIGN)
& 25.32 & 95.45 & 26.67 & 94.50 & 38.57 & 90.95 & 57.32 & 84.41 & \textbf{36.97} & \textbf{91.33}\\
\bottomrule
\end{tabular}
\end{table}

\subsection{VLM Backbones}
\label{app: differ-vlm}
In this section, we conduct experiments with different CLIP vision encoders to investigate the performance of EOE. Furthermore, we also report additional experimental results beyond CLIP models.

Table~\ref{tab: differ-vision-encoder} shows the performance of ImageNet-1K(ID) with different CLIP vision encoders, including ViT-B/32\footnote{\url{https://huggingface.co/openai/clip-vit-base-patch32}}, ViT-L/14\footnote{\url{https://huggingface.co/openai/clip-vit-large-patch14}}, RN50\footnote{\url{https://github.com/openai/CLIP}}, RN50x4, RN50x16 and RN101. Our EOE achieves the best results across all CLIP vision encoders. Compared with ViT-B/16, EOE yields an enhancement of $1.74\%$ and $0.29\%$ in FPR95 and AUROC based on ViT-L/14, respectively. Moreover, DOC achieves the best OOD detection performance compared to both zero-shot methods and fine-tuning methods in terms of FPR95 based on ViT-L/14. It should be noted that CLIPN achieves worse performance when using the VIT-L/14 as the backbone than when using VIT-B/16. Instead, our EOE is more generalizable to different backbones and produces clearly better OOD performance than CLIPN when using VIT-L/14 as the backbone. Compared to other zero-shot methods, our EOE also achieved the best results on the ResNet backbone. The results of fine-tuning methods are reported by MCM~\citep{ming2022delving}. Due to limitations in computational resources, we do not report the results of other fine-tuning methods on the ResNet backbone here.

The results for other vision-language models are shown in Table~\ref{tab: differ-vlm}. We perform experiments with GroupViT\footnote{\url{https://huggingface.co/nvidia/groupvit-gcc-yfcc}}~\citep{xu2022groupvit}, AltCLIP\footnote{\url{https://huggingface.co/BAAI/AltCLIP}}~\citep{chen2022altclip} and ALIGN\footnote{\url{https://huggingface.co/kakaobrain/align-base}}~\citep{jia2021scaling}.  
When using GroupViT as the backbone, our EOE’s performance~(\textbf{$37.16\%$} in terms of FPR95) is significantly better than MCM's~(\textbf{$60.38\%$} in terms of FPR95). These results indicate that our EOE is more generalizable to different VLMs.

\subsection{Score Functions}\label{app: score function}
Here, we present the specific form of the score function designed in the ablation study. They are $S_{\text{MAX}}$, $S_{\text{MSP}}$, $S_{\text{Energy}}$ and $S_{\text{MaxLogit}}$.
First, we review the definition of label-wise matching score $s_i(x)$:
\begin{equation}
    s_i(x) = \frac{\mathcal{I}(x) \cdot \mathcal{T}(t_i)}{\lVert \mathcal{I}(x)\rVert \cdot \lVert \mathcal{T}(t_i) \rVert};~~~i \in [1, K+L],~t_i \in \mathcal{Y}_\text{id} \cup \mathcal{Y}_\text{outlier}.
\end{equation}
The specific form of $S_{\text{MAX}}$ is as follows:
\begin{equation}
    S_{\text{MAX}}
    (x;\mathcal{Y}_\text{id},\mathcal{Y}_\text{outlier},\mathcal{T},\mathcal{I}) =\begin{cases} 
      \frac{1}{K} &  \underset{i \in [1, K]}\max s_i < \underset{j \in [K+1, L]}\max s_j\\
      \underset{{i \in [1, K]}}\max \frac{e^{s_i(x)}}{\sum_{j=1}^{K} e^{s_j(x)}}& \underset{i \in [1, K]}\max s_i \ge \underset{{j \in [K+1, L]}}\max s_j
   \end{cases},
\end{equation}
$S_{\text{MAX}}$ indicates that if the $s_j~(j \in [K+1, L])$ of an input sample is larger than the $s_k~(j \in [1, K])$, this sample is recognized to be an OOD sample. This means that the highest degree of similarity observed between the input sample and the outlier label exceeds that between the input sample and any in-distribution (ID) class label. Otherwise, the input sample is calculated according to maximum softmax probabilities~(MSP). 

$S_{\text{MSP}}$ is an adaptation of MSP, as defined in Eq.~\ref{eq: s_msp}. Similarly, $S_{\text{Energy}}$ and $S_{\text{MaxLogit}}$ are modifications of the Energy and MaxLogit metrics, respectively, incorporating outlier classes into their secondary components.
\begin{equation}
    S_{\text{Energy}}(x;\mathcal{Y}_\text{id},\mathcal{Y}_\text{outlier},\mathcal{T},\mathcal{I}) = -T \left(\log \sum_{i=1}^K e^{f_i(x) / T} - \log \sum_{j=K+1}^L e^{f_j(x) / T}\right),
\end{equation}

\begin{equation}
    S_{\text{MaxLogit}}(x;\mathcal{Y}_\text{id},\mathcal{Y}_\text{outlier},\mathcal{T},\mathcal{I}) = \underset{\scriptscriptstyle i \in [1, K]}\max s_i(x)- \underset{\scriptscriptstyle j \in [K+1, K+L]}\max s_j(x).
\end{equation}

We conducted additional ablation study experiments for score functions under three more challenging scenarios, including two large-scale benchmark settings and one fine-grained OOD detection setting in Table~\ref{app-tab:score-function}. Specifically, for large-scale benchmarks, we evaluate far OOD detection (ImageNet-1K \textit{vs.} Texture) and near OOD detection (ImageNet-1K \textit{vs.} NINCO, organized by OpenOOD). For fine-grained OOD detection, we evaluate the setting of Food-50 \textit{vs.} Food-51. Our $S_\text{EOE}$ consistently outperform the commonly used scores of $S_\text{MAX}$, $S_\text{Energy}$, and $S_\text{MaxLogit}$. Compared between two variants of our scores,  \textit{i.e.}, $S_\text{EOE}$ and $S_\text{MSP}$, our weighted version ($S_\text{EOE}$, $\beta$=0.25) achieves the best results in most cases and on average.

\begin{table}[t]
\centering
\caption{Additional ablation studies on score functions. The \textbf{bold} indicates the best performance on each dataset.}
\label{app-tab:score-function}
\fontsize{7}{8}\selectfont
\setlength\tabcolsep{4.5pt}
\begin{tabular}{clcccccccc}
\toprule

\multirow{2}{*}{\textbf{Score Funtion}} & \textbf{ID} & \multicolumn{2}{c}{ImageNet-1K} & \multicolumn{2}{c}{ImageNet-1K}& \multicolumn{2}{c}{Food-50} & \multicolumn{2}{c}{\multirow{2}{*}{\textbf{Average}}} \\
 & \textbf{OOD}        &  \multicolumn{2}{c}{Texture} & \multicolumn{2}{c}{NINCO}& \multicolumn{2}{c}{Food-50} \\

\cmidrule(rl){3-4}
\cmidrule(rl){5-6}
\cmidrule(rl){7-8}
\cmidrule(rl){9-10}
&&  \textbf{FPR95$\downarrow$} & \textbf{AUROC$\uparrow$}               
&  \textbf{FPR95$\downarrow$} & \textbf{AUROC$\uparrow$}
&  \textbf{FPR95$\downarrow$} & \textbf{AUROC$\uparrow$}
&  \textbf{FPR95$\downarrow$} & \textbf{AUROC$\uparrow$}   \\
\midrule
$S_\text{MAX}$
&&60.58 & 85.48 & 100.00 & 73.18 & 100.00  & 85.39 & 86.86 & 81.35 \\
$S_\text{Energy}$
&&78.82 & 78.43 & 78.98 & 75.64& 45.89 & 88.28& 67.90 & 80.78 \\
$S_\text{MaxLogit}$
&& 75.65 & 77.72& 74.21 & 75.97&44.90 & 87.89& 64.92 & 80.53 \\
$S_\text{MSP}$
&& \textbf{55.88} &\textbf{86.56} &78.05 & 74.87& 41.04 & 91.61& 58.32 & 84.35 \\
$S_\text{EOE}$
&& 57.55 & 85.64 &\textbf{73.63} & \textbf{76.93} & \textbf{37.95} & \textbf{91.96} &  \textbf{56.37} & \textbf{84.85} \\
\bottomrule
\end{tabular}
\end{table}

\subsection{LLM Prompts}\label{app: ablation-llm-prompts}
We performed additional ablation study experiments for LLM prompts on the same three datasets as in Section~\ref{app: score function}. The results are shown in Table~\ref{app:tab-llm-prompts}, and our resemble prompt consistently outperforms other prompts in these challenging scenarios.
We provide specific examples of \textit{`dissimilar'} and \textit{`irrelevant'} LLM prompts in Figure~\ref{app fig: dissimilar LLM prompts} and Figure~\ref{app fig: irrelevant LLM prompts}, respectively. The LLM chosen is GPT-3.5-turbo-16k.

\begin{table}[t]
\centering
\caption{Additional ablation studies on LLM prompts. The \textbf{bold} indicates the best performance on each dataset.}
\label{app:tab-llm-prompts}
\fontsize{7}{8}\selectfont
\setlength\tabcolsep{4.5pt}
\begin{tabular}{clcccccccc}
\toprule

\multirow{2}{*}{\textbf{LLM Prompt}} & \textbf{ID} & \multicolumn{2}{c}{ImageNet-1K} & \multicolumn{2}{c}{ImageNet-1K}& \multicolumn{2}{c}{Food-50} & \multicolumn{2}{c}{\multirow{2}{*}{\textbf{Average}}} \\
 & \textbf{OOD}        &  \multicolumn{2}{c}{Texture} & \multicolumn{2}{c}{NINCO}& \multicolumn{2}{c}{Food-50} \\

\cmidrule(rl){3-4}
\cmidrule(rl){5-6}
\cmidrule(rl){7-8}
\cmidrule(rl){9-10}
&&  \textbf{FPR95$\downarrow$} & \textbf{AUROC$\uparrow$}               
&  \textbf{FPR95$\downarrow$} & \textbf{AUROC$\uparrow$}
&  \textbf{FPR95$\downarrow$} & \textbf{AUROC$\uparrow$}
&  \textbf{FPR95$\downarrow$} & \textbf{AUROC$\uparrow$}   \\
\midrule
Irrelevant
&& 60.69 & 85.14&79.36 & 74.30& 41.89 & 90.74& 60.64 & 83.39 \\
Dissimilar
&&  58.86 & 85.29&75.10 & 76.78& 40.00 & 91.67& 57.98 & 84.58\\
Resemble
&& \textbf{57.55} & \textbf{85.64}& \textbf{73.63} & \textbf{76.93}  &\textbf{37.95} & \textbf{91.96}& \textbf{56.37} &\textbf{84.85}\\
\bottomrule
\end{tabular}
\end{table}

\begin{figure}[!ht]
\centering
\begin{tikzpicture}
\small
\definecolor{chatcolor1}{HTML}{5fedb7} 
\definecolor{shadecolor}{gray}{0.9}
\fontfamily{cmss}\selectfont

\node[align=left, text width=0.85\textwidth, fill=shadecolor, rounded corners=1mm, anchor=north west] (node1) at ([yshift=-0.4cm]node1.south west |- node2.south) {
\textbf{Q:} Given the image category [water jug], please suggest visually dissimilar categories that are not directly related or belong to the same primary group as [water jug]. Provide suggestions that do not share visual characteristics but are from broader and different domains than [water jug].\\
\textbf{A:} There are three classes dissimilar to [water jug], and they are from broader and different domains than [water jug]:\\
- trumpets\\
- helmets\\
- rucksacks\\
\vspace{1\baselineskip}
\textbf{Q:} Given the image category [horse], please suggest visually dissimilar categories that are not directly related or belong to the same primary group as [horse]. Provide suggestions that do not share visual characteristics but are from broader and different domains than [horse].\\
\textbf{A:} There are three classes dissimilar to [horse], and they are from broader and different domains than [horse]:};
\node[align=left, text width=0.12\textwidth, fill=chatcolor1, rounded corners=1mm, anchor=north east] (node2) at ([yshift=-0.2cm]node1.south -| {$(0,0)+(0.95\textwidth,0)$}) {- pineapple\\
- laptop\\
- mountain\\};
\node[anchor=west, font=\selectfont] at (node1.south west |- node2) {ID class label: horse};

\node[draw, black, thick, rounded corners=3mm, inner sep=7pt, fit=(node1) (node2)] {};

\end{tikzpicture}
\vspace{+2pt}
\caption{Instance of \textit{`dissimilar'} LLM prompt for OOD detection, ID class label: \textit{horse}. Note that, the gray is the LLM prompt for near OOD detection, and the green is the LLM actually returns. 
}
\label{app fig: dissimilar LLM prompts}
\end{figure}

\begin{figure}[!ht]
\centering
\begin{tikzpicture}
\small
\definecolor{chatcolor1}{HTML}{5fedb7} 
\definecolor{shadecolor}{gray}{0.9}
\fontfamily{cmss}\selectfont

\node[align=left, text width=0.85\textwidth, fill=shadecolor, rounded corners=1mm, anchor=north west] (node1) at ([yshift=-0.4cm]node1.south west |- node2.south) {
\textbf{Q:} Given the image category [water jug], please suggest categories that are not directly related or belong to the same primary group as [water jug].\\
\textbf{A:} There are three classes from broader and different domains than [water jug]:\\
- trumpets\\
- helmets\\
- rucksacks\\
\vspace{1\baselineskip}
\textbf{Q:} Given the image category [horse], please suggest categories that are not directly related or belong to the same primary group as [horse].\\
\textbf{A:} There are three classes from broader and different domains than [horse]:};
\node[align=left, text width=0.12\textwidth, fill=chatcolor1, rounded corners=1mm, anchor=north east] (node2) at ([yshift=-0.2cm]node1.south -| {$(0,0)+(0.95\textwidth,0)$}) {- pineapple\\
- laptop\\
- sunglasses\\};
\node[anchor=west, font=\selectfont] at (node1.south west |- node2) {ID class label: horse};

\node[draw, black, thick, rounded corners=3mm, inner sep=7pt, fit=(node1) (node2)] {};

\end{tikzpicture}
\vspace{+2pt}
\caption{Instance of \textit{`irrelevant'} LLM prompt for OOD detection, ID class label: \textit{horse}. Note that gray is the LLM prompt for near OOD detection, and green is the LLM that actually returns.}
\label{app fig: irrelevant LLM prompts}
\end{figure}

\subsection{Various LLMs}\label{app:ablaton-llms}
We conducted additional ablation studies on various LLMs using the same three datasets mentioned in Section~\ref{app: score function}. The results, presented in Table~\ref{app:tab-llms}, indicate that our strategy is effective across different LLMs. Note that due to safety precautions, Claude2 declines to process large-scale datasets such as ImageNet-1K. And LLaMA2-7B's responses do not adhere strictly to our predefined JSON format. Consequently, we employed alternative LLMs, including Mixtral-8x7B-Chat~\citep{jiang2024mixtral}, Gemini-Pro~\citep{reid2024gemini}, Claude-3-Haiku~\citep{anthropic2024claude}, and GPT-4~\citep{gpt4}.

\begin{table}[t]
\centering
\caption{Additional ablation studies on various LLMs. The \textbf{bold} indicates the best performance on each dataset.}
\label{app:tab-llms}
\fontsize{7}{8}\selectfont
\setlength\tabcolsep{4.5pt}
\begin{tabular}{clcccccccc}
\toprule

\multirow{2}{*}{\textbf{LLM}} & \textbf{ID} & \multicolumn{2}{c}{ImageNet-1K} & \multicolumn{2}{c}{ImageNet-1K}& \multicolumn{2}{c}{Food-50} & \multicolumn{2}{c}{\multirow{2}{*}{\textbf{Average}}} \\
 & \textbf{OOD}        &  \multicolumn{2}{c}{Texture} & \multicolumn{2}{c}{NINCO}& \multicolumn{2}{c}{Food-50} \\

\cmidrule(rl){3-4}
\cmidrule(rl){5-6}
\cmidrule(rl){7-8}
\cmidrule(rl){9-10}
&&  \textbf{FPR95$\downarrow$} & \textbf{AUROC$\uparrow$}               
&  \textbf{FPR95$\downarrow$} & \textbf{AUROC$\uparrow$}
&  \textbf{FPR95$\downarrow$} & \textbf{AUROC$\uparrow$}
&  \textbf{FPR95$\downarrow$} & \textbf{AUROC$\uparrow$}   \\
\midrule
MCM
&&57.85 & 86.11 &79.40 & 74.27  & 43.48 & 91.75 & 60.21 & 84.04 \\
EOE(Ours, Mixtral-8x7B-Chat)
&& 56.95 & 86.02 &73.70 & 75.50 & 33.48 & 93.28& 54.71 & 84.93\\
EOE(Ours, Claude-3-Haiku)
&&  54.4 & 86.58 & 73.80 & 75.99 &37.08 &92.44 &55.09 & 85.00\\
EOE(Ours, Gemini-Pro) 
&& 56.56 &  86.85 & 72.88 &  75.67 & 33.73 &  93.02&  54.39 &  85.19\\
EOE(Ours, GPT-3.5-turbo-16k)
&& 57.55 & 85.64 & 73.63 & 76.93 &37.95 & 91.96 & 56.37 & 84.85\\
EOE(Ours, GPT-4)
&& 53.65 & 87.05 & 73.41 & 74.96& 35.66 & 93.10& 54.24 & 85.04
\\
\bottomrule
\end{tabular}
\end{table}

\section{Limitations}\label{app: limitation}
The results are shown in Table~\ref{app: limitation-tab}, without prior knowledge of the OOD task. 
In the majority of scenarios, the performance of our EOE remains superior to that of the MCM.

\begin{table}[htbp!]
\centering
\caption{Using \textbf{far} OOD detection prompt to envision outlier class labels on Zero-shot \textbf{near} and \textbf{fine-grained} OOD detection tasks. The \textcolor{teal}{green} indicates Our EOE outperforms the strong baseline MCM.}
\fontsize{7}{8}\selectfont
\setlength\tabcolsep{2.8pt}
\begin{tabular}{cl*{12}c}
\toprule
 && \multicolumn{4}{c}{near OOD detection} & \multicolumn{8}{c}{fine-grained OOD detection}\\
\cmidrule(rl){3-6}
\cmidrule(rl){7-14}
\multirow{2}{*}{\textbf{Method}} & \textbf{ID} & \multicolumn{2}{c}{ImageNet-10} & \multicolumn{2}{c}{ImageNet-20} &\multicolumn{2}{c}{CUB-100} & \multicolumn{2}{c}{Stanford-Cars-98} & \multicolumn{2}{c}{Food-50} & \multicolumn{2}{c}{Oxford-Pet-18} \\
 & \textbf{OOD}   & \multicolumn{2}{c}{ImageNet-20} & \multicolumn{2}{c}{ImageNet-10}     &  \multicolumn{2}{c}{CUB-100} & \multicolumn{2}{c}{Stanford-Cars-98} & \multicolumn{2}{c}{Food-51} & \multicolumn{2}{c}{Oxford-Pet-19}  \\

\cmidrule(rl){3-4}
\cmidrule(rl){5-6}
\cmidrule(rl){7-8}
\cmidrule(rl){9-10}
\cmidrule(rl){11-12}
\cmidrule(rl){13-14}
&&  \textbf{FPR95$\downarrow$} & \textbf{AUROC$\uparrow$}               
&  \textbf{FPR95$\downarrow$} & \textbf{AUROC$\uparrow$}
&  \textbf{FPR95$\downarrow$} & \textbf{AUROC$\uparrow$}
&  \textbf{FPR95$\downarrow$} & \textbf{AUROC$\uparrow$}
&  \textbf{FPR95$\downarrow$} & \textbf{AUROC$\uparrow$}
&  \textbf{FPR95$\downarrow$} & \textbf{AUROC$\uparrow$}   \\
\midrule
MCM   &&
5.00 & 98.71  & 17.40 & 97.87 &
83.58 & 67.51 & 83.99 & 68.71 & 43.38 & 91.75 & 63.92 & 84.88 \\
EOE~(Ours)   
&& 
4.20 & 99.09 & 13.93 & 98.10 &
74.74 & 73.41 & 76.83 & 71.60 & 37.95 & 91.96 & 52.55 & 90.33 \\
EOE~(Ours) (far prompt) && 
5.47 &  \textcolor{teal}{98.74} & \textcolor{teal}{11.40} &  \textcolor{teal}{98.21} &
\textcolor{teal}{77.84} & \textcolor{teal}{72.29} & \textcolor{teal}{76.24} & \textcolor{teal}{73.03} & \textcolor{teal}{39.98} & 91.62 & 67.27 & \textcolor{teal}{88.21} \\
\bottomrule
\end{tabular}
\label{app: limitation-tab}
\end{table}

\section{Specific example of LLM prompts}
\label{example of LLM prompt}
To obtain the outlier class labels automatically generated by the LLM, we provide a question and answer template for LLM and append `-' to the end of the answer to produce a bullet-point list output. It is worth noting that the provided template does not contain any ID class content.
Below, we present specific examples for three LLM prompts. The LLM prompt for far OOD detection is depicted in Figure~\ref{app fig: far LLM prompts}. The LLM prompts for near OOD detection and fine-grained OOD detection are showcased in Figure~\ref{app fig: near LLM prompts} and Figure~\ref{app fig: fine-grained LLM prompts}, respectively. 

To better understand the outlier class labels generated by LLMs, we show in Figure~\ref{app fig: outlier_imagenet-1k} the outlier class labels generated by GPT-3.5-turbo-16k for ImageNet-1K~(ID), which were instructed under the far OOD prompt. 

\begin{figure}[!ht]
\centering
\begin{tikzpicture}
\small
\definecolor{chatcolor1}{HTML}{5fedb7} 
\definecolor{shadecolor}{gray}{0.9}
\fontfamily{cmss}\selectfont

\node[align=left, text width=0.85\textwidth, fill=shadecolor, rounded corners=1mm, anchor=north west] (node1) at ([yshift=-0.4cm]node1.south west |- node2.south) {
\textbf{Q:} I have gathered images of 4 distinct categories: ['Husky dog', 'Garfield cat', 'churches', 'truck']. Summarize what broad categories these categories might fall into based on visual features. Now, I am looking to identify 5 categories that visually resemble these broad categories but have no direct relation to these broad categories. Please list these 5 items for me.\\
\textbf{A:} These 5 items are:\\
- black stone\\
- mountain\\
- Ginkgo Tree\\
- river\\
- Rapeseed\\
\vspace{1\baselineskip}
\textbf{Q:} I have gathered images of 100 distinct categories: ['Apple pie', 'Baby back ribs', 'Baklava', 'Beef carpaccio', 'Beef tartare', 'Beet salad', 'Beignets', 'Bibimbap', 'Bread pudding', 'Breakfast burrito', 'Bruschetta', 'Caesar salad', 'Cannoli', 'Caprese salad', 'Carrot cake', 'Ceviche', 'Cheesecake', 'Cheese plate', 'Chicken curry', 'Chicken quesadilla', 'Chicken wings', 'Chocolate cake', 'Chocolate mousse', 'Churros', 'Clam chowder', 'Club sandwich', 'Crab cakes', 'Creme brulee', 'Croque madame', 'Cup cakes', 'Deviled eggs', 'Donuts', 'Dumplings', 'Edamame', 'Eggs benedict', 'Escargots', 'Falafel', 'Filet mignon', 'Fish and chips', 'Foie gras', 'French fries', 'French onion soup', 'French toast', 'Fried calamari', 'Fried rice', 'Frozen yogurt', 'Garlic bread', 'Gnocchi', 'Greek salad', 'Grilled cheese sandwich', 'Grilled salmon', 'Guacamole', 'Gyoza', 'Hamburger', 'Hot and sour soup', 'Hot dog', 'Huevos rancheros', 'Hummus', 'Ice cream', 'Lasagna', 'Lobster bisque', 'Lobster roll sandwich', 'Macaroni and cheese', 'Macarons', 'Miso soup', 'Mussels', 'Nachos', 'Omelette', 'Onion rings', 'Oysters', 'Pad thai', 'Paella', 'Pancakes', 'Panna cotta', 'Peking duck', 'Pho', 'Pizza', 'Pork chop', 'Poutine', 'Prime rib', 'Pulled pork sandwich', 'Ramen', 'Ravioli', 'Red velvet cake', 'Risotto', 'Samosa', 'Sashimi', 'Scallops', 'Seaweed salad', 'Shrimp and grits', 'Spaghetti bolognese', 'Spaghetti carbonara', 'Spring rolls', 'Steak', 'Strawberry shortcake', 'Sushi', 'Tacos', 'Takoyaki', 'Tiramisu', 'Tuna tartare', 'Waffles']. Summarize what broad categories these categories might fall into based on visual features. Now, I am looking to identify 50 classes that visually resemble these broad categories but have no direct relation to these broad categories. Please list these 50 items for me.\\
\textbf{A:} These 50 items are:\\};
\node[align=left, text width=0.15\textwidth, fill=chatcolor1, rounded corners=1mm, anchor=north east] (node2) at ([yshift=-0.2cm]node1.south -| {$(0,0)+(0.95\textwidth,0)$}) {- Orchid\\
- Watermelon\\
- Sunflower\\
- ...\\
- Neon graffiti\\};
\node[anchor=west, font=\selectfont] at (node1.south west |- node2) {ID dataset: Food-101};

\node[draw, black, thick, rounded corners=3mm, inner sep=7pt, fit=(node1) (node2)] {};

\end{tikzpicture}
\vspace{+2pt}
\caption{Instance of LLM prompt for far OOD detection, ID dataset: Food-101. Note that, the gray is the LLM prompt for far OOD detection, and the green is the LLM actually returns. 
}
\label{app fig: far LLM prompts}
\end{figure}

\begin{figure}[!ht]
\centering
\begin{tikzpicture}
\small
\definecolor{chatcolor1}{HTML}{5fedb7} 
\definecolor{shadecolor}{gray}{0.9}
\fontfamily{cmss}\selectfont

\node[align=left, text width=0.85\textwidth, fill=shadecolor, rounded corners=1mm, anchor=north west] (node1) at ([yshift=-0.4cm]node1.south west |- node2.south) {
\textbf{Q:} Given the image category [water jug], please suggest visually similar categories that are not directly related or belong to the same primary group as [water jug]. Provide suggestions that share visual characteristics but are from broader and different domains than [water jug].\\
\textbf{A:} There are three classes similar to [water jug], and they are from broader and different domains than [water jug]:\\
- trumpets\\
- helmets\\
- rucksacks\\
\vspace{1\baselineskip}
\textbf{Q:} Given the image category [horse], please suggest visually similar categories that are not directly related or belong to the same primary group as [horse]. Provide suggestions that share visual characteristics but are from broader and different domains than [horse].\\
\textbf{A:} There are three classes similar to [horse], and they are from broader and different domains than [horse]:};
\node[align=left, text width=0.1\textwidth, fill=chatcolor1, rounded corners=1mm, anchor=north east] (node2) at ([yshift=-0.2cm]node1.south -| {$(0,0)+(0.95\textwidth,0)$}) {- zebra\\
- giraffe\\
- deer\\};
\node[anchor=west, font=\selectfont] at (node1.south west |- node2) {ID class label: horse};

\node[draw, black, thick, rounded corners=3mm, inner sep=7pt, fit=(node1) (node2)] {};

\end{tikzpicture}
\vspace{+2pt}
\caption{Instance of LLM prompt for near OOD detection, ID class label: \textit{horse}. Note that, the gray is the LLM prompt for near OOD detection, and the green is the LLM actually returns. 
}
\label{app fig: near LLM prompts}
\end{figure}

\begin{figure}[!ht]
\centering
\begin{tikzpicture}
\small
\definecolor{chatcolor1}{HTML}{5fedb7} 
\definecolor{shadecolor}{gray}{0.9}
\fontfamily{cmss}\selectfont

\node[align=left, text width=0.85\textwidth, fill=shadecolor, rounded corners=1mm, anchor=north west] (node1) at ([yshift=-0.4cm]node1.south west |- node2.south) {
\textbf{Q:} I have a dataset containing 10 unique species of dogs. I need a list of 10 distinct dog species that are NOT present in my dataset, and ensure there are no repetitions in the list you provide. For context, the species in my dataset are: ['husky dog', 'alaskan Malamute', 'cossack sled dog', 'golden retriever', 'German Shepherd', 'Beagle', 'Bulldog', 'Poodle', 'Dachshund', 'Doberman Pinscher']\\
\textbf{A:} The other 10 dog species not in the dataset are:\\
- Labrador Retriever\\
- Rottweiler\\
- Boxer\\
- Border Collie\\
- Shih Tzu\\
- Akita\\
- Saint Bernard\\
- Australian Shepherd\\
- Great Dane\\
- Boston Terrier\\
\vspace{1\baselineskip}
\textbf{Q:} I have a dataset containing 50 different species of food. I need a list of 50 distinct food species that are NOT present in my dataset, and ensure there are no repetitions in the list you provide. For context, the species in my dataset are: ['Prime Rib', 'Dumplings', 'Strawberry Shortcake', 'Frozen Yogurt', 'Seaweed Salad', 'Tiramisu', 'Red Velvet Cake', 'Omelette', 'Beef Carpaccio', 'Lasagna', 'Donuts', 'Sushi', 'Beignets', 'Chicken Wings', 'Carrot Cake', 'Gnocchi', 'Lobster Bisque', 'Spaghetti Bolognese', 'Greek Salad', 'Oysters', 'Caprese Salad', 'Panna Cotta', 'Shrimp And Grits', 'Baby Back Ribs', 'Creme Brulee', 'Gyoza', 'Escargots', 'Churros', 'Grilled Cheese Sandwich', 'Scallops', 'Breakfast Burrito', 'Cheesecake', 'Huevos Rancheros', 'Cheese Plate', 'Steak', 'Apple Pie', 'Mussels', 'Crab Cakes', 'Pancakes', 'Pulled Pork Sandwich', 'Bruschetta', 'Hot Dog', 'Risotto', 'Chicken Curry', 'Paella', 'Cannoli', 'Eggs Benedict', 'Fried Calamari', 'French Fries', 'Lobster Roll Sandwich']\\
\textbf{A:} The other 50 food species not in the dataset are:};
\node[align=left, text width=0.15\textwidth, fill=chatcolor1, rounded corners=1mm, anchor=north east] (node2) at ([yshift=-0.2cm]node1.south -| {$(0,0)+(0.95\textwidth,0)$}) {
- Lychee\\
- chicken tacos\\
- Durian\\
- ...\\
- affogato\\
- pizza\\};
\node[anchor=west, font=\selectfont] at (node1.south west |- node2) {ID dataset: Food50};

\node[draw, black, thick, rounded corners=3mm, inner sep=7pt, fit=(node1) (node2)] {};

\end{tikzpicture}
\vspace{+2pt}
\caption{Instance of LLM prompt for fine-grained OOD detection, ID dataset: Food-50. Note that, the gray is the LLM prompt for fine-grained OOD detection, and the green is the LLM actually returns. 
}
\label{app fig: fine-grained LLM prompts}
\end{figure}

\begin{figure}[!ht]
\centering
\begin{tikzpicture}
\small
\definecolor{chatcolor1}{HTML}{5fedb7} 
\definecolor{shadecolor}{gray}{0.9}
\fontfamily{cmss}\selectfont

\node[align=left, text width=0.95\textwidth, fill=shadecolor, rounded corners=1mm, anchor=north west] (node1) at ([yshift=-0.4cm]node1.south west |- node2.south) {
\textbf{Outlier label for ImageNet-1K:}"Sunflower", "Waterfall", "Desert", "Sunset", "Rainbow", "Snowflake", "Aurora borealis", "Lightning", "Galaxy", "Moon", "Star", "Cloud", "Forest", "Mountain range", "Beach sunset", "Autumn leaves", "Spring flowers", "Winter landscape", "Summer beach", "City skyline", "Countryside", "Ocean waves", "Water droplet", "Fireworks", "Hot air balloon", "Iceberg", "Sand dunes", "Tropical island", "Rainforest", "Canyon", "Glacier", "Tornado", "Volcanic eruption", "Underwater coral reef", "Safari wildlife", "Desert oasis", "Northern lights", "Full moon", "Shooting star", "Thunderstorm", "Milky Way", "Rolling hills", "Waterfall", "Sunset over the ocean", "Snow-capped mountains", "Lush green meadow", "Colorful autumn foliage", "Blooming cherry blossoms", "Frozen lake", "Sandy beach with palm trees- Desert cactus", "Tropical rainforest", "Snowy mountain peak", "Vibrant sunset sky", "Misty waterfall", "Sandy beachscape", "Autumn forest scene", "Lush green valley", "Colorful flower field", "Majestic ocean view", "Serene lake scene", "Rolling countryside hills", "Urban cityscape skyline", "Starry night sky", "Thunderstorm lightning", "Fiery sunset horizon", "Tranquil river scene", "Snowy winter landscape", "Blossoming cherry trees", "Vibrant coral reef", "Rocky canyon walls", "Frozen Arctic tundra", "Exotic tropical island", "Dense jungle foliage", "Majestic volcano eruption", "Serene moonlit night", "Milky Way galaxy", "Vibrant rainbow arc", "Sandy desert dunes", "Crystal clear waterfall", "Autumn foliage reflection", "Snow-covered pine forest", "Sunflower field", "Vibrant coral reef", "Rolling green hills", "Sandy beach with palm trees", "Colorful hot air balloons", "Frozen lake with mountains", "Lush green meadow", "Blooming cherry blossoms", "Rocky mountain range", "Serene ocean waves", "Thunderstorm with lightning", "Milky Way galaxy", "Vibrant sunset over water", "Snowy alpine landscape", "Peaceful countryside scene", "Colorful spring flowers", "Tropical island paradise", "Desert oasis with palm trees- Cherry blossom tree", "Snowy mountain range", "Golden sunset beach", "Misty forest scene", "Colorful tulip field", "Serene lake reflection", "Rolling green meadows", "Vibrant butterfly garden", "Majestic waterfall cascade", "Tranquil riverbank scene", "Snowy winter wonderland", "Blooming sunflower field", "Urban city skyline", "Starry night landscape", "Thunderstorm cloudscape", "Fiery sunset over mountains", "Crystal clear river", "Autumn foliage carpet", "Tropical coral reef", "Rocky desert landscape", "Frozen Arctic wilderness", "Exotic palm-fringed island", "Dense tropical rainforest", "Majestic volcanic crater", "Serene moonlit beach", "Milky Way starry sky", "Vibrant double rainbow", "Sandy desert oasis", "Cascading waterfall pool", "Autumn forest path", "Snow-covered alpine peaks", "Sunflower garden", "Vibrant underwater coral", "Rolling countryside scenery", "Sandy beach paradise", "Colorful hot air balloons", "Frozen lake panorama", "Lush green pasture", "Blooming cherry orchard", "Rocky mountain summit", "Serene ocean sunset", "Thunderstorm lightning bolts", "Milky Way galaxy cluster", "Vibrant sunset over mountains", "Snowy pine forest", "Peaceful countryside road", "Colorful spring garden", "Tropical island getaway", "Desert oasis with camels- Lavender field", "Snow-capped peaks", "Golden sunset sky", "Misty woodland scene", "Colorful wildflowers meadow", "Serene lake landscape", "Rolling green hills", "Vibrant butterfly garden", "Majestic waterfall view", "Tranquil riverbank scenery", "Snowy winter forest", "Blooming poppy field", "Urban cityscape view", "Starry night panorama", "Thunderstorm clouds", "Fiery sunset over ocean", "Crystal clear stream", "Autumn foliage trail", "Tropical coral reef", "Rocky desert canyon", "Frozen Arctic landscape", "Exotic palm-fringed beach", "Dense rainforest canopy", "Majestic volcanic eruption", "Serene moonlit lake", "Milky Way galaxy view", "Vibrant rainbow colors", "Sandy desert dunes", "Cascading waterfall pool", "Autumn forest path", "Snow-covered alpine scenery", "Sunflower bouquet", "Vibrant underwater life", "Rolling countryside fields", "Sandy beach paradise", "Colorful hot air balloons", "Frozen lake reflection", "Lush green meadow", "Blooming cherry blossom", "Rocky mountain range", "Serene ocean waves", "Thunderstorm lightning", "Milky Way galaxy cluster", "Vibrant sunset over water", "Snowy pine forest", "Peaceful countryside road", "Colorful spring garden", "Tropical island getaway", "Desert oasis with camels- Lavender field", "Snow-capped peaks", "Golden sunset sky", "Misty woodland scene", "Colorful wildflowers meadow", "Serene lake landscape", "Rolling green hills", "Vibrant butterfly garden", "Majestic waterfall view", "Tranquil riverbank scenery", "Snowy winter forest", "Blooming poppy field", "Urban cityscape view", "Starry night panorama", "Thunderstorm clouds", "Fiery sunset over ocean", "Crystal clear stream", "Autumn foliage trail", "Tropical coral reef", "Rocky desert canyon", "Frozen Arctic landscape", "Exotic palm-fringed beach", "Dense rainforest canopy", "Majestic volcanic eruption", "Serene moonlit lake", "Milky Way galaxy view", "Vibrant rainbow colors", "Sandy desert dunes", "Cascading waterfall pool", "Autumn forest path", "Snow-covered alpine scenery", "Sunflower bouquet", "Vibrant underwater life", "Rolling countryside fields", "Sandy beach paradise", "Colorful hot air balloons", "Frozen lake reflection", "Lush green meadow", "Blooming cherry blossom", "Rocky mountain range", "Serene ocean waves", "Thunderstorm lightning", "Milky Way galaxy cluster", "Vibrant sunset over water", "Snowy pine forest", "Peaceful countryside road", "Colorful spring garden", "Tropical island getaway", "Desert oasis with camels- Lavender field", "Snow-capped peaks", "Golden sunset sky", "Misty woodland scene", "Colorful wildflowers meadow", "Serene lake landscape", "Rolling green hills", "Vibrant butterfly garden", "Majestic waterfall view", "Tranquil riverbank scenery", "Snowy winter forest", "Blooming poppy field", "Urban cityscape view", "Starry night panorama", "Thunderstorm clouds", "Fiery sunset over ocean", "Crystal clear stream", "Autumn foliage trail", "Tropical coral reef", "Rocky desert canyon", "Frozen Arctic landscape", "Exotic palm-fringed beach", "Dense rainforest canopy", "Majestic volcanic eruption", "Serene moonlit lake", "Milky Way galaxy view", "Vibrant rainbow colors", "Sandy desert dunes", "Cascading waterfall pool", "Autumn forest path", "Snow-covered alpine scenery", "Sunflower bouquet", "Vibrant underwater life", "Rolling countryside fields", "Sandy beach paradise", "Colorful hot air balloons", "Frozen lake reflection", "Lush green meadow", "Blooming cherry blossom", "Rocky mountain range", "Serene ocean waves", "Thunderstorm lightning", "Milky Way galaxy cluster", "Vibrant sunset over water", "Snowy pine forest", "Peaceful countryside road", "Colorful spring garden", "Tropical island getaway", "Desert oasis with camels"};
\node[draw, black, thick, rounded corners=3mm, inner sep=7pt, fit=(node1)]{};
\end{tikzpicture}
\vspace{+2pt}
\caption{Outlier class labels generated by GPT-3.5-turbo-16k for ImageNet-1K. Due to space constraints, we show results for $L=300$ here. 
}
\label{app fig: outlier_imagenet-1k}
\end{figure}

\end{document}